\documentclass[fleqn,twoside]{article}
\usepackage{espcrc1}

\newcommand{\ttbs}{\char'134}
\newcommand{\AmS}{{\protect\the\textfont2
  A\kern-.1667em\lower.5ex\hbox{M}\kern-.125emS}}

\title{Time-interval balancing in multi-processor
 scheduling of composite modular jobs\\
 (preliminary description)}

\author{Mark Sh. Levin
\thanks{
 Mark Sh. Levin:~
 Inst. for Inform. Transmission Problems,
 Russian Academy of Sciences;
  http://www.mslevin.iitp.ru;
 email: mslevin@acm.org
  } }

\begin{document}

\maketitle

\begin{abstract}
 The article describes a special time-interval balancing in multi-processor
 scheduling of composite modular jobs.
 This scheduling problem is close
 to just-in-time planning approach.
%
%
 First,
 brief literature surveys are presented on
 just-in-time scheduling and
 due-data/due-window scheduling problems.
 Further,
 the problem  and its formulation are proposed for
 the time-interval balanced scheduling
 of composite modular jobs.

 The illustrative real world planning example for
 modular home-building is described.
 Here, the main objective function consists in a balance  between
 production
 of the typical building modules (details)
 and the assembly processes of the building(s)
 (by several teams).
%
%
%
 The assembly plan has to be modified
 to satisfy the balance requirements.
%
%
%
%
%
%
 The solving framework is based on the following:
 (i) clustering of initial set of modular detail types to obtain
 about ten basic detail types that correspond to
 main manufacturing conveyors;
 (ii) designing a preliminary plan of assembly for buildings
 (time interval is about 2 years);
 (iii) detection of unbalanced time periods,
 (iv) modification of the planning solution to improve
 the schedule balance.
 The framework implements a metaheuristic  based on local
 optimization approach.

 Two other applications
 (supply chain management, information transmission systems)
 are briefly described.
%

~~


\vspace{1pc}
\end{abstract}

\maketitle

\tableofcontents

\newcounter{cms}
\setlength{\unitlength}{1mm}

\newpage
\section{Introduction}

  In general,
 some special problems are under examination in which
  chains/sequences of objects/items are clustered/located.
 For example, clustering and location
 of  chains into bins are considered in \cite{batu12}.
 Usually, applications of the problems correspond to planning in
 manufacturing systems and computing systems.
%
%
 %
 It is reasonable to point out,
 various balanced clustering problems and
 balanced combinatorial optimization problems
 have been examined
 (e.g., balanced knapsack problems,
 balanced assignment problems,
 balanced bin packing problems,
 balanced scheduling problems).
%
%
 The basic author glance at balanced clustering problems
 is contained in \cite{lev17bal1,lev17balj}.
%
%
 In the paper,
 a special new time-interval balancing
 in multi-processor scheduling of
 composite modular jobs is examined.
 Here, the following hierarchy is under consideration:
 (1) a set of basic elements (i.e., basic modules);
 (2) an initial set of composite jobs/tasks (for multi-processor scheduling)
 consisting of the above-mentioned basic elements;
 and
 (3) the resultant multi-processor scheduling while taking into account
 proportion-based constraints
 (by proportion of the used basic elements)
  for each time interval.

 The consideration is mainly based on a special scheduling problem
 for time-interval balanced assembly in modular home-building.
 Fig. 1 depicts a relationship of the examined problem
 and some close JIT scheduling/planning problems.

\begin{center}
\begin{picture}(120,87)

\put(16,00){\makebox(0,0)[bl]{Fig. 1.
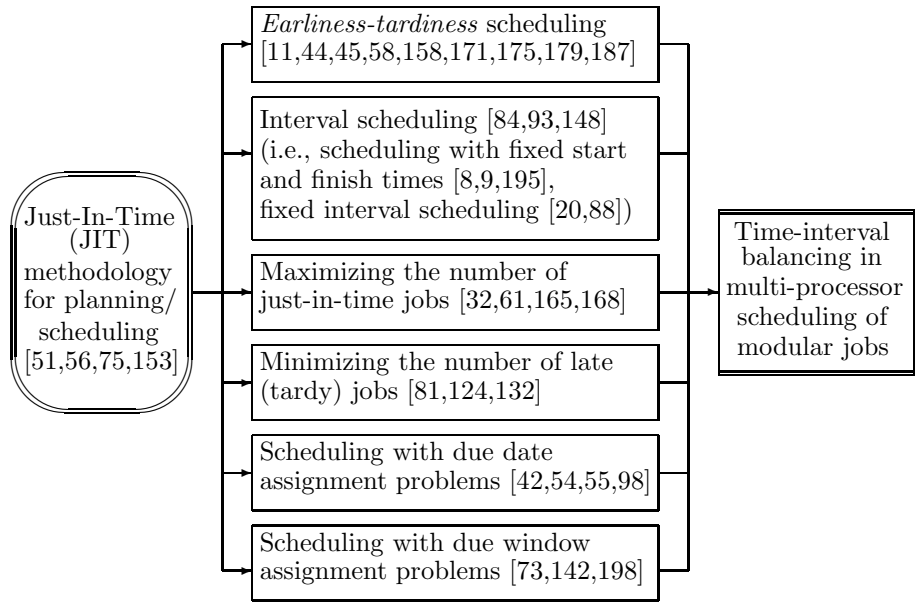
  Scheme of some JIT relative scheduling problems}}

\put(33,81){\makebox(0,0)[bl]{{\it Earliness-tardiness}
 scheduling}}

\put(33,77){\makebox(0,0)[bl]{\cite{bak90,gar79,gar88,hall91,ronc12,sidn77,sourd08,tan12,val12}
  }}

\put(32,75){\line(1,0){54}} \put(32,85){\line(1,0){54}}
\put(32,75){\line(0,1){10}} \put(86,75){\line(0,1){10}}

\put(28,80){\vector(1,0){4}} \put(86,80){\line(1,0){4}}

\put(33,68){\makebox(0,0)[bl]{Interval scheduling
  \cite{kolen07,krum11,papa98} }}


\put(33,64){\makebox(0,0)[bl]{(i.e., scheduling with fixed start}}

\put(33,60){\makebox(0,0)[bl]{and finish times
 \cite{ang11,ark87,woe94}, }}

\put(33,56){\makebox(0,0)[bl]{fixed interval scheduling
  \cite{branda16,kov07}) }}


\put(32,54){\line(1,0){54}} \put(32,73){\line(1,0){54}}
\put(32,54){\line(0,1){19}} \put(86,54){\line(0,1){19}}

\put(28,65.5){\vector(1,0){4}} \put(86,65.5){\line(1,0){4}}

\put(33,48){\makebox(0,0)[bl]{Maximizing the number of}}
\put(33,44){\makebox(0,0)[bl]{just-in-time jobs
 \cite{choi07,hir02,shab12,shab12b} }}

\put(32,42){\line(1,0){54}} \put(32,52){\line(1,0){54}}
\put(32,42){\line(0,1){10}} \put(86,42){\line(0,1){10}}

\put(28,47){\vector(1,0){4}} \put(86,47){\line(1,0){4}}




\put(33,36){\makebox(0,0)[bl]{Minimizing the number of late}}
\put(33,32){\makebox(0,0)[bl]{(tardy) jobs
 \cite{kay14,mhal05,moor68} }}

\put(32,30){\line(1,0){54}} \put(32,40){\line(1,0){54}}
\put(32,30){\line(0,1){10}} \put(86,30){\line(0,1){10}}

\put(28,35){\vector(1,0){4}} \put(86,35){\line(1,0){4}}

\put(33,24){\makebox(0,0)[bl]{Scheduling with due date}}
\put(33,20){\makebox(0,0)[bl]{assignment problems
 \cite{emm87,gord02,gord04,lauf04} }}

\put(32,18){\line(1,0){54}} \put(32,28){\line(1,0){54}}
\put(32,18){\line(0,1){10}} \put(86,18){\line(0,1){10}}

\put(28,23){\vector(1,0){4}} \put(86,23){\line(1,0){4}}




\put(33,12){\makebox(0,0)[bl]{Scheduling with due window}}
\put(33,08){\makebox(0,0)[bl]{assignment problems
 \cite{jana15,mos09b,yeu01} }}

\put(32,06){\line(1,0){54}} \put(32,16){\line(1,0){54}}
\put(32,06){\line(0,1){10}} \put(86,06){\line(0,1){10}}

\put(28,10){\vector(1,0){4}} \put(86,10){\line(1,0){4}}



\put(90,10){\line(0,1){70}}

\put(96,54){\makebox(0,0)[bl]{Time-interval}}
\put(97,50){\makebox(0,0)[bl]{balancing in}}
\put(95,46){\makebox(0,0)[bl]{multi-processor}}
\put(96.5,42){\makebox(0,0)[bl]{scheduling of}}
\put(96.5,38){\makebox(0,0)[bl]{modular jobs}}

\put(94,36.5){\line(1,0){26}} \put(94,57.5){\line(1,0){26}}

\put(94,36){\line(1,0){26}} \put(94,58){\line(1,0){26}}
\put(94,36){\line(0,1){22}} \put(120,36){\line(0,1){22}}

\put(90,47){\vector(1,0){4}}

\put(01.6,56){\makebox(0,0)[bl]{Just-In-Time  }}
\put(07.6,52){\makebox(0,0)[bl]{(JIT) }}
\put(01.6,48){\makebox(0,0)[bl]{methodology}}
\put(01.6,44){\makebox(0,0)[bl]{for planning/}}
\put(03.6,40){\makebox(0,0)[bl]{scheduling }}

\put(01.6,36){\makebox(0,0)[bl]{\cite{golh91,groe93,joze10,rios12}}}

\put(12,47){\oval(24,32)} \put(12,47){\oval(23,31)}


\put(24,47){\line(1,0){4}} \put(28,10){\line(0,1){70}}





\end{picture}
\end{center}

 At the end of the paper,
 two other application domains for the considered problem are
  pointed out.
 This material can be considered as a special part of balanced
 clustering domain and the author
  project
 on combinatorial clustering
  \cite{lev15c,lev15d} including
  dynamic clustering
 \cite{lev15restr,lev17d}
 and balanced clustering
 \cite{lev17bal1,lev17balj}.

\newpage
\section{On Just-In-Time scheduling}

 Just-In-Time  (JIT) scheduling problems are mainly based on
 earliness-tardiness planning models
 \cite{alv12,alv15,gar79,gar88,joze10,lauf04,li06,pin16,ronc12,ros17,tan12,val12,vale09}
%
%
 and
 interval scheduling models
  \cite{kolen07,kov07,krum11,papa98}.
 JIT scheduling models assume an existence of job due dates and
 penalize  both early and tardy jobs
 (e.g., earliness penalty/cost,
 tardiness cost as late charge,  etc.).

 Further, brief literature surveys on some research directions
 in the field of JIT
 are presented as follows:

 (a) general issues on JIT scheduling (Table 1),

 (b) JIT scheduling models (Table 2),

 (c) solving (algorithmic) approaches in JIT scheduling (Table 3),
 and

 (d) basic application domains of JIT scheduling (Table 4).

%
%

\begin{center}
\begin{small}
 {\bf Table 1.} General issues on JIT scheduling\\
%
\begin{tabular}{| c | l| l |}
\hline
 No.&Issue(s)& Source(s)\\

\hline

 1.&
 Basic scheduling problems (books, surveys) &\cite{joze10,pin16,rios12}\\


 2.&JIT philosophy: a literature review &\cite{golh91}\\


 3.&Some surveys on JIT systems  &\cite{kub93,rios12}\\

 4.&Schedules for mixed-model, multi level just-in-time assembly systems
     &\cite{golh91}\\

 5.&JIT management of  building projects&\cite{pheng12}\\

 6.&JIT-KANBAN systems (literature review)
    & \cite{send07}\\


 7.&Controlling just-in-sequence flow-production&\cite{mei10}\\

 8.&Design and operational issues of kanban systems (overview)
    & \cite{akt99}\\


 9.&Robust design methodology for Kanban system design&\cite{moo97}\\

 10.&JIT production leveling &\cite{gior15}\\

 11.&JIT manufacturing system (introduction, implementation)
  &\cite{koot13}\\

\hline
\end{tabular}
\end{small}
\end{center}

%

 Let us describe two simplified illustrative examples.
 Here, an initial job (task) set is given
  \(A=\{a_{1},...,a_{i},...,a_{n}\}\),
  there are parameters for each job
  \(a_{i} \in A\):
 (a) processing time \(\theta (a_{i})\);
 (b) general time interval
  \( [0,T] \)
 for processing the job set \(A\);
 (c) interval for job \( a_{i} \) processing
 (i.e., early time and tardy time:
   \( \lambda^{a_{i}} = [ t^{a_{i}}_{1}, t^{a_{i}}_{2} ] \),
   \( t^{a_{i}}_{1} < t^{a_{i}}_{2} \);
   \( t^{a_{i}}_{1},t^{a_{i}}_{2} \in [0,T]\)).
%
%
 Numerical examples are depicted as follows:

 (a) one-machine/processor scheduling (Fig. 2, Table 5),
 \(A = \{ a_{1},...,a_{7}\}\),
 general time interval  \( [0,5.0] \);

(b) three-machine/processor scheduling (Fig. 3, Table 6),
 \(A = \{ a_{1},...,a_{12}\}\),
  general time interval  \( [0,5.0] \).

\begin{center}
\begin{picture}(84,66)
\put(03,00){\makebox(0,0)[bl]{Fig. 2.
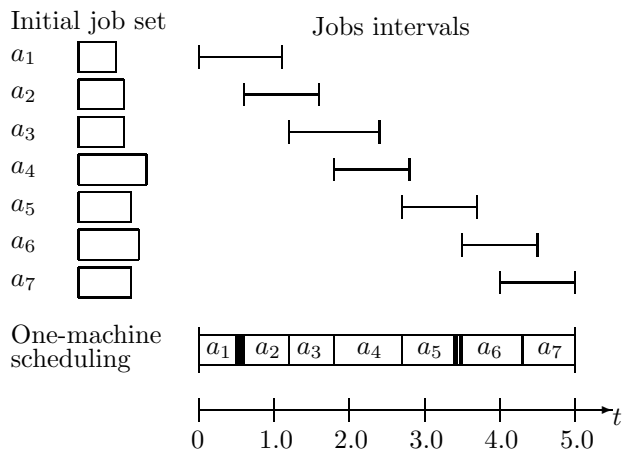
 Illustration for one-machine scheduling }}

\put(00,60){\makebox(0,0)[bl]{Initial job set}}

\put(040,60){\makebox(0,0)[bl]{Jobs intervals}}

\put(00,56){\makebox(0,0)[bl]{\(a_{1}\)}}
\put(09,55){\line(1,0){05}} \put(09,59){\line(1,0){05}}
\put(09,55){\line(0,1){04}} \put(14,55){\line(0,1){04}}

\put(25,55.5){\line(0,1){03}} \put(36,55.5){\line(0,1){03}}
\put(25,57){\line(1,0){11}}

\put(00,51){\makebox(0,0)[bl]{\(a_{2}\)}}
\put(09,50){\line(1,0){06}} \put(09,54){\line(1,0){06}}
\put(09,50){\line(0,1){04}} \put(15,50){\line(0,1){04}}

\put(31,50.5){\line(0,1){03}} \put(41,50.5){\line(0,1){03}}
\put(31,52){\line(1,0){10}}

\put(00,46){\makebox(0,0)[bl]{\(a_{3}\)}}
\put(09,45){\line(1,0){06}} \put(09,49){\line(1,0){06}}
\put(09,45){\line(0,1){04}} \put(15,45){\line(0,1){04}}

\put(37,45.5){\line(0,1){03}} \put(49,45.5){\line(0,1){03}}
\put(37,47){\line(1,0){12}}

\put(00,41){\makebox(0,0)[bl]{\(a_{4}\)}}
\put(09,40){\line(1,0){09}} \put(09,44){\line(1,0){09}}
\put(09,40){\line(0,1){04}} \put(18,40){\line(0,1){04}}

\put(43,40.5){\line(0,1){03}} \put(53,40.5){\line(0,1){03}}
\put(43,42){\line(1,0){10}}

\put(00,36){\makebox(0,0)[bl]{\(a_{5}\)}}
\put(09,35){\line(1,0){07}} \put(09,39){\line(1,0){07}}
\put(09,35){\line(0,1){04}} \put(16,35){\line(0,1){04}}

\put(52,35.5){\line(0,1){03}} \put(62,35.5){\line(0,1){03}}
\put(52,37){\line(1,0){10}}

\put(00,31){\makebox(0,0)[bl]{\(a_{6}\)}}
\put(09,30){\line(1,0){08}} \put(09,34){\line(1,0){08}}
\put(09,30){\line(0,1){04}} \put(17,30){\line(0,1){04}}

\put(60,30.5){\line(0,1){03}} \put(70,30.5){\line(0,1){03}}
\put(60,32){\line(1,0){10}}

\put(00,26){\makebox(0,0)[bl]{\(a_{7}\)}}
\put(09,25){\line(1,0){07}} \put(09,29){\line(1,0){07}}
\put(09,25){\line(0,1){04}} \put(16,25){\line(0,1){04}}

\put(65,25.5){\line(0,1){03}} \put(75,25.5){\line(0,1){03}}
\put(65,27){\line(1,0){10}}


\put(25,10){\vector(1,0){55}}

\put(80,08){\makebox(0,0)[bl]{\(t\)}}

\put(25,8.5){\line(0,1){03}} 
\put(35,8.5){\line(0,1){03}} 
\put(45,8.5){\line(0,1){03}} 
\put(55,8.5){\line(0,1){03}} 
\put(65,8.5){\line(0,1){03}} 
\put(75,8.5){\line(0,1){03}}

\put(24,5){\makebox(0,0)[bl]{\(0\)}}
\put(33,5){\makebox(0,0)[bl]{\(1.0\)}}
\put(43,5){\makebox(0,0)[bl]{\(2.0\)}}
\put(53,5){\makebox(0,0)[bl]{\(3.0\)}}
\put(63,5){\makebox(0,0)[bl]{\(4.0\)}}
\put(73,5){\makebox(0,0)[bl]{\(5.0\)}}

\put(00,19){\makebox(0,0)[bl]{One-machine}}
\put(00,15){\makebox(0,0)[bl]{scheduling}}

\put(25,15){\line(0,1){06}} \put(75,15){\line(0,1){06}}

\put(25,16){\line(1,0){50}} \put(25,20){\line(1,0){50}}

\put(26,17){\makebox(0,0)[bl]{\(a_{1}\)}}
\put(25,16){\line(0,1){04}} \put(30,16){\line(0,1){04}}

\put(30.33,16){\line(0,1){04}} \put(30.66,16){\line(0,1){04}}

\put(32.5,17){\makebox(0,0)[bl]{\(a_{2}\)}}
\put(31,16){\line(0,1){04}} \put(37,16){\line(0,1){04}}

\put(38,17){\makebox(0,0)[bl]{\(a_{3}\)}}
\put(37,16){\line(0,1){04}} \put(43,16){\line(0,1){04}}

\put(46,17){\makebox(0,0)[bl]{\(a_{4}\)}}
\put(43,16){\line(0,1){04}} \put(52,16){\line(0,1){04}}

\put(54,17){\makebox(0,0)[bl]{\(a_{5}\)}}
\put(52,16){\line(0,1){04}} \put(59,16){\line(0,1){04}}

\put(59.33,16){\line(0,1){04}} \put(59.66,16){\line(0,1){04}}

\put(62,17){\makebox(0,0)[bl]{\(a_{6}\)}}
\put(60,16){\line(0,1){04}} \put(68,16){\line(0,1){04}}

\put(70,17){\makebox(0,0)[bl]{\(a_{7}\)}}
\put(68,16){\line(0,1){04}}

\end{picture}
\end{center}

 It is reasonable to point out (separately) the basic objective functions
 which are usually used in JIT-like scheduling problems:

  {\it 1.} makespan (i.e., total completion time of all jobs ) maximization
  in scheduling
  (e.g., \cite{deren10}),

 {\it 2.} maximizing the number of JIT jobs in scheduling
 (e.g., \cite{choi07,shab12b}),

\newpage
\begin{center}
\begin{small}
 {\bf Table 2.} JIT scheduling models\\
%
\begin{tabular}{| c | l| l |}
\hline
 No.&Model(s)& Source(s)\\
%

\hline

 1.&Maximizing the  number of JIT jobs:&\\

 1.1.&Scheduling to maximize the number of JIT  jobs  (survey)
   & \cite{shab12b}\\

 1.2.&Parallel machine scheduling to maximize the weighted number
 of JIT jobs
   & \cite{adam10,sung05}\\





 1.3.&Scheduling of parallel identical machines to maximize the
 weighted number
   &\cite{hir02}\\

  &of JIT jobs&\\


 1.4.&Maximizing the number of JIT jobs in flow-shop scheduling
   &\cite{choi07}\\


\hline

 2.&JIT scheduling with equal-size (unit processing) jobs:&\\

 2.1.&JIT scheduling with equal-size jobs (survey)
   & \cite{souk12}\\


 2.2.&Due dates assignment and JIT scheduling with equal-size jobs
  &\cite{tuo10}\\


 2.3.& Minmax weighted earliness-tardiness with identical processing
 times
   &\cite{ger17}\\

 &(and two competing agents)&\\


\hline
 3.&JIT scheduling in flow shop scheduling systems:&\\

 3.1.&JIT scheduling problem in flow shop scheduling systems
   &\cite{shab12}\\


 3.2.&Maximizing the number of just-in-time jobs in flow-shop scheduling
   &\cite{choi07}\\


 3.3.&Parameterized tractability of the just-in-time flow-shop
 scheduling
   &\cite{herm17}\\


\hline
 4.&Multicriteria JIT scheduling problems:&\\

 4.1.&Multicriteria JIT scheduling problems&\cite{tki06}\\

 4.2.&Multi-criteria scheduling in JIT approach&\cite{bol10}\\


 4.3.&Multicriteria earliness-tardiness scheduling
  &\cite{hoog05}\\

 4.4.&Bicriterion approach to common flow allowances due window
  scheduling
   &\cite{wangd17,yin16}\\

  &with controllable processing times&\\



 4.5.&Multi-criteria scheduling with due-window assignment problem
  &\cite{mos08a}\\


\hline
 5.&Some basic JIT scheduling problems under uncertainty:&\\

 5.1.&Scheduling problem with uncertain parameters in just in time
 system
 &\cite{boz14}\\


 5.2.&Stochastic scheduling with minimizing the number of tardy jobs
  &\cite{ely13}\\


 5.3.&Minimizing the number of tardy jobs with stochastically-ordered
   &\cite{tri08}\\

  & processing times&\\


 5.4.&Stochastic single machine scheduling with quadratic early-tardy
 penalties
 & \cite{mitt93} \\


 5.5.&Minimization of the weighted number of tardy jobs
   &\cite{de91}\\

 &with random processing times and deadline&\\


 5.6.&Scheduling stochastic jobs with asymmetric earliness and
 tardiness penalties
   &\cite{cai97}\\


 5.7.&Single machine stochastic scheduling to minimize the expected
 number
   &\cite{seo05}\\

 &of tardy  jobs  using mathematical programming models&\\


 5.8.&Stochastic flow-shop scheduling with minimizing the expected
 number
   & \cite{ely13b}\\

 &of tardy jobs&\\


 5.9.&Stochastic scheduling on parallel machines subject to random
 breakdowns
  & \cite{cai99}\\

 &to minimize expected costs for earliness and tardy jobs &\\


 5.10.&Single machine stochastic JIT scheduling problem
 & \cite{tang08}\\

 &subject to machine breakdowns&\\


 5.11.&Fixed interval scheduling under uncertainty&\cite{branda16}\\



\hline
 6.&Some special models:&\\

 6.1.&Makespan minimization of multi-slot JIT scheduling on
 single
   & \cite{deren10}\\

 &\&  parallel machines&\\


 6.2.&Mixed-model, multi level JIT processes, assembly systems
   &\cite{boy11,kub97,milt89,ste93}\\




 6.3.&Maximum deviation JIT scheduling problem&\cite{bra04,ste93}\\

 6.4.&Online JIT scheduling framework as weighted bipartite matching
 &\cite{mcg08,pin16}\\

 6.5.&Cyclic scheduling for
  production system
  under JIT delivery policy
     &\cite{nori96}\\


 6.6.&Two-agent single-machine scheduling problem with JIT jobs
   & \cite{choi14}\\

 6.7.&JIT scheduling with controllable processing times on
 parallel machines
    &\cite{ley10}\\


 6.8.&JIT scheduling with competing agents&\cite{chung12}\\

 6.9.&Single machine weighted mean squared deviation problem
  & \cite{per17} \\


\hline
\end{tabular}
\end{small}
\end{center}

\newpage
\begin{center}
\begin{small}
 {\bf Table 3.} Solving (algorithmic) approaches \\
\begin{tabular}{| c | l| l |}
\hline
 No.&Solving approach & Source(s)\\

\hline
 1.&General solving  approaches:&\\

 1.1.&Sequencing approaches for mixed-model JIT production systems (review)
   &\cite{dha09}\\


 1.2.&Decomposition method for multi-product kanban systems
 with setup times
   &\cite{kri02}\\

 &and lost sales&\\


 1.3.&Decomposition algorithms for parallel machine JIT scheduling
  &\cite{chen99}\\


 1.4.&Dynamic programming algorithms (e.g., for
   mixed-model, JIT production systems)
    &\cite{milt90,tan12}\\


 1.5.&Unifying approach by goal programming approach for JIT manufacturing
     &\cite{li06}\\


 1.6.&Simulation analysis of  JIT techniques for  production systems
     &\cite{huang83}\\


 \hline
 2.&Constrained programming approaches:&\\
 2.1.&Constrained programming approach for JIT scheduling
    &\cite{mon09}\\


 2.2.&Change constrained programming for stochastic scheduling
  &\cite{ely13}\\

 & with minimizing the number of tardy jobs&\\


\hline
 3.&Some polynomial methods:&\\

 3.1.& Polynomial cases and PTAS for JIT  scheduling on
 parallel machines
   & \cite{tuo08b}\\

  &around a common due-date&\\


 3.2.&Some new polynomial cases in JIT scheduling problems
 with multiple due dates
   & \cite{tuo08} \\


 3.3.&Quadratic time algorithm to maximize the number of JIT jobs
   &\cite{cep05b}\\

 &on identical parallel machines&\\


 3.4.&Polynomial algorithm  for minimization of earliness, tardiness and
   &\cite{dro12}\\

 &due date penalties on uniform parallel machines with identical jobs&\\


 3.5.&FPTAS for the weighted earliness-tardiness problem&\cite{kov99}\\

 3.6.&FPTAS for for the total weighted tardiness with a common due date
    &\cite{kacem10}\\


 3.7.&Fast FPTAS for
  minimization of  total weighted earliness and tardiness
   & \cite{kel18}\\

 &(single machine scheduling, about large common du date)&\\


\hline
 4.&Enumerative techniques
  (e.g., branch-and-bound (B\&B), dynamic programming (DP)):&\\
 4.1.&
 B\&B algorithm for
   single machine earliness and tardiness scheduling
     &\cite{ronc10}\\


 4.2.&
 B\&B algorithm for minimizing earliness\&tardiness costs
   &\cite{bak14}\\

 &in single-machine stochastic scheduling &\\


 4.3.&
 B\&B algorithm for the single machine sequence-dependent group
   &\cite{kes15}\\

 &scheduling with earliness and tardiness penalties &\\


 4.4.&Faster
 B\&B algorithm for the
 earliness–tardiness scheduling problem
   &\cite{sourd08}\\


 4.5.&Branch-and-cut algorithm for single machine JIT scheduling
 to minimize the sum of
  &\cite{per17}\\

 &weighted  mean squared deviation of completion times&\\

 & with respect  a common due date&\\


 4.6.&
 DP algorithm for scheduling mixed-model,
 JIT production systems
    &\cite{milt90}\\


\hline
 5.&Heuristic (macroheuristic) approaches (including evolutionary methods):&\\

 5.1.&Heuristics for JIT scheduling in parallel machines&\cite{lag91}\\

 5.2.& GA for single machine
 scheduling with linear earliness\&
 quadratic tardiness penalties
   & \cite{vale09}\\


 5.3.&
 Hybrid Genetic Bees Algorithm applied to single machine scheduling
  &\cite{yuc17}\\

 & with earliness and tardiness penalties&\\


 5.4.&Hybrid GA-SA algorithm for JIT scheduling of multi-level assemblies
  &\cite{roa96}\\


 5.5.&Tabu search for
 JIT sequencing for mixed-model assembly lines with setups
   &\cite{mcm98}\\


 5.6.&Tabu search in JIT scheduling problem with uncertain parameters
 &\cite{boz14}\\


 5.7.&Tabu search for fixed interval scheduling under uncertainty
   & \cite{branda16}\\

 5.8.&Local search metaheuristic
  for the JIT scheduling problem
   & \cite{chet15}\\

 5.9.& Metaheuristics for scheduling on parallel machine
  &\cite{adam12b}\\

 &to minimize weighted number of early and tardy jobs&\\


 5.10.&Metaheuristics for multi-criteria scheduling with JIT approach
     (genetic
  &\cite{bol10}\\

 &algorithm (GA), particle swam optimization (PSO),
 differential evolution (DE))&\\


 5.11.&Hybrid metaheuristics for scheduling to
    minimize  weighted earliness-tardiness
    &\cite{alv15}\\

 &penalties on parallel identical machines&\\



\hline
\end{tabular}
\end{small}
\end{center}

  {\it 3.} maximization of the weighted number of JIT jobs in scheduling
   (e.g., \cite{adam10,hir02,sung05}),

 {\it 4.} minimizing the number of tardy jobs in scheduling (e.g., \cite{ely13}),

 {\it 5.} minimization of weighted earliness and tardiness penalties in
 scheduling
  (e.g., \cite{alv15,jana08}),


\newpage
\begin{center}
\begin{small}
 {\bf Table 4.} Some JIT
 application domains\\
\begin{tabular}{| c | l| l |}
\hline
 No.&Application domain& Source(s)\\

\hline

 1.& JIT scheduling for production (manufacturing) systems
 &\cite{joze10,milt90,mond11}\\

 2.&JIT schedules for flexible transfer lines&\cite{kub94}\\

 3.& JIT scheduling for register optimization (computer design)
 &\cite{van92}\\


 4.& JIT scheduling for multichannel
  Ethernet passive optical networks  (EPONs)
  &\cite{mcg08,pin16}\\

 & (online JIT scheduling framework as weighted bipartite matching)&\\


 5.&JIT scheduling for real-time sensor data dissemination
     &\cite{liu06}\\


 6.&JIT signaling for WDM optical burst switching networks
    &\cite{wei00}\\

 7.&Scheduling with JIT approach in transportation
  &\cite{alv09,bol10}\\



 8.&JIT management of  building projects&\cite{pheng12}\\


 9.&
 Computer wiring \&
  bandwidth allocation of communication channels
    &\cite{bar15,chenb02,gup79,kov07}\\






 10.&VLSI circuit design&\cite{carl95}\\


 11.&Timetabling (for bus drivers, for aircraft services, for class scheduling,
   &\cite{cart92,kroon97,mar86}\\

 & for satellite data transmission, etc.)&\\


 12.&Identification of protein-encoding genes by
  spectroscopical methods
   & \cite{chenzz03}\\


 13.&Scheduling a maintenance activity
    & \cite{kroon97,mos09bb}\\






\hline
\end{tabular}
\end{small}
\end{center}

\begin{center}
\begin{small}
 {\bf Table 5.} Date for examples of interval scheduling \\
\begin{tabular}{| c | c| c |c |c| }
\hline
 No.&Job&Processing&Processing interval    & Position in \\
  &\( a_{i}\)&time \(\theta(a_{i})\)& \(\lambda^{a_{i}}=[t^{a_{i}}_{1},t^{a_{i}}_{2}]\)&schedule\\

\hline
 1.&\(a_{1}\) &\(0.5\) &  \([ 0.0,1.1 ]\)  &\(1\)\\
 2.&\(a_{2}\) &\(0.6\) &  \([ 0.6,1.6 ]\)  &\(2\)\\
 3.&\(a_{3}\) &\(0.6\) &  \([ 1.2,2.4 ]\)  &\(3\)\\
 4.&\(a_{4}\) &\(0.9\) &  \([ 1.8,2.8 ]\)  &\(4\)\\
 5.&\(a_{5}\) &\(0.7\) &  \([ 2.7,3.7 ]\)  &\(5\)\\
 6.&\(a_{6}\) &\(0.8\) &  \([ 3.5,4.5 ]\)  &\(6\)\\
 7.&\(a_{7}\) &\(0.7\) &  \([ 4.0,5.0 ]\)  &\(7\)\\

\hline
\end{tabular}
\end{small}
\end{center}

%
\begin{center}
\begin{picture}(90,100)
\put(03,00){\makebox(0,0)[bl]{Fig. 3.
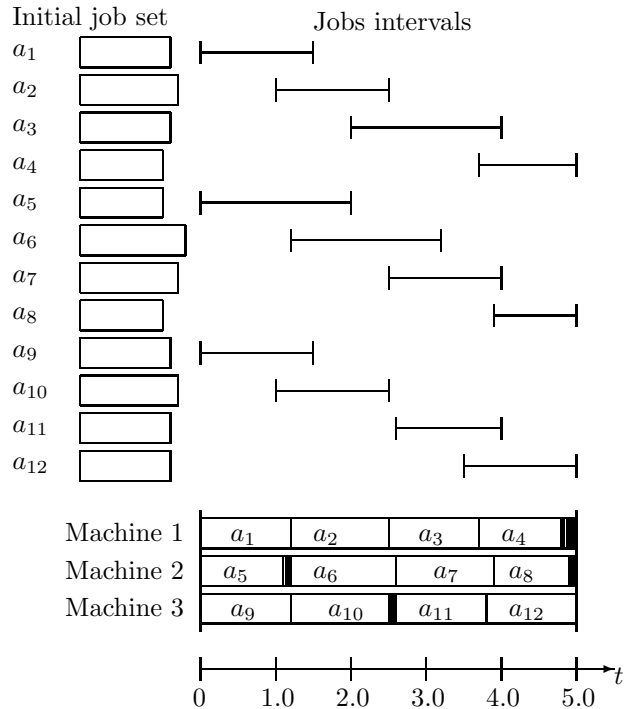
 Illustration for three-machine scheduling }}

\put(00,95){\makebox(0,0)[bl]{Initial job set}}

\put(040,95){\makebox(0,0)[bl]{Jobs intervals}}

\put(00,91){\makebox(0,0)[bl]{\(a_{1}\)}}
\put(09,90){\line(1,0){12}} \put(09,94){\line(1,0){12}}
\put(09,90){\line(0,1){04}} \put(21,90){\line(0,1){04}}

\put(25,90.5){\line(0,1){03}} \put(40,90.5){\line(0,1){03}}
\put(25,92){\line(1,0){15}}

\put(00,86){\makebox(0,0)[bl]{\(a_{2}\)}}
\put(09,85){\line(1,0){13}} \put(09,89){\line(1,0){13}}
\put(09,85){\line(0,1){04}} \put(22,85){\line(0,1){04}}

\put(35,85.5){\line(0,1){03}} \put(50,85.5){\line(0,1){03}}
\put(35,87){\line(1,0){15}}

\put(00,81){\makebox(0,0)[bl]{\(a_{3}\)}}
\put(09,80){\line(1,0){12}} \put(09,84){\line(1,0){12}}
\put(09,80){\line(0,1){04}} \put(21,80){\line(0,1){04}}

\put(45,80.5){\line(0,1){03}} \put(65,80.5){\line(0,1){03}}
\put(45,82){\line(1,0){20}}

\put(00,76){\makebox(0,0)[bl]{\(a_{4}\)}}
\put(09,75){\line(1,0){11}} \put(09,79){\line(1,0){11}}
\put(09,75){\line(0,1){04}} \put(20,75){\line(0,1){04}}

\put(62,75.5){\line(0,1){03}} \put(75,75.5){\line(0,1){03}}
\put(62,77){\line(1,0){13}}

\put(00,71){\makebox(0,0)[bl]{\(a_{5}\)}}
\put(09,70){\line(1,0){11}} \put(09,74){\line(1,0){11}}
\put(09,70){\line(0,1){04}} \put(20,70){\line(0,1){04}}

\put(25,70.5){\line(0,1){03}} \put(45,70.5){\line(0,1){03}}
\put(25,72){\line(1,0){20}}

\put(00,66){\makebox(0,0)[bl]{\(a_{6}\)}}
\put(09,65){\line(1,0){14}} \put(09,69){\line(1,0){14}}
\put(09,65){\line(0,1){04}} \put(23,65){\line(0,1){04}}

\put(37,65.5){\line(0,1){03}} \put(57,65.5){\line(0,1){03}}
\put(37,67){\line(1,0){20}}

\put(00,61){\makebox(0,0)[bl]{\(a_{7}\)}}
\put(09,60){\line(1,0){13}} \put(09,64){\line(1,0){13}}
\put(09,60){\line(0,1){04}} \put(22,60){\line(0,1){04}}

\put(50,60.5){\line(0,1){03}} \put(65,60.5){\line(0,1){03}}
\put(50,62){\line(1,0){15}}

\put(00,56){\makebox(0,0)[bl]{\(a_{8}\)}}
\put(09,55){\line(1,0){11}} \put(09,59){\line(1,0){11}}
\put(09,55){\line(0,1){04}} \put(20,55){\line(0,1){04}}

\put(64,55.5){\line(0,1){03}} \put(75,55.5){\line(0,1){03}}
\put(64,57){\line(1,0){11}}

\put(00,51){\makebox(0,0)[bl]{\(a_{9}\)}}
\put(09,50){\line(1,0){12}} \put(09,54){\line(1,0){12}}
\put(09,50){\line(0,1){04}} \put(21,50){\line(0,1){04}}

\put(25,50.5){\line(0,1){03}} \put(40,50.5){\line(0,1){03}}
\put(25,52){\line(1,0){15}}

\put(00,46){\makebox(0,0)[bl]{\(a_{10}\)}}
\put(09,45){\line(1,0){13}} \put(09,49){\line(1,0){13}}
\put(09,45){\line(0,1){04}} \put(22,45){\line(0,1){04}}

\put(35,45.5){\line(0,1){03}} \put(50,45.5){\line(0,1){03}}
\put(35,47){\line(1,0){15}}

\put(00,41){\makebox(0,0)[bl]{\(a_{11}\)}}
\put(09,40){\line(1,0){12}} \put(09,44){\line(1,0){12}}
\put(09,40){\line(0,1){04}} \put(21,40){\line(0,1){04}}

\put(51,40.5){\line(0,1){03}} \put(65,40.5){\line(0,1){03}}
\put(51,42){\line(1,0){14}}

\put(00,36){\makebox(0,0)[bl]{\(a_{12}\)}}
\put(09,35){\line(1,0){12}} \put(09,39){\line(1,0){12}}
\put(09,35){\line(0,1){04}} \put(21,35){\line(0,1){04}}

\put(60,35.5){\line(0,1){03}} \put(75,35.5){\line(0,1){03}}
\put(60,37){\line(1,0){15}}

\put(25,10){\vector(1,0){55}}

\put(80,08){\makebox(0,0)[bl]{\(t\)}}

\put(25,8.5){\line(0,1){03}} 
\put(35,8.5){\line(0,1){03}} 
\put(45,8.5){\line(0,1){03}} 
\put(55,8.5){\line(0,1){03}} 
\put(65,8.5){\line(0,1){03}} 
\put(75,8.5){\line(0,1){03}}

\put(24,5){\makebox(0,0)[bl]{\(0\)}}
\put(33,5){\makebox(0,0)[bl]{\(1.0\)}}
\put(43,5){\makebox(0,0)[bl]{\(2.0\)}}
\put(53,5){\makebox(0,0)[bl]{\(3.0\)}}
\put(63,5){\makebox(0,0)[bl]{\(4.0\)}}
\put(73,5){\makebox(0,0)[bl]{\(5.0\)}}


\put(07,27){\makebox(0,0)[bl]{Machine 1}}

\put(25,25){\line(0,1){06}} \put(75,25){\line(0,1){06}}
\put(25,26){\line(1,0){50}} \put(25,30){\line(1,0){50}}

\put(29,26.5){\makebox(0,0)[bl]{\(a_{1}\)}}
\put(25,26){\line(0,1){04}} \put(37,26){\line(0,1){04}}

\put(40,26.5){\makebox(0,0)[bl]{\(a_{2}\)}}
\put(37,26){\line(0,1){04}} \put(50,26){\line(0,1){04}}

\put(54,26.5){\makebox(0,0)[bl]{\(a_{3}\)}}
\put(50,26){\line(0,1){04}} \put(62,26){\line(0,1){04}}

\put(65,26.5){\makebox(0,0)[bl]{\(a_{4}\)}}
\put(62,26){\line(0,1){04}} \put(73,26){\line(0,1){04}}

\put(73.33,26){\line(0,1){04}} \put(73.66,26){\line(0,1){04}}
\put(74,26){\line(0,1){04}}
\put(74.33,26){\line(0,1){04}} \put(74.66,26){\line(0,1){04}}

\put(07,22){\makebox(0,0)[bl]{Machine 2}}

\put(25,20){\line(0,1){06}} \put(75,20){\line(0,1){06}}
\put(25,21){\line(1,0){50}} \put(25,25){\line(1,0){50}}

\put(28,21.5){\makebox(0,0)[bl]{\(a_{5}\)}}
\put(25,21){\line(0,1){04}} \put(36,21){\line(0,1){04}}

\put(36.33,21){\line(0,1){04}} \put(36.66,21){\line(0,1){04}}

\put(40,21.5){\makebox(0,0)[bl]{\(a_{6}\)}}
\put(37,21){\line(0,1){04}} \put(51,21){\line(0,1){04}}

\put(56,21.5){\makebox(0,0)[bl]{\(a_{7}\)}}
\put(51,21){\line(0,1){04}} \put(64,21){\line(0,1){04}}

\put(66,21.5){\makebox(0,0)[bl]{\(a_{8}\)}}
\put(64,21){\line(0,1){04}} \put(74,21){\line(0,1){04}}

\put(74.33,21){\line(0,1){04}} \put(74.66,21){\line(0,1){04}}


\put(07,17){\makebox(0,0)[bl]{Machine 3}}

\put(25,15){\line(0,1){06}} \put(75,15){\line(0,1){06}}
\put(25,16){\line(1,0){50}} \put(25,20){\line(1,0){50}}

\put(29,16.5){\makebox(0,0)[bl]{\(a_{9}\)}}
\put(25,16){\line(0,1){04}} \put(37,16){\line(0,1){04}}

\put(42,16.5){\makebox(0,0)[bl]{\(a_{10}\)}}
\put(37,16){\line(0,1){04}} \put(50,16){\line(0,1){04}}

\put(50.33,16){\line(0,1){04}} \put(50.66,16){\line(0,1){04}}

\put(54,16.5){\makebox(0,0)[bl]{\(a_{11}\)}}
\put(51,16){\line(0,1){04}} \put(63,16){\line(0,1){04}}

\put(66,16.5){\makebox(0,0)[bl]{\(a_{12}\)}}
\put(63,16){\line(0,1){04}} \put(75,16){\line(0,1){04}}



\end{picture}
\end{center}

\newpage
\begin{center}
\begin{small}
 {\bf Table 6.} Date for examples of interval scheduling \\
\begin{tabular}{| c | c| c |c |c| c|}
\hline
 No.&Job&Processing&Processing interval &Number of&Position in\\
 &\(a_{i}\)&time \(\theta(a_{i})\)&\(\lambda^{a_{i}} = [t^{a_{i}}_{1},t^{a_{i}}_{2} ]\)&machine&schedule\\

\hline
 1. &\(a_{1}\) &\(1.2\)& \([0.0,1.5]\) &\(1\) &\(1\)\\
 2. &\(a_{2}\) &\(1.3\)& \([1.0,2.5]\) &\(1\) &\(2\)\\
 3. &\(a_{3}\) &\(1.2\)& \([2.0,4.0]\) &\(1\) &\(3\)\\
 4. &\(a_{4}\) &\(1.1\)& \([3.7,5.0]\) &\(1\) &\(4\)\\

 5. &\(a_{5}\) &\(0.7\)& \([0.0,2.0]\) &\(2\) &\(1\)\\
 6. &\(a_{6}\) &\(0.6\)& \([1.7,2.7]\) &\(2\) &\(2\)\\
 7. &\(a_{7}\) &\(0.7\)& \([2.5,4.0]\) &\(2\) &\(3\)\\
 8. &\(a_{8}\) &\(1.0\)& \([3.9,5.0]\) &\(2\) &\(4\)\\

 9. &\(a_{9}\) &\(1.2\)& \([0.0,1.5 ]\) &\(3\) &\(1\)\\
 10.&\(a_{10}\)&\(1.3\)& \([1.0,2.5]\) &\(3\) &\(2\)\\
 11.&\(a_{11}\)&\(1.2\)& \([2.6,4.0]\) &\(3\) &\(3\)\\
 12.&\(a_{12}\)&\(1.2\)& \([3.0,5.0]\) &\(3\) &\(4\)\\

\hline
\end{tabular}
\end{small}
\end{center}

 {\it 6.} minimizing the weighted number of early and tardy jobs in scheduling
  (e.g., \cite{jana08}),


  {\it 7.} maximum deviation in JIT scheduling problems
  (e.g., \cite{bra04,per17,ste93}),

  {\it 8.} minimization of the sum of weighted mean squared deviation of the completion times
   (e.g., \cite{per17}),


  {\it 9.} minimizing variation of production rates in JIT systems
   (e.g., \cite{kub93})

  {\it 10.} minimization of the expected number of tardy jobs in scheduling
   (e.g., \cite{ely13b}),

  {\it 11.} minimization of  expected costs for earliness
  and tardy jobs in scheduling
  (e.g., \cite{cai99}).

 In recent years, multicriteria JIT-like scheduling problems
 are under examination (i.e., combination of the objective
 functions above are used)
 (e.g., \cite{bol10,hoog05,tki06,wangd17,yin16}).


 It may be reasonable to present an example of basic problem formulations as
 follows.
 Let jobs (tasks)
 \(A = \{a_{1},...,a_{i},...,a_{n} \}\)
 are non-preemptive and are scheduled on \(m\)
 identical machines (processors)
 \(P=\{P_{1},...,P_{j},...,P_{m}\}\).
 Each \(P_{j}\) can handle at most one job at a time,
 each \(a_{i}\) can be completely processed on any machine.
 Each job \(a_{i}\) has parameters:
 (a) processing time \(\theta (a_{i})\),
 (b) a time interval (i.e., window)
 for its processing
 \( \lambda^{a_{i}} =[ t^{a_{i}}_{1}, t^{a_{i}}_{2} ] \),
 (c) completion time in a solution \(S\) (i.e., a schedule)
  \(C(a_{i})\),
 (d) earliness
 \(u (a_{i}) = \max \{ 0, t^{a_{i}}_{1} - C (a_{i}) \} \),
 (e) tardiness
  \(v (a_{i}) = \max \{ 0, C(a_{i}) - t^{a_{i}}_{2}  \} \).
 In addition, values of non-negative costs (penalties)
 of earliness and tardiness are given:
 \(\alpha~u(a_{i}) \) and
 \(\beta~v(a_{i}) \),
 respectively.
 The problem is (for sum of penalties):

~~

   Find a schedule \(S\) such that
   \( F^{sum}(S)=\sum_{i=1}^{n} ~[\alpha~ u(a_{i}) + \beta ~v(a_{i}) ]\)
 is minimized.

~~

  In the case of maximum total penalty, the objective function is:

  to  minimize
   \( F^{max}(S)= \max_{1\leq i \leq n} ~ \{ \alpha ~u(a_{i}),\beta ~v(a_{i}) \} \)

~~

 In addition, it is necessary to point out combinatorial
 optimization models which are close to
 (or are used in)
  JIT-like scheduling problems:
  \(k\)-coloring, maximum weight clique, assignment/allocation,
  timetabling,
  maximum weight independent set, weighted matching
 (e.g., \cite{kov07,mcg08,pin16}).


%
 Some simplified special JIT scheduling/planning models can be solvable by
 polynomial algorithms
 (e.g.,  \cite{cep05b,jana15,mos10b,tuo08,tuo08b,valen03,wangx10})
%
%
 or PTAS/FPTAS
 (e.g., \cite{kacem10,kel18,kov99,tuo08b}).
 In the main, the models are NP-hard
 (e.g., \cite{gar79,gar88,hoog05,jana15,kes15,papa98,tki06}
  and,
%
 as a result, the following solving approaches are used:
 (i) enumerative methods (e.g., branch-and-bound and dynamic programming algorithms
 (e.g., \cite{bak14,kes15,milt90}),
%
%
 (ii)  various heuristics/metaheristics
 (local optimization, VNS methods, evolutionary algorithms, etc.)
 (e.g., \cite{lag91,roa96}).
%
 Analogical situation (i.e., usage of enumerative algorithms and/or heuristics)
  exists in the filed of  multicriteria  JIT scheduling problems
 (e.g.,  \cite{bol10,tki06}).


%
 In recent decades,
 due-data (due-window) assignment scheduling problems
 have been intensively studied (Table 7, Table 8).
 Here, it is reasonable to point out the basic surveys
 \cite{bak90,gord02,gord04,jana15,kam04,lauf04}.
%
%
%
%
 The situation with problem complexities and the used
 solving methods is analogical
 (as it was pointed out for JIT scheduling).

\newpage

\begin{center}
\begin{small}
 {\bf Table 7.} Some studies in scheduling with
  due-date assignment, part 1 \\
\begin{tabular}{| c | l| l |}
\hline
 No.&Issue(s)& Source(s)\\

\hline

 1.&Surveys:&\\

 1.1.&Due date quotation models and algorithms&\cite{kam04}\\

 1.2.&Survey of the state-of-the-art of common
 due-date assignment and scheduling
 & \cite{gord02} \\


 1.3.&Scheduling with due date assignment (survey)&\cite{gord04}\\

 1.4.&Scheduling with common due date, earliness and tardiness
 penalties
  &\cite{lauf04}\\

 & for multiple machine problems&\\




\hline
 2.&Due dates assignment and JIT scheduling with unit-time or equal-size
 jobs:&\\

 2.1.&Minmax earliness-tardiness costs with unit processing time jobs
   &\cite{mos01}\\


 2.2.&Minimizing weighted earliness–tardiness and
 due date cost with unit–time jobs
 &\cite{mos06a}\\


 2.3.&Scheduling of unit-time jobs distinct due windows on parallel
  processors
    &\cite{jana08}\\


 2.4.&Due dates assignment and JIT scheduling with equal-size jobs
  &\cite{tuo10}\\


 2.5.& Minimizing earliness,
 tardiness, and due-date costs for equal-sized jobs
 & \cite{lic08} \\


 2.6.&Minimization of earliness, tardiness and due date penalties
   &\cite{dro12}\\

 &on uniform parallel machines with identical jobs&\\


\hline
 3.& Single machine due-date assignment problems:&\\

 3.1.& Assigning a common due-date for all the jobs (CON),
  &\cite{bak90,gord02,shab12c}\\


 3.2.&Assigning job-dependent due-dates which are (linear)
 functions
   &\cite{gord92d,gord99a,jana07}  \\

 & of the job processing times  (SLK)
    &\cite{kara93,wangx10}  \\

  3.3.&Assigning job-dependent due-dates which are penalized
   & \cite{seid81,shab08,shab10}\\

  &if exceed prespecified deadlines (DIF) & \\


  3.4.&Assigning generalized due dates (GDD)
    &\cite{hall91b,mosh04d,sris90a}\\


 3.5.&Minmax due-date assignment problem with lead-time cost
   &\cite{morb13}\\


 3.6.&Common due date assignment to minimize total penalty
  &\cite{panw82}\\

 & for the one machine scheduling problem&\\


 3.7.&Single-machine scheduling problems involving due date
 determination decisions
   & \cite{yin13}\\


 3.8.&Stochastic single machine scheduling with proportional job
 weights
 &\cite{jia03}\\

 &to minimize deviations
 of completion times from a common due date&\\


\hline
 4.&Common due-date assignment and scheduling:&\\

 4.1.&Survey of the state-of-the-art of common
 due-date assignment and scheduling
   &\cite{gord02}\\


 4.2.& Assigning a common due-date for all the jobs (CON),
  &\cite{bak90,gord02,shab12c}\\


 4.3.&Scheduling to a common due-date on parallel uniform processors
   &\cite{emm87}\\


 4.4.&Common due date assignment to minimize total penalty
 for  one machine scheduling
  &\cite{panw82}\\



 4.5.&Scheduling around a small common due date&\cite{ali97,hoog91}\\




 4.6.&Common due date assignment with generalized eaqrliness/tardness
 penalties
   & \cite{koul17} \\


 4.7.&
 Single machine scheduling
 with different ready times and common due date
    & \cite{birg12}\\


 4.8.&Single machine weighted earliness-tardiness penalty problem
  & \cite{modal01}\\

 & with a common due date&\\


  4.9.& Unrelated parallel machine scheduling with a common due date,
   &\cite{bank01}\\

 &release dates, and linear earliness and tardiness penalties&\\


 4.10.&Common due date assignment and scheduling with ready times
    & \cite{cheng02}\\


 4.11.&Two-machine flow shop scheduling for the minimization
  &\cite{sak09}\\

 &of the mean absolute deviation from a common due date&\\


 4.12.&Scheduling on parallel identical machines to
    minimize  weighted
    &\cite{alv15}\\

 & earliness-tardiness penalties
     with respect to a common due date&\\


\hline
 5.&Stochastic/uncertaint scheduling with due dates:&\\

 5.1.&Sequencing stochastic jobs on a single machine with a
 common due date
  & \cite{sar91}\\

 &and stochastic processing times&\\


 5.2.&Single machine scheduling with stochastic processing times or
  & \cite{soro06}\\

 &stochastic due-dates to minimize the number of early and tardy
 jobs&\\


 5.3.& Stochastic single machine scheduling to minimize the weighted
 number
 &\cite{soro07}\\

 &of early and tardy jobs
 (random processing time, due dates)&\\


 5.4.&Stochastic single machine scheduling with random common due date
   & \cite{benm12}\\


 5.5.&Stochastic single machine scheduling with proportional job
 weights
 &\cite{jia03}\\

 &to minimize deviations
 of completion times from a common due date&\\


 5.6.&Stochastic scheduling with release dates and due dates
    &\cite{pined83,pin16}\\



 5.7.&Scheduling stochastic jobs with due dates on parallel machines
   &\cite{emm90}\\


 5.8.&Fuzzy due-date scheduling problem with fuzzy processing times
 &\cite{itoh99}\\



\hline
\end{tabular}
\end{small}
\end{center}

\newpage
\begin{center}
\begin{small}
 {\bf Table 7.} Some studies in scheduling with
  due-date assignment, part 2 \\
\begin{tabular}{| c | l| l |}
\hline
 No.&Issue(s)& Source(s)\\

\hline
 6.&Special problems:&\\
 6.1.&Group scheduling and due date assignment&\cite{lis11}\\

 6.2.&Due-date assignment on uniform machines&\cite{mos09}\\

 6.3.&Optimal coordination of resource allocation,
 due date assignment in scheduling
   & \cite{shab16}\\


 6.4.&Single machine multiple common due dates scheduling
 with learning effects
  &\cite{wang10b}\\


 6.5.&Multiple common due dates assignment and scheduling
 problems with resource
  &\cite{yang13}\\

 &allocation
 and general position-dependent deterioration effect&\\


 6.6.&Dynamic due-date assignment models in flexible
  manufacturing systems
   &\cite{jos11} \\


\hline
\end{tabular}
\end{small}
\end{center}

\begin{center}
\begin{small}
 {\bf Table 8.} Some studies in scheduling with due-window assignment \\
\begin{tabular}{| c | l| l |}
\hline
 No.&Issue(s)& Source(s)\\

\hline

 1.&Surveys:&\\

 1.1.&Survey on scheduling problems with due windows&\cite{jana15}\\

 1.2.&Common due-window scheduling problems&\cite{jana13a,kram93}\\



 1.3.& Due-window scheduling for parallel machines&\cite{kram94}\\

 1.4.& Minmax scheduling problems with a common due-window
  & \cite{mos09b} \\



\hline
 2.&Single-machine scheduling with a common due window:&\\

 2.1.&Single-machine scheduling with a common due window assignment
  &\cite{jana04,mos10b,yeu01}\\



 2.2.&Determination of common due window location in
  single machine scheduling
   & \cite{lim96,lim98}\\



 2.3.&Single machine scheduling problem with common due window
   &\cite{lim97}\\

 &and controllable processing times&\\


 2.4.&Single-machine due window assignment and scheduling  with
  &\cite{yin14}\\

 &a common allowance and controllable job processing time&\\


\hline
 3.&Due-window assignment problems with unit-time (or equal) jobs:&\\

 3.1.&Due-window assignment problems with unit-time jobs&\cite{ger13,ger13c}\\



 3.2.&Due-window assignment with identical jobs
 on parallel uniform machines
   &\cite{ger13b}\\


 3.3.&Due-window assignment with unit processing time jobs&\cite{mos04}\\

 3.4.&Soft due-window assignment and scheduling
 of unit-time jobs
   &\cite{jana12}\\

  &on parallel machines&\\



\hline
 4.&Scheduling with a common due window:&\\

 4.1.&Common due-window scheduling&\cite{kram93}\\

 4.2.& Minmax scheduling problems with a common due-window&\cite{mos09b}\\

 4.3.&Two-stage flow shop earliness and tardiness machine
 scheduling
 &\cite{yeu04}\\

 &involving a common due window&\\


 4.4.&Two-machine flow shop scheduling with a common due window
 &\cite{yeu09}\\

 &to minimize weighted number of early and tardy jobs&\\


 4.5.&Parallel machine scheduling with common due-windows
   &\cite{huang10}\\


 4.6.&
 Parallel machine scheduling and common due window assignment
  &\cite{jana13b}\\

 & with job independent earliness and tardiness costs&\\


\hline
 5.&Special problems:&\\

 5.1.&Due-window assignment problem with position-dependent
 processing times
   &\cite{mos08}\\


 5.2.& Due-window scheduling for parallel machines&\cite{kram94}\\

 5.3.&Flexible job shop scheduling with due window&\cite{huang13b}\\

 5.4.&Due window assignment and resource allocation scheduling problems
   &\cite{liu17}\\

 &with learning and general positional effects&\\


 5.5.&Scheduling a maintenance activity and due-window assignment
   &\cite{morb15}\\


 5.6.& Scheduling problems with multiple due windows assignment
   &\cite{yang14}\\


 5.7.&Dynamic window-constrained scheduling of real-time streams in
 media servers
   &\cite{west04}\\


 5.8.&Multi-criteria scheduling with due-window assignment problem
  &\cite{mos08a}\\


\hline
 6.&Soft due-window assignment and scheduling:&\\

 6.1.&Soft due-window assignment and scheduling on parallel machines
   &\cite{jana07}\\


 6.2.&Soft due-window assignment and scheduling
 of unit-time jobs
   &\cite{jana12}\\

 & on parallel machines&\\


\hline
\end{tabular}
\end{small}
\end{center}

\newpage
\section{Time-interval balancing scheduling of modular jobs}

 It is assumed a three-stage system:
 production stage (manufacturing of a basic module set),
 transportation (transmission) stage, and utilization (e.g., assembly) stage.
 The problem consists in scheduling of composite jobs
 at  the third stage
 while taking into account balance constraints of
 the first stage and the second stage.
 Note, the main goal consists in deletion of  a buffer-based subsystem
 (as in JIT approaches)
 or to use a very simplified buffer-based subsystem.

 First, an illustration example is described (Fig. 4).
 There is a set of five basic elements (details, building elements, elementary jobs, modules):
 ~\( \overline{\Delta} = \{\Delta_{1},\Delta_{2},\Delta_{3},\Delta_{4},\Delta_{5}\} \).
 The elements are produced by corresponding five manufacturing organizations (e.g., producers, conveyors).
 The system is targeted to use (e.g., to assembly)
 of a specified set of four composite modular jobs
 (as module chains or typical modular building):
 \(a_{1}=<\Delta_{1} \rightarrow \Delta_{2} \rightarrow \Delta_{3}>\),
 \(a_{2}=<\Delta_{2} \rightarrow \Delta_{5}>\),
 \(a_{3}=<\Delta_{1} \rightarrow \Delta_{2} \rightarrow \Delta_{4} \rightarrow \Delta_{5}>\),
%
 \(a_{4} = <\Delta_{1} \rightarrow \Delta_{2} \rightarrow \Delta_{2}
 \rightarrow \Delta_{3} \rightarrow \Delta_{4} \rightarrow \Delta_{5}>\).

 In the example, the system assembly process is executed by three
 processors/teams (Fig. 4).
  The system plan (general schedule)
  consists of schedules for three processors (machines, teams)
 \(S = \{S_{1},S_{2},S_{3}\}\), where
 \(a_{0}\) denotes an ``empty'' element/time period):
%
 (i) team (processor) \(P_{1}\):
  \(S_{1} =< a_{4} \rightarrow a_{4}>\),
 (ii) team (processor) \(P_{2}\):
 \(S_{2} =<a_{2} \rightarrow a_{0} \rightarrow a_{3} \rightarrow a_{0}
 \rightarrow a_{3}>\),
 and
 (iii) team (processor) \(P_{3}\):
 \(S_{3} =<a_{3} \rightarrow a_{1} \rightarrow a_{3} \rightarrow a_{0}>\).


\begin{center}
\begin{picture}(110,73)
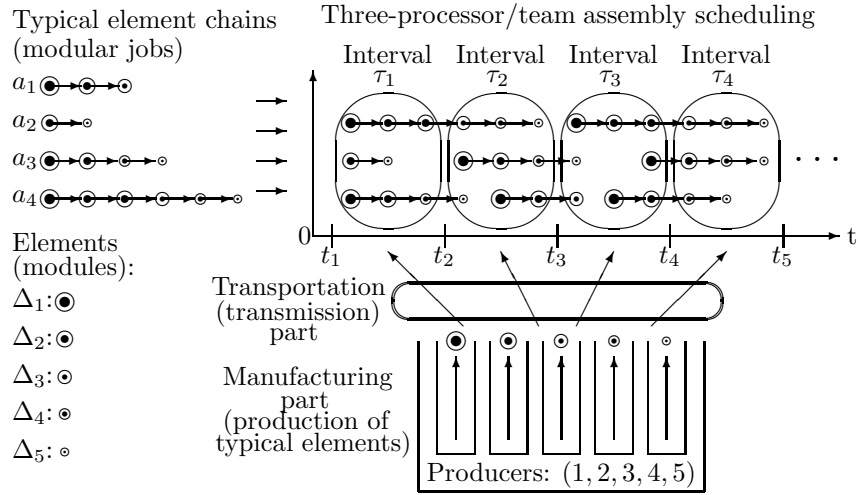
\put(00,00){\makebox(0,0)[bl]{Fig. 4. Multi-processor schedule
 (interval balance by element
  structure) }}

\put(00,67){\makebox(0,0)[bl]{Typical element chains}}
\put(00,63){\makebox(0,0)[bl]{(modular jobs)}}

\put(00,59){\makebox(0,0)[bl]{\(a_{1}\)}}

\put(05,60){\circle*{1.4}} \put(05,60){\circle{2.3}}
\put(10,60){\circle*{1.2}} \put(10,60){\circle{2.0}}
\put(15,60){\circle*{0.8}} \put(15,60){\circle{1.8}}

\put(05,60){\vector(1,0){4.2}} \put(10,60){\vector(1,0){4.2}}


\put(00,54){\makebox(0,0)[bl]{\(a_{2}\)}}

\put(05,55){\circle*{1.2}} \put(05,55){\circle{2.0}}
\put(10,55){\circle*{0.5}} \put(10,55){\circle{1.2}}

\put(05,55){\vector(1,0){4.2}}



\put(00,44){\makebox(0,0)[bl]{\(a_{4}\)}}


\put(05,45){\circle*{1.4}} \put(05,45){\circle{2.3}}

\put(10,45){\circle*{1.2}} \put(10,45){\circle{2.0}}
\put(15,45){\circle*{1.2}} \put(15,45){\circle{2.0}}

\put(20,45){\circle*{0.8}} \put(20,45){\circle{1.8}}

\put(25,45){\circle*{0.7}} \put(25,45){\circle{1.4}}

\put(30,45){\circle*{0.5}} \put(30,45){\circle{1.2}}

\put(05,45){\vector(1,0){4.2}} \put(10,45){\vector(1,0){4.2}}
\put(15,45){\vector(1,0){4.2}} \put(20,45){\vector(1,0){4.2}}
\put(25,45){\vector(1,0){4.2}}

\put(00,49){\makebox(0,0)[bl]{\(a_{3}\)}}

\put(05,50){\circle*{1.4}} \put(05,50){\circle{2.3}}
\put(10,50){\circle*{1.2}} \put(10,50){\circle{2.0}}
\put(15,50){\circle*{0.7}} \put(15,50){\circle{1.4}}
\put(20,50){\circle*{0.5}} \put(20,50){\circle{1.2}}

\put(05,50){\vector(1,0){4}} \put(10,50){\vector(1,0){4}}
\put(15,50){\vector(1,0){4}}

\put(00,38){\makebox(0,0)[bl]{Elements}}
\put(00,34){\makebox(0,0)[bl]{(modules):}}


\put(00,30){\makebox(0,0)[bl]{\(\Delta_{1}\):}}
\put(07,31.3){\circle*{1.4}} \put(07,31.3){\circle{2.3}}

\put(00,25){\makebox(0,0)[bl]{\(\Delta_{2}\):}}
\put(07,26.3){\circle*{1.2}} \put(07,26.3){\circle{2.0}}

\put(00,20){\makebox(0,0)[bl]{\(\Delta_{3}\):}}
\put(07,21.3){\circle*{0.8}} \put(07,21.3){\circle{1.8}}

\put(00,15){\makebox(0,0)[bl]{\(\Delta_{4}\):}}
\put(07,16.3){\circle*{0.7}} \put(07,16.3){\circle{1.4}}

\put(00,10){\makebox(0,0)[bl]{\(\Delta_{5}\):}}
\put(07,11.3){\circle*{0.5}} \put(07,11.3){\circle{1.2}}

\put(32.5,58){\vector(1,0){04}} \put(32.5,54){\vector(1,0){04}}
\put(32.5,50){\vector(1,0){04}} \put(32.5,46){\vector(1,0){04}}


\put(41,67.5){\makebox(0,0)[bl]{Three-processor/team
 assembly scheduling}}

\put(44,63){\makebox(0,0)[bl]{Interval}}
\put(48,60){\makebox(0,0)[bl]{\(\tau_{1}\)}}

\put(59,63){\makebox(0,0)[bl]{Interval}}
\put(63,60){\makebox(0,0)[bl]{\(\tau_{2}\)}}

\put(74,63){\makebox(0,0)[bl]{Interval}}
\put(78,60){\makebox(0,0)[bl]{\(\tau_{3}\)}}

\put(89,63){\makebox(0,0)[bl]{Interval}}
\put(93,60){\makebox(0,0)[bl]{\(\tau_{4}\)}}

\put(40,40){\vector(0,1){22}} \put(40,40){\vector(1,0){70}}

\put(38,39){\makebox(0,0)[bl]{\(0\)}}
\put(111,39){\makebox(0,0)[bl]{t}}

\put(104,50){\makebox(0,0)[bl]{{\bf .~.~.}}}


\put(45,45){\circle*{1.4}} \put(45,45){\circle{2.3}}
\put(50,45){\circle*{1.2}} \put(50,45){\circle{2.0}}
\put(55,45){\circle*{0.7}} \put(55,45){\circle{1.4}}
\put(60,45){\circle*{0.5}} \put(60,45){\circle{1.2}}

\put(45,45){\vector(1,0){4.2}} \put(50,45){\vector(1,0){4.2}}
\put(55,45){\vector(1,0){4.2}}


\put(65,45){\circle*{1.4}} \put(65,45){\circle{2.3}}
\put(70,45){\circle*{1.2}} \put(70,45){\circle{2.0}}
\put(75,45){\circle*{0.8}} \put(75,45){\circle{1.8}}

\put(65,45){\vector(1,0){4.2}} \put(70,45){\vector(1,0){4.2}}

\put(80,45){\circle*{1.4}} \put(80,45){\circle{2.3}}
\put(85,45){\circle*{1.2}} \put(85,45){\circle{2.0}}
\put(90,45){\circle*{0.7}} \put(90,45){\circle{1.4}}
\put(95,45){\circle*{0.5}} \put(95,45){\circle{1.2}}

\put(80,45){\vector(1,0){4.2}} \put(85,45){\vector(1,0){4.2}}
\put(90,45){\vector(1,0){4.2}}

\put(45,50){\circle*{1.2}} \put(45,50){\circle{2.0}}
\put(50,50){\circle*{0.5}} \put(50,50){\circle{1.2}}

\put(45,50){\vector(1,0){4.2}}


\put(60,50){\circle*{1.4}} \put(60,50){\circle{2.3}}
\put(65,50){\circle*{1.2}} \put(65,50){\circle{2.0}}
\put(70,50){\circle*{0.7}} \put(70,50){\circle{1.4}}
\put(75,50){\circle*{0.5}} \put(75,50){\circle{1.2}}


\put(60,50){\vector(1,0){4}} \put(65,50){\vector(1,0){4}}
\put(70,50){\vector(1,0){4}}


\put(85,50){\circle*{1.4}} \put(85,50){\circle{2.3}}
\put(90,50){\circle*{1.2}} \put(90,50){\circle{2.0}}
\put(95,50){\circle*{0.7}} \put(95,50){\circle{1.4}}
\put(100,50){\circle*{0.5}} \put(100,50){\circle{1.2}}


\put(85,50){\vector(1,0){4}} \put(90,50){\vector(1,0){4}}
\put(95,50){\vector(1,0){4}}


\put(45,55){\circle*{1.4}} \put(45,55){\circle{2.3}}
\put(50,55){\circle*{1.2}} \put(50,55){\circle{2.0}}
\put(55,55){\circle*{1.2}} \put(55,55){\circle{2.0}}
\put(60,55){\circle*{0.8}} \put(60,55){\circle{1.9}}
\put(65,55){\circle*{0.7}} \put(65,55){\circle{1.4}}
\put(70,55){\circle*{0.5}} \put(70,55){\circle{1.2}}

\put(45,55){\vector(1,0){4.2}} \put(50,55){\vector(1,0){4.2}}
\put(55,55){\vector(1,0){4.2}} \put(60,55){\vector(1,0){4.2}}
\put(65,55){\vector(1,0){4.2}}


\put(75,55){\circle*{1.4}} \put(75,55){\circle{2.3}}
\put(80,55){\circle*{1.2}} \put(80,55){\circle{2.0}}
\put(85,55){\circle*{1.2}} \put(85,55){\circle{2.0}}
\put(90,55){\circle*{0.8}} \put(90,55){\circle{1.9}}
\put(95,55){\circle*{0.7}} \put(95,55){\circle{1.4}}
\put(100,55){\circle*{0.5}} \put(100,55){\circle{1.2}}

\put(75,55){\vector(1,0){4.2}} \put(80,55){\vector(1,0){4.2}}
\put(85,55){\vector(1,0){4.2}} \put(90,55){\vector(1,0){4.2}}
\put(95,55){\vector(1,0){4.2}}

\put(50,50){\oval(14,18)} \put(65,50){\oval(14,18)}
\put(80,50){\oval(14,18)} \put(95,50){\oval(14,18)}

\put(42.5,38.5){\line(0,1){3}} \put(57.5,38.5){\line(0,1){3}}
\put(72.5,38.5){\line(0,1){3}} \put(87.5,38.5){\line(0,1){3}}
\put(102.5,38.5){\line(0,1){3}}

\put(41,36){\makebox(0,0)[bl]{\(t_{1}\)}}
\put(56,36){\makebox(0,0)[bl]{\(t_{2}\)}}
\put(71,36){\makebox(0,0)[bl]{\(t_{3}\)}}
\put(86,36){\makebox(0,0)[bl]{\(t_{4}\)}}
\put(101,36){\makebox(0,0)[bl]{\(t_{5}\)}}

\put(26.5,31.5){\makebox(0,0)[bl]{Transportation}}
\put(27,28){\makebox(0,0)[bl]{(transmission)}}
\put(34.2,25.5){\makebox(0,0)[bl]{part}}

\put(72.5,31.4){\oval(44,5)} \put(72.5,31.4){\oval(43.5,4.5)}

\put(60,28){\vector(-1,1){10}} \put(70,28){\vector(-1,2){5}}
\put(75,28){\vector(1,2){5}} \put(85,28){\vector(1,1){10}}

\put(28,19.5){\makebox(0,0)[bl]{Manufacturing}}
\put(35.7,16.5){\makebox(0,0)[bl]{part}}
\put(28,13.5){\makebox(0,0)[bl]{(production of}}
\put(27,10.5){\makebox(0,0)[bl]{typical elements)}}

\put(54,06){\line(0,1){20}} \put(92,06){\line(0,1){20}}
\put(54,06){\line(1,0){38}}

\put(56.5,11){\line(0,1){15}} \put(61.5,11){\line(0,1){15}}
\put(56.5,11){\line(1,0){5}} \put(59,13){\vector(0,1){11}}
\put(59,26){\circle*{1.4}} \put(59,26){\circle{2.3}}

\put(63.5,11){\line(0,1){15}} \put(68.5,11){\line(0,1){15}}
\put(63.5,11){\line(1,0){5}} \put(66,13){\vector(0,1){11}}
\put(66,26){\circle*{1.2}} \put(66,26){\circle{2.0}}

\put(70.5,11){\line(0,1){15}} \put(75.5,11){\line(0,1){15}}
\put(70.5,11){\line(1,0){5}} \put(73,13){\vector(0,1){11}}
\put(73,26){\circle*{0.8}} \put(73,26){\circle{1.8}}

\put(77.5,11){\line(0,1){15}} \put(82.5,11){\line(0,1){15}}
\put(77.5,11){\line(1,0){5}} \put(80,13){\vector(0,1){11}}
\put(80,26){\circle*{0.7}} \put(80,26){\circle{1.4}}


\put(84.5,11){\line(0,1){15}} \put(89.5,11){\line(0,1){15}}
\put(84.5,11){\line(1,0){5}} \put(87,13){\vector(0,1){11}}
\put(87,26){\circle*{0.5}} \put(87,26){\circle{1.2}}

\put(55.1,06.5){\makebox(0,0)[bl]{Producers: (\(1,2,3,4,5\))}}

\end{picture}
\end{center}

 The basic requirement for the general schedule \(S\)
 is targeted to designing the assembly schedule
 with  balanced (by element/module structure) time intervals,
 here:
  \(\tau_{1} = [t_{1},t_{2}]\), \(\tau_{2}= [t_{2},t_{3}]\),
   \(\tau_{3}= [t_{3},t_{4}]\), \(\tau_{4}= [t_{4},t_{5}]\)).
 At each time interval, elements are used
 (additional 6th ``empty'' element type is used \(\Delta_{6}\), the type corresponds
 to ``empty'' element/time period \(a_{0}\)):

 (1)
 \(X_{\tau_{1}}=\{\Delta_{1},\Delta_{1},\Delta_{2}, \Delta_{2},\Delta_{2},\Delta_{2},
   \Delta_{3},\Delta_{5},\Delta_{6} \}\),
 (2)
 \(X_{\tau_{2}}=\{\Delta_{1},\Delta_{1},\Delta_{2}, \Delta_{2},\Delta_{3},\Delta_{4},
 \Delta_{4},\Delta_{5},\Delta_{5}\}\),

 (3)
 \(X_{\tau_{3}}=\{\Delta_{1},\Delta_{1},\Delta_{1}, \Delta_{2},\Delta_{2},\Delta_{2},
 \Delta_{3},\Delta_{5},\Delta_{6}\}\),
 (4)
 \(X_{\tau_{4}}=\{\Delta_{2},\Delta_{3},\Delta_{4},
 \Delta_{4},\Delta_{4},\Delta_{5}, \Delta_{5},\Delta_{5},\Delta_{6}\}\).

  Evidently, the clustering solution is:
 ~\(\widetilde{X}(S) = \{X_{\tau_{1}},X_{\tau_{2}},X_{\tau_{3}},X_{\tau_{4}}\}\).

 The corresponding multiset estimates are
 (by number of basic element types  \cite{lev12a,lev15}):~

%
 \(e(X_{\tau_{1}}) = (2,4,1,0,1,1)\), \(e(X_{\tau_{2}}) = (2,2,1,2,2,0)\),
 \(e(X_{\tau_{3}}) = (3,3,1,0,1,1)\), \(e(X_{\tau_{4}}) = (0,1,1,3,3,1)\).

 The basic reference multiset estimate has to be in correspondence to
 an output structure (i.e., productivity)
 of the manufacturing (production) system
 (while taking into account transportation system),
 for example:
 \(e_{0} = (2,3,2,1,1,0)\).

 The proximity between multiset estimates
 (\( \delta (e_{0}, e(X_{\tau_{\iota}}) ) \), \( \iota =\overline{1,4}\))
 are shown in Table 9.

\begin{center}
\begin{small}
 {\bf Table 9.} Proximities between cluster estimates and reference structure estimate
 (\( \delta (e_{0}, e(X_{\tau_{\iota}}) ) \)
 \\
\begin{tabular}{| c | l c c c c|}
\hline
  Reference &Cluster
   &\(e(X_{\tau_{1}})=\)&\(e(X_{\tau_{2}})=\)&\(e(X_{\tau_{3}})=\)
   &\(e(X_{\tau_{4}})=\) \\

   estimate &estimate:
   &\((2,4,1,0,1,1)\)&\((2,2,1,2,2,0)\)&\((3,3,1,0,1,1)\)
   &\((0,1,1,3,3,1)\) \\

\hline


 \(e_{0}= (2,3,2,1,1,0)\)&&\(3\)&\(3\)&\(4\)&\(15\)\\

\hline
\end{tabular}
\end{small}
\end{center}

 Generally, the set of processors/team is
 \(P = \{P_{1},...,P_{\xi},...,P_{m}\}  \).
 The scheduling problem is:

~~

 Find the general schedule \(S\) and
 corresponding clustering solution
 \(\widetilde{X}(S) =
  \{X_{\tau_{1}},X_{\tau_{2}},X_{\tau_{3}},..,X_{\tau_{k}}\}\)
 such that
 (i) the number of time intervals (the number \(k\)), the length of schedule, makespan)
 is minimized,
 (ii) proximity between each interval element structure and
 reference structure is limited
  (i.e.,
 \( \max_{\iota=\overline{1,k} }~~ \delta (e_{0}, e(X_{\tau_{\iota}}) )
  \leq \delta^{0}\),
 \(\delta^{0}\) is a joint constraint of manufacturing and transportation part).

~~

 The general formal model is as follows (here: \(m\) processor/teams, \(L(S) = k\)):
 \[
 \min~ L(S) = \max_{\xi=\overline{1,m}} L(S_{\xi})
 ~~~ s.t. ~~
 \max_{\iota=\overline{1,k} }~~ \delta (e_{0}, e( X_{\tau_{\iota}}) )
  \leq \delta^{0}.
%
%
 \]
 In the considered numerical example,
 \(L(S)=4\),
 \( \widehat{B} (\widetilde{X}(S))  =  4\).

 An illustration of the examined hierarchy
 is depicted in Fig. 5.

\begin{center}
\begin{picture}(80,48)
\put(06,00){\makebox(0,0)[bl]{Fig. 5.
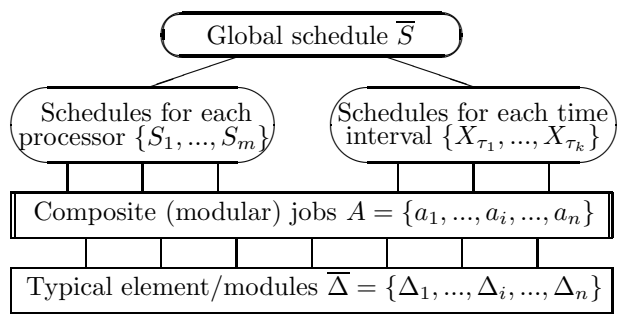
 Hierarchy of examined components }}

\put(26,41.8){\makebox(0,0)[bl]{Global schedule  \( \overline{S}\) }}

\put(40,43){\oval(40,06)} \put(40,43){\oval(39.4,05.5)}
\put(30,40){\line(-3,-1){12}} \put(50,40){\line(3,-1){12}}

\put(04,31.5){\makebox(0,0)[bl]{Schedules for each}}
\put(01,27.5){\makebox(0,0)[bl]{processor \(\{S_{1},...,S_{m}\}\) }}

%
\put(17.5,31){\oval(35,10)}

\put(07.5,22){\line(0,1){04}} \put(17.5,22){\line(0,1){04}}
\put(27.5,22){\line(0,1){04}}

\put(43.4,31.5){\makebox(0,0)[bl]{Schedules for each time }}
\put(44.5,27.5){\makebox(0,0)[bl]{interval \(\{X_{\tau_{1}},...,X_{\tau_{k}} \}\) }}

%
\put(61.5,31){\oval(38,10)}

\put(51.5,22){\line(0,1){04}} \put(61.5,22){\line(0,1){04}}
\put(71.5,22){\line(0,1){04}}

\put(03,17.5){\makebox(0,0)[bl]{Composite (modular) jobs
  \( A =
   \{a_{1},...,a_{i},...,a_{n}  \}\) }}

\put(00,16){\line(1,0){80}} \put(00,22){\line(1,0){80}}
\put(00,16){\line(0,1){06}} \put(80,16){\line(0,1){06}}
\put(0.5,16){\line(0,1){06}} \put(79.5,16){\line(0,1){06}}

\put(10,12){\line(0,1){04}} \put(20,12){\line(0,1){04}}
\put(30,12){\line(0,1){04}} \put(40,12){\line(0,1){04}}
\put(50,12){\line(0,1){04}} \put(60,12){\line(0,1){04}}
\put(70,12){\line(0,1){04}}

\put(02,07.5){\makebox(0,0)[bl]{Typical element/modules
  \( \overline{\Delta} =
   \{\Delta_{1},...,\Delta_{i},...,\Delta_{n}  \}\) }}

\put(00,06){\line(1,0){80}} \put(00,12){\line(1,0){80}}
\put(00,06){\line(0,1){06}} \put(80,06){\line(0,1){06}}
%

\end{picture}
\end{center}

 Clearly, many kinds (versions) of the proposed problem can be examined
 (e.g.,
 various scales for the parameters estimates,
 various objective functions, various constraints, various levels of uncertainty).
 The described problem types can de useful for planning several domains:
 manufacturing systems, home-building systems,
 logistics (supply chain management),
 information transmission systems.

 In addition, it is reasonable to point out the following notes:

 {\it Note 1.} It is possible to examine various model formulation for the above-mentioned problem
 (i.e., various constraints, various objective functions).

 {\it Note 2.} The problems of the considered kind
  are very complicated, i.e,   NP-hard
  (outside some simplified cases)
 Evidently, additional studies of the problem complexities have to
 be conducted.

 {\it Note 3.} Various heuristics and metaheuristics
 can be considered as prospective ones
 as the solving schemes for the problem above.

\newpage
\section{Time-interval balanced planning in modular house-building}

\subsection{Life-cycle and home-building conveyor}

 In general, the life cycle of home-building
  can be considered as the following
  chain of phases (Fig. 6):

~~

 {\bf 0.} Preliminary phase:
 analysis of requirements to the building,
 land planning,
 analysis of the building functionality,
 geological exploration, etc.

 {\bf 1.} Ideation and design phase:
 (1.1) architectural design (generation of the basic architectural ideas),
 (1.2) construction design (real design stage to prepare design
 documentation).

 {\bf 2.} Manufacturing phase:
  manufacturing of building details and components.

 {\bf 3.} Transportation and assembly phase:
 (3.1) transportation of building components to
 the assembly place,
 (3.2) assembly of the building.

 {\bf 4.} Utilization phase: utilization including maintenance.

 {\bf 5.} Recycling phase:
  (5.1) destruction of the building,
  (5.2) recycling of the building components.

~~

 Modern industrial technology for modular homebuilding provides
 faster and lower cost building process
 (e.g., \cite{laws14,mull11}).
%
%
 In 1980,
 a new system of big structural panel based homebuilding in Moscow has been
 suggested
  \cite{mak82,roch80}.
 The homebuilding system was based on
  catalogue big structural panel elements
  (``method KOPE'').
%
 In the homebuilding system,
 a coordination (i.e., balancing) between structural panel manufacturing stage and
 homebuilding (assembly) stage was a bottleneck.
 Here,
 the material is targeted to the coordination above (i.e., the bottleneck).
 The considered planning approach is based on
 a real-world home-building conveyor
 (DSK-2, Ochakovo/Moscow, 1982) (Fig. 7) \cite{lev83}:

~~

 {\bf I.} Architectural part:~
 1.1. general architectural design,
 1.2. construction design.

 {\bf II.} Production part:~
 2.1. manufacturing system
 (industry based manufacturing of home details-modules/structural panels),
 2.2. transportation system
 (transportation of the details/modules to home places),
 2.3. assembly systems
  (assembly of the building from the details/modules).

~~



 Note,
 preliminary stages for the conveyor
 involve land planning, geological exploration, etc.

\begin{center}
\begin{picture}(150,30)
\put(34.5,00){\makebox(0,0)[bl]{Fig. 6.
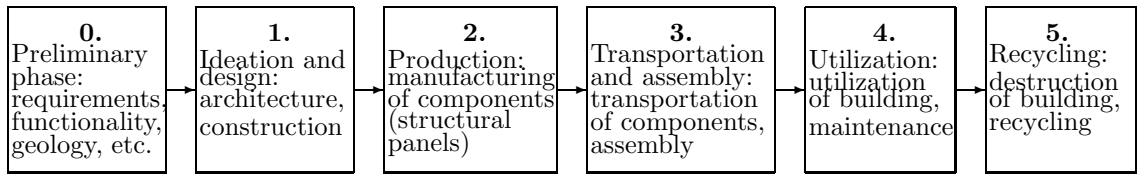
 Scheme of general home-building life cycle }}

\put(09.5,23.5){\makebox(0,0)[bl]{{\bf 0.}}}
\put(0.5,20){\makebox(0,0)[bl]{Preliminary}}
\put(0.5,17){\makebox(0,0)[bl]{phase:}}
\put(0.5,14){\makebox(0,0)[bl]{requirements,}}
\put(0.5,11){\makebox(0,0)[bl]{functionality,}}
\put(0.5,08){\makebox(0,0)[bl]{geology,  etc.}}

\put(00,06){\line(1,0){21}} \put(00,28){\line(1,0){21}}
\put(00,06){\line(0,1){22}} \put(21,06){\line(0,1){22}}

\put(21,17){\vector(1,0){4}}

\put(34.5,23.5){\makebox(0,0)[bl]{{\bf 1.}}}
\put(25.5,20){\makebox(0,0)[bl]{Ideation and}}
\put(25.5,17){\makebox(0,0)[bl]{design: }}
\put(25.5,14){\makebox(0,0)[bl]{architecture,}}
\put(25.5,11){\makebox(0,0)[bl]{construction}}

\put(25,06){\line(1,0){21}} \put(25,28){\line(1,0){21}}
\put(25,06){\line(0,1){22}} \put(46,06){\line(0,1){22}}

\put(46,17){\vector(1,0){4}}

\put(60.5,23.5){\makebox(0,0)[bl]{{\bf 2.}}}
\put(50.5,20){\makebox(0,0)[bl]{Production:}}
\put(50.5,17){\makebox(0,0)[bl]{manufacturing}}
\put(50.5,14){\makebox(0,0)[bl]{of components}}
\put(50.5,11){\makebox(0,0)[bl]{(structural }}
\put(50.5,08){\makebox(0,0)[bl]{panels)}}

\put(50,06){\line(1,0){23}} \put(50,28){\line(1,0){23}}
\put(50,06){\line(0,1){22}} \put(73,06){\line(0,1){22}}

\put(73,17){\vector(1,0){4}}

\put(88,23.5){\makebox(0,0)[bl]{{\bf 3.}}}
\put(77.5,20){\makebox(0,0)[bl]{Transportation}}
\put(77.5,17){\makebox(0,0)[bl]{and assembly:}}
\put(77.5,14){\makebox(0,0)[bl]{transportation}}
\put(77.5,11){\makebox(0,0)[bl]{of components, }}
\put(77.5,08){\makebox(0,0)[bl]{assembly}}

\put(77,06){\line(1,0){25}} \put(77,28){\line(1,0){25}}
\put(77,06){\line(0,1){22}} \put(102,06){\line(0,1){22}}

\put(102,17){\vector(1,0){4}}

\put(115,23.5){\makebox(0,0)[bl]{{\bf 4.}}}
\put(106.5,20){\makebox(0,0)[bl]{Utilization: }}
\put(106.5,17){\makebox(0,0)[bl]{utilization}}
\put(106.5,14){\makebox(0,0)[bl]{of building,}}
\put(106.5,11){\makebox(0,0)[bl]{maintenance }}
\put(106.5,08){\makebox(0,0)[bl]{}}

\put(106,06){\line(1,0){20}} \put(106,28){\line(1,0){20}}
\put(106,06){\line(0,1){22}} \put(126,06){\line(0,1){22}}

\put(126,17){\vector(1,0){4}}

\put(138,23.5){\makebox(0,0)[bl]{{\bf 5.}}}
\put(130.5,20){\makebox(0,0)[bl]{Recycling:}}
\put(130.5,17){\makebox(0,0)[bl]{destruction}}
\put(130.5,14){\makebox(0,0)[bl]{of building, }}
\put(130.5,11){\makebox(0,0)[bl]{recycling}}

\put(130,06){\line(1,0){20}} \put(130,28){\line(1,0){20}}
\put(130,06){\line(0,1){22}} \put(150,06){\line(0,1){22}}

\end{picture}
\end{center}

\begin{center}
\begin{picture}(131,31)
\put(018,00){\makebox(0,0)[bl]{Fig. 7.
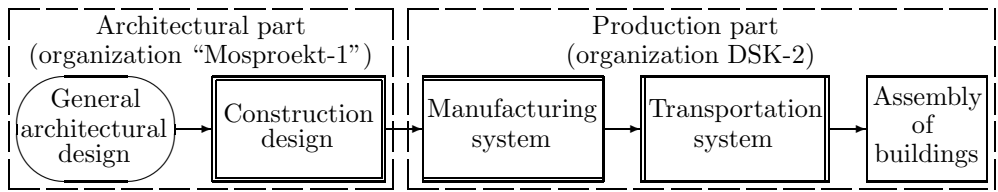
 Scheme of home-building conveyor (Moscow, 1982)}}

\put(05.5,17){\makebox(0,0)[bl]{General}}
\put(01.5,13){\makebox(0,0)[bl]{architectural }}
\put(06.5,09){\makebox(0,0)[bl]{design}}


\put(11.5,14){\oval(21,14)}

\put(22,14){\vector(1,0){5}}

\put(28.5,15){\makebox(0,0)[bl]{Construction}}
\put(33.5,11){\makebox(0,0)[bl]{design}}

\put(27,07){\line(1,0){23}} \put(27,21){\line(1,0){23}}
\put(27,07){\line(0,1){14}} \put(50,07){\line(0,1){14}}

\put(27.5,07.5){\line(1,0){22}} \put(27.5,20.5){\line(1,0){22}}
\put(27.5,07.5){\line(0,1){13}} \put(49.5,07.5){\line(0,1){13}}

\put(50,14){\vector(1,0){5}}

\put(11.5,26){\makebox(0,0)[bl]{Architectural part}}
\put(02.6,22){\makebox(0,0)[bl]{(organization ``Mosproekt-1'')}}

\put(00,06){\line(1,0){04}} \put(05,06){\line(1,0){04}}
\put(10,06){\line(1,0){04}} \put(15,06){\line(1,0){04}}
\put(20,06){\line(1,0){04}} \put(25,06){\line(1,0){04}}
\put(30,06){\line(1,0){04}} \put(35,06){\line(1,0){04}}
\put(40,06){\line(1,0){05}} \put(46,06){\line(1,0){05}}

\put(00,30){\line(1,0){04}} \put(05,30){\line(1,0){04}}
\put(10,30){\line(1,0){04}} \put(15,30){\line(1,0){04}}
\put(20,30){\line(1,0){04}} \put(25,30){\line(1,0){04}}
\put(30,30){\line(1,0){04}} \put(35,30){\line(1,0){04}}
\put(40,30){\line(1,0){05}} \put(46,30){\line(1,0){05}}

\put(00,06){\line(0,1){04}} \put(00,11){\line(0,1){04}}
\put(00,16){\line(0,1){04}} \put(00,21){\line(0,1){04}}
\put(00,26){\line(0,1){04}}

\put(51,06){\line(0,1){04}} \put(51,11){\line(0,1){04}}
\put(51,16){\line(0,1){04}} \put(51,21){\line(0,1){04}}
\put(51,26){\line(0,1){04}}

\put(55.6,15){\makebox(0,0)[bl]{Manufacturing}}
\put(61.5,11){\makebox(0,0)[bl]{system}}

\put(55,07){\line(1,0){24}} \put(55,21){\line(1,0){24}}
\put(55,07.5){\line(1,0){24}} \put(55,20.5){\line(1,0){24}}

\put(55,07){\line(0,1){14}} \put(79,07){\line(0,1){14}}

\put(79,14){\vector(1,0){5}}

\put(85,15){\makebox(0,0)[bl]{Transportation}}
\put(91,11){\makebox(0,0)[bl]{system}}

\put(84,07){\line(1,0){25}} \put(84,21){\line(1,0){25}}
\put(84,07){\line(0,1){14}} \put(109,07){\line(0,1){14}}
\put(84.5,07){\line(0,1){14}} \put(108.5,07){\line(0,1){14}}

\put(109,14){\vector(1,0){5}}

\put(114.8,16.5){\makebox(0,0)[bl]{Assembly}}
\put(120,13.5){\makebox(0,0)[bl]{of}}
\put(115,09){\makebox(0,0)[bl]{buildings}}

\put(114,07){\line(1,0){16}} \put(114,21){\line(1,0){16}}
\put(114,07){\line(0,1){14}} \put(130,07){\line(0,1){14}}

\put(77.5,26){\makebox(0,0)[bl]{Production  part}}
\put(74,22){\makebox(0,0)[bl]{(organization DSK-2)}}

\put(53,06){\line(1,0){04}} \put(58,06){\line(1,0){04}}
\put(63,06){\line(1,0){04}} \put(68,06){\line(1,0){04}}
\put(73,06){\line(1,0){04}} \put(78,06){\line(1,0){04}}
\put(83,06){\line(1,0){04}} \put(88,06){\line(1,0){04}}
\put(93,06){\line(1,0){04}} \put(98,06){\line(1,0){04}}
\put(103,06){\line(1,0){04}} \put(108,06){\line(1,0){04}}
\put(113,06){\line(1,0){05}} \put(119,06){\line(1,0){05}}
\put(125,06){\line(1,0){05}}

\put(53,30){\line(1,0){04}} \put(58,30){\line(1,0){04}}
\put(63,30){\line(1,0){04}} \put(68,30){\line(1,0){04}}
\put(73,30){\line(1,0){04}} \put(78,30){\line(1,0){04}}
\put(83,30){\line(1,0){04}} \put(88,30){\line(1,0){04}}
\put(93,30){\line(1,0){04}} \put(98,30){\line(1,0){04}}
\put(103,30){\line(1,0){04}} \put(108,30){\line(1,0){04}}
\put(113,30){\line(1,0){05}} \put(119,30){\line(1,0){05}}
\put(125,30){\line(1,0){05}}

\put(53,06){\line(0,1){04}} \put(53,11){\line(0,1){04}}
\put(53,16){\line(0,1){04}} \put(53,21){\line(0,1){04}}
\put(53,26){\line(0,1){04}}

\put(131,06){\line(0,1){04}} \put(131,11){\line(0,1){04}}
\put(131,16){\line(0,1){04}} \put(131,21){\line(0,1){04}}
\put(131,26){\line(0,1){04}}

\end{picture}
\end{center}

 The result of the conveyor consists in building(s).
 Here, the manufacturing system is  the most
 capitalized component of the conveyor.
 Thus,  conveyor components
 as transportation and assembly
 have to be time-balanced
 (i.e., about synchronized)
 on the basis of
 time-productivity of manufacturing part.
 Buffers between manufacturing part and
 assembly is not reliable and conveyor is based on the principle
 ``just-in-time'' \cite{groe93,joze10,mond11,rios12}.
%
%
%
 The real world time requirement (i.e., time constraint)
 for store of module/detail between transportation system and
 assembly
 is: ~\( \leq \)~ three days.

 Finally,
 the examined planning problem of assembly of modular buildings
 is targeted to designing a balanced assembly process
 (i.e., time-synchronized with manufacturing process)
 where time unbalance is
 for each module/detail is ~\( \leq \)~ three days.

%

 Further, a problem of balanced planning in house-building is examined.
 The problem description is based in the designed and implemented
  applied software planning system \cite{lev83}.
 In general, the system involves three stages (Fig. 8):
 (i) manufacturing of typical building modules
 (details, elements, structural panels);
 (ii) transportation of the elements to areas of house-building
 (with very small buffers);
 and
 (iii) assembly of buildings.
 Note, the small buffers require organizational decisions
 as quasi ``just-in-time''.

 In homebuilding systems, a real time interval can be equal to
 one week or to ten days.
 In the next section (in the numerical example),
 the considered time interval
 equals one month (for simplification).

\begin{center}
\begin{picture}(29,36)

\put(00,26){\makebox(0,0)[bl]{MANAGEMENT}}
\put(06,22){\makebox(0,0)[bl]{LAYER}}

\put(02,12){\makebox(0,0)[bl]{``PHYSICAL''}}
\put(06,08){\makebox(0,0)[bl]{LAYER}}
\end{picture}
%
\begin{picture}(110,36)
\put(00,00){\makebox(0,0)[bl]{Fig. 8.
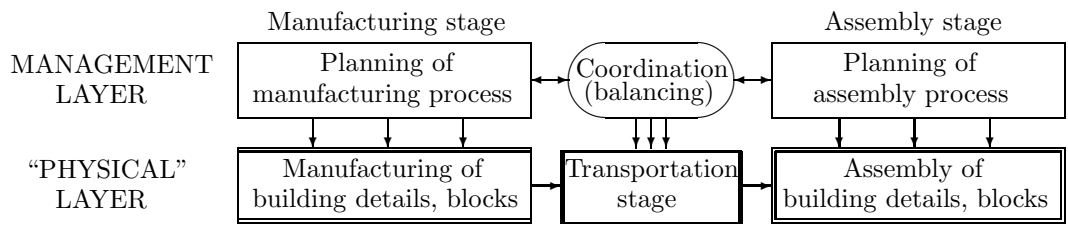
 Scheme of manufacturing and house-building}}

\put(04,31){\makebox(0,0)[bl]{Manufacturing stage}}

\put(11,25.5){\makebox(0,0)[bl]{Planning of}}
\put(02,21.5){\makebox(0,0)[bl]{manufacturing process}}

\put(00,20){\line(1,0){39}} \put(00,30){\line(1,0){39}}
\put(00,20){\line(0,1){10}} \put(39,20){\line(0,1){10}}


\put(10,20){\vector(0,-1){4}} \put(20,20){\vector(0,-1){4}}
\put(30,20){\vector(0,-1){04}}


\put(06,11.5){\makebox(0,0)[bl]{Manufacturing of}}
\put(02,07.5){\makebox(0,0)[bl]{building details, blocks}}

\put(00,06){\line(1,0){39}} \put(00,16){\line(1,0){39}}
\put(00,06.5){\line(1,0){39}} \put(00,15.5){\line(1,0){39}}
\put(00,06){\line(0,1){10}} \put(39,06){\line(0,1){10}}


\put(39,11){\vector(1,0){04}}

\put(40,25){\vector(1,0){04}} \put(43,25){\vector(-1,0){04}}
\put(67,25){\vector(1,0){04}} \put(70,25){\vector(-1,0){04}}

\put(45,25.5){\makebox(0,0)[bl]{Coordination}}
\put(46,21.5){\makebox(0,0)[bl]{(balancing)}}

\put(55,25){\oval(22,10)}

\put(55,20){\vector(0,-1){04}} \put(57,20){\vector(0,-1){04}}
\put(53,20){\vector(0,-1){04}}

\put(43,06){\line(1,0){24}} \put(43,15.5){\line(1,0){24}}
\put(43,06){\line(0,1){09.5}} \put(67,06){\line(0,1){09.5}}

\put(43.3,06){\line(0,1){09.5}} \put(66.7,06){\line(0,1){09.5}}

\put(43.5,011.5){\makebox(0,0)[bl]{Transportation}}
\put(50.5,07.5){\makebox(0,0)[bl]{stage}}

\put(67,11){\vector(1,0){04}}

\put(78,31){\makebox(0,0)[bl]{Assembly stage}}

\put(80.5,25.5){\makebox(0,0)[bl]{Planning of}}
\put(76.5,21.5){\makebox(0,0)[bl]{assembly process}}

\put(71,20){\line(1,0){39}} \put(71,30){\line(1,0){39}}
\put(71,20){\line(0,1){10}} \put(110,20){\line(0,1){10}}


\put(80,20){\vector(0,-1){4}} \put(90,20){\vector(0,-1){4}}
\put(100,20){\vector(0,-1){04}}


\put(80.5,11.5){\makebox(0,0)[bl]{Assembly of}}
\put(72.5,07.5){\makebox(0,0)[bl]{building details, blocks}}

\put(71,06){\line(1,0){39}} \put(71,16){\line(1,0){39}}
\put(71,06){\line(0,1){10}} \put(110,06){\line(0,1){10}}

\put(71.5,06.5){\line(1,0){38}} \put(71.5,15.5){\line(1,0){38}}
\put(71.5,06.5){\line(0,1){09}} \put(109.5,06.5){\line(0,1){09}}

\end{picture}
\end{center}

\subsection{Real world illustrative numerical example}

 In this section,
 a realistic numerical example of a two-year plan for home-building conveyor
  is described
 (``method KOPE'', Moscow home-building system, DSK-2, Ochakovo/Moscow, 1982..1983) \cite{lev83}.
  The example is based on
   FORTRAN-software that was
  designed by the author
  \cite{lev83}.

 The following basic architectural sections
 (the structural modules correspond to building ``column'' ) are used:

 (a) catalogue structural elements (corresponding to apartment):
   \(g_{1}\), \(g_{2}\), \(g_{5}\), \(g_{9}\);

 (b) composition elements (for connection of the sections)
   \(w_{1}\), \(w_{2}\), \(w_{3}\), \(w_{6}\), \(w_{7}\).


 The following basic types of structural details (manufacturing stage)
 are considered:
%
%
 (1) literal walls (panels) \(d_{1}\),
%
 (2) separation walls (panels)
 for 1-st floor  \(d_{2}\),
%
 (3) separation walls (panels)
  (2nd floor and so on) \(d_{3}\),
%
 (4) floor panels (intermediate slabs, floor slab panels) \(d_{4}\),
%
 (5) roof panels (slubs, cover bulkheads) \(d_{5}\),
%
 (6) fliers (stair flights)  \(d_{6}\),
%
 (7) elements of blocks and engineering communication \(d_{7}\),
 (8) balcony elements \(d_{8}\).
%
%
 Note, real set of the basic
 structural details in ``method KOPE'' involves about 500
 elements, but for the balancing problem  the above-mentioned \(8\)
  basic  integrated types (groups).
 Note,  detail \(d_{1}\) corresponds to the main structural detail
 while taking into account production subsystem because
 production of \(d_{1}\) is the basic manufacturing process (by cost and by time).
 As a result, the number of the required
 details of this kind defines (mainly) the balance
 between the production stage and the assembly stage.

 Here, the following typical floors are examined:
 (i) crawl space \(r_{1}\),
%
 (2) floor 1 \(r_{2}\),
 (3) bottom floor \(r_{3}\),
 (4) medium floor  \(r_{4}\),
 (5) top floor \(r_{5}\),
 (6) next floor \(r_{6}\),
 (7) the upper storey \(r_{7}\),
 and
 (8) engineering storey (mechanical floor) \(r_{8}\).
%
 The construction hierarchy (i.e., hierarchical system)
  of structural panels-buildings is depicted in Fig. 9.

\begin{center}
\begin{picture}(100,61)
\put(04,00){\makebox(0,0)[bl]{Fig. 9.
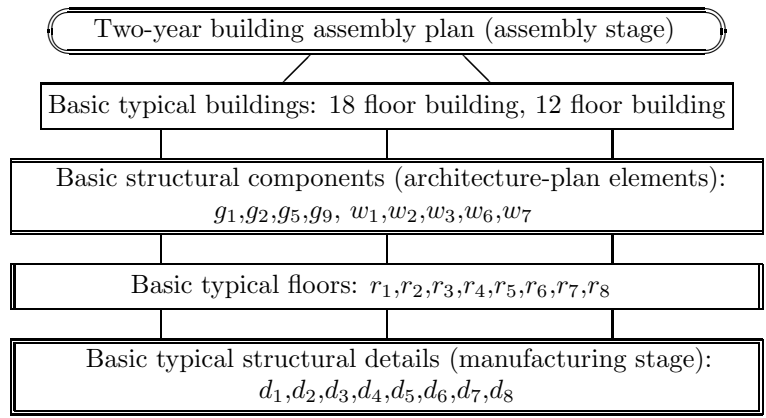
 Construction hierarchy: structural panels-buildings }}

\put(50,57){\oval(90,6)} \put(50,57){\oval(89,5)}



\put(11,55.3){\makebox(0,0)[bl]{Two-year building assembly plan
 (assembly stage)}}

\put(40,54){\line(-1,-1){04}} \put(60,54){\line(1,-1){04}}


\put(04,44){\line(1,0){92}} \put(04,50){\line(1,0){92}}
\put(04,44){\line(0,1){06}} \put(96,44){\line(0,1){06}}

\put(05,45.5){\makebox(0,0)[bl]{Basic typical buildings:
  18 floor building, 12 floor building }}

\put(20,40){\line(0,1){04}} \put(50,40){\line(0,1){04}}
\put(80,40){\line(0,1){04}}

\put(00,30.4){\line(1,0){100}} \put(00,39.6){\line(1,0){100}}

\put(00,30){\line(1,0){100}} \put(00,40){\line(1,0){100}}
\put(00,30){\line(0,1){10}} \put(100,30){\line(0,1){10}}

\put(06,35.5){\makebox(0,0)[bl]{Basic structural components
(architecture-plan elements):}}

\put(27,31.5){\makebox(0,0)[bl]
 {\(g_{1}\),\(g_{2}\),\(g_{5}\),\(g_{9}\),
 \(w_{1}\),\(w_{2}\),\(w_{3}\),\(w_{6}\),\(w_{7}\)  }}

\put(20,26){\line(0,1){04}} \put(50,26){\line(0,1){04}}
\put(80,26){\line(0,1){04}}

\put(00,20){\line(1,0){100}} \put(00,26){\line(1,0){100}}
\put(00,20){\line(0,1){06}} \put(100,20){\line(0,1){06}}
\put(00.5,20){\line(0,1){06}} \put(99.5,20){\line(0,1){06}}

\put(016,21.5){\makebox(0,0)[bl]{Basic typical floors:
\(r_{1}\),\(r_{2}\),\(r_{3}\),\(r_{4}\),\(r_{5}\),\(r_{6}\),\(r_{7}\),\(r_{8}\)}}

\put(20,16){\line(0,1){04}} \put(50,16){\line(0,1){04}}
\put(80,16){\line(0,1){04}}

\put(00,06){\line(1,0){100}} \put(00,16){\line(1,0){100}}
\put(00,06){\line(0,1){10}} \put(100,06){\line(0,1){10}}

\put(0.5,06.5){\line(1,0){99}} \put(0.5,15.5){\line(1,0){99}}
\put(0.5,06.5){\line(0,1){09}} \put(99.5,06.5){\line(0,1){09}}

\put(010,11.5){\makebox(0,0)[bl]{Basic typical structural details
 (manufacturing stage):}}

\put(033,07.5){\makebox(0,0)[bl]{\(d_{1}\),\(d_{2}\),\(d_{3}\),\(d_{4}\),\(d_{5}\),\(d_{6}\),\(d_{7}\),\(d_{8}\)}}

\end{picture}
\end{center}

 The examined problem consists in coordination (balancing) between two stages:

 1. assembly stage
 (i.e., two-year building assembly plan) and

 2. production stage
 (i.e., requirements of basic typical structural details).

 The suggested heuristic solving scheme for
 coordination (balancing) of assembly stage and production
 stage
 is based on correction of
 a generated initial schedule (solution) (Fig. 10).

\begin{center}
\begin{picture}(120,69)
\put(11.5,00){\makebox(0,0)[bl]{Fig. 10.
 Heuristic solving scheme for coordination/balancing }}

\put(05,60){\line(1,0){115}} \put(05,68){\line(1,0){115}}
\put(05,60){\line(0,1){08}} \put(120,60){\line(0,1){08}}

\put(19,64){\makebox(0,0)[bl]{Calculation of required
 structural components (details) }}

\put(30,61){\makebox(0,0)[bl]{for each time interval of the
 schedule }}

\put(58,60){\vector(0,-1){04}}

\put(05,48){\line(1,0){115}} \put(05,56){\line(1,0){115}}
\put(05,48){\line(0,1){08}} \put(120,48){\line(0,1){08}}

\put(06.5,52){\makebox(0,0)[bl]{Comparison (by structural details)
 of manufacturing system productivity}}

\put(09,49){\makebox(0,0)[bl]{and required structural components
(details) (for each time interval)}}

\put(58,48){\vector(0,-1){04}}


\put(58,41){\oval(64,06)}

\put(29,39.5){\makebox(0,0)[bl]{Evaluation of the balance
 (unbalance) }}

\put(58,38){\vector(0,-1){06}}

\put(20.3,34){\makebox(0,0)[bl]{Balance is not sufficient }}

\put(90,41){\vector(1,0){19}}

\put(93,42){\makebox(0,0)[bl]{Balance}}
\put(94,38){\makebox(0,0)[bl]{is OK }}

\put(114,41){\oval(10,06)} \put(114,41){\oval(09,05)}
\put(110.6,40){\makebox(0,0)[bl]{End}}

\put(05,21){\line(1,0){115}} \put(05,32){\line(1,0){115}}
\put(05,21){\line(0,1){11}} \put(120,21){\line(0,1){11}}

\put(012,28){\makebox(0,0)[bl]{Selection of unbalanced time
 interval(s) of the assembly schedule }}

\put(10,25){\makebox(0,0)[bl]{while taking into account
 productivity of manufacturing subsystem}}

\put(28,22){\makebox(0,0)[bl]{and requirements of the assembly
 schedule}}


\put(58,21){\vector(0,-1){04}}

\put(05,11.5){\line(-1,0){05}} \put(00,11.5){\line(0,1){29.5}}
\put(00,41){\vector(1,0){26}}

\put(05,06){\line(1,0){115}} \put(05,17){\line(1,0){115}}
\put(05,06){\line(0,1){11}} \put(120,06){\line(0,1){11}}

\put(21,13){\makebox(0,0)[bl]{Correction (improvement) of the
assembly schedule: }}

\put(13,10){\makebox(0,0)[bl]{1. by shifts of schedule parts
 (i.e., composite modular jobs), }}

\put(13,07){\makebox(0,0)[bl]{2. by exchange of schedule parts
 (i.e., composite modular jobs). }}

\end{picture}
\end{center}

 Table 10 contains configurations of the considered
 18-floor building and 22-floor
 building by the typical floors
 (i.e., binary relation \(R^{buil,r}\)).
  An illustration of the relationships of
 the examined objects is depicted in Fig. 11.
 Further,
 the weighted binary relations
 (correspondences)
 \(R^{r,d}\),
 \(R^{a,g\&w}\),
 \(R^{r,g\&w}\),
 \(R^{\tau,r}\),
  \(R^{\tau,d}\),
  are presented in tables.
 In the considered planning problem,
 relation \(R^{\tau,d}\) is the basic one.

\begin{center}
\begin{small}
 {\bf Table 10.} Configuration of typical building by typical floors  \\
\begin{tabular}{|c|c | c    c c c c  c c c c| }
\hline
 No.&Typy of building&Typical floor: &\(r_{1}\)&\(r_{2}\)&\(r_{3}\)&\(r_{4}\)
                &\(r_{5}\)&\(r_{6}\)&\(r_{7}\)&\(r_{8}\)\\

\hline

 1.& \(18\)-floor fuilding&
   &\(0\)&\(1\)&\(0\)&\(11\)
   &\(5\)&\(1\)&\(1\)&\(1\)\\

 1.& \(22\)-floor building&
   &\(0\)&\(1\)&\(4\)&\(11\)
   &\(5\)&\(1\)&\(1\)&\(1\)\\

\hline
\end{tabular}
\end{small}
\end{center}

\begin{center}
\begin{picture}(95,72)
\put(06,00){\makebox(0,0)[bl]{Fig. 11.
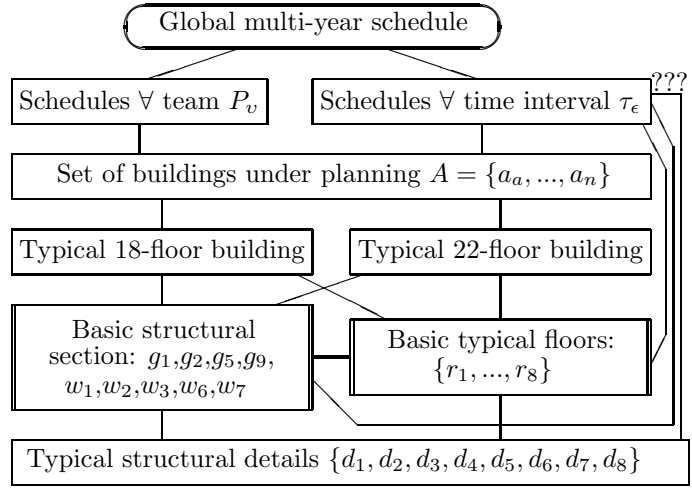
 Relationships of examined components }}

\put(19.6,65.8){\makebox(0,0)[bl]{Global multi-year schedule}}


\put(40,67){\oval(50,06)} \put(40,67){\oval(49.4,05.5)}

\put(30,64){\line(-3,-1){12}} \put(50,64){\line(3,-1){12}}

\put(01,55.5){\makebox(0,0)[bl]{Schedules \(\forall \) team \(P_{\upsilon}\)}}

\put(00,54){\line(1,0){34}} \put(00,60){\line(1,0){34}}
\put(00,54){\line(0,1){06}} \put(34,54){\line(0,1){06}}
%

\put(17,54){\line(0,-1){04}}

\put(41,55.5){\makebox(0,0)[bl]{Schedules \(\forall \) time interval \(\tau_{\epsilon}\)}}

\put(40,54){\line(1,0){45}} \put(40,60){\line(1,0){45}}
\put(40,54){\line(0,1){06}} \put(85,54){\line(0,1){06}}
%

\put(62.5,54){\line(0,-1){04}}

\put(85,58.5){\makebox(0,0)[bl]{???}}

\put(85,58){\line(1,0){04}}
\put(89,58){\line(0,-1){46}}

\put(85,57){\line(1,-2){03}}
\put(88,51){\line(0,-1){37}}
\put(88,14){\line(-1,0){42}}
\put(46,14){\line(-1,1){6}}


\put(84,54){\line(1,-2){03}}
\put(87,48){\line(0,-1){22}}
\put(87,26){\line(-1,-2){02}}

\put(06,45.5){\makebox(0,0)[bl]{Set of buildings under planning
  \( A = \{a_{a},...,a_{n}\} \) }}

\put(00,44){\line(1,0){85}} \put(00,50){\line(1,0){85}}
\put(00,44){\line(0,1){06}} \put(85,44){\line(0,1){06}}

\put(20,44){\line(0,-1){04}}

\put(65,44){\line(0,-1){04}}

\put(46,35.5){\makebox(0,0)[bl]{Typical 22-floor building }}

\put(45,34){\line(1,0){40}} \put(45,40){\line(1,0){40}}
\put(45,34){\line(0,1){06}} \put(85,34){\line(0,1){06}}

\put(65,34){\line(0,-1){06}}

\put(47,34){\line(-3,-1){12}}

\put(01,35.5){\makebox(0,0)[bl]{Typical 18-floor building }}

\put(00,34){\line(1,0){40}} \put(00,40){\line(1,0){40}}
\put(00,34){\line(0,1){06}} \put(40,34){\line(0,1){06}}

\put(20,34){\line(0,-1){04}}

\put(38,34){\line(2,-1){12}}

\put(50,23.5){\makebox(0,0)[bl]{Basic typical floors: }}
\put(56,19.5){\makebox(0,0)[bl]{\(\{r_{1},...,r_{8}\)\} }}

\put(45,18){\line(1,0){40}} \put(45,28){\line(1,0){40}}
\put(45,18){\line(0,1){10}} \put(85,18){\line(0,1){10}}
\put(45.5,18){\line(0,1){10}} \put(84.5,18){\line(0,1){10}}

\put(65,18){\line(0,-1){06}}

\put(40,23.15){\line(1,0){05}}
\put(40,22.9){\line(1,0){05}}

\put(08,25.5){\makebox(0,0)[bl]{Basic structural  }}
\put(05,21.5){\makebox(0,0)[bl]{section:  \(g_{1}\),\(g_{2}\),\(g_{5}\),\(g_{9}\), }}
\put(07,17.5){\makebox(0,0)[bl]{\(w_{1}\),\(w_{2}\),\(w_{3}\),\(w_{6}\),\(w_{7}\) }}

\put(00,16){\line(1,0){40}} \put(00,30){\line(1,0){40}}
\put(00,16){\line(0,1){14}} \put(40,16){\line(0,1){14}}
\put(0.5,16){\line(0,1){14}} \put(39.5,16){\line(0,1){14}}


\put(20,16){\line(0,-1){04}}

\put(02,07.5){\makebox(0,0)[bl]{Typical structural details
 \(\{d_{1},d_{2},d_{3},d_{4},d_{5},d_{6},d_{7},d_{8}\}\) }}

\put(00,06){\line(1,0){90}} \put(00,12){\line(1,0){90}}
\put(00,06){\line(0,1){06}} \put(90,06){\line(0,1){06}}

\end{picture}
\end{center}

 Correspondences between
 typical floors and
  basic types of structural details (manufacturing stage)
 for each structural section
 are pointed out in the following tables:
 (1) for structural section \(g_{1}\) (Table 11),
 (2) for structural section \(g_{2}\) (Table 12),
 (3) for structural section \(g_{5}\) (Table 13),
 (4) for structural section \(g_{9}\) (Table 14),
%
%
 (5) for structural section \(w_{1}\) (Table 15),
 (6) for structural section \(w_{2}\) (Table 16),
 (7) for structural section \(w_{3}\) (Table 17),
 (8) for structural section \(w_{6}\) (Table 18),
 and
 (9) for structural section \(w_{7}\) (Table 19).


 Further, a numerical example for 9 building is described.
 Table 20 contains the buildings (as a versions of composite jobs)
 which are under planning (i.e., construction/assembly).

\begin{center}
\begin{small}
 {\bf Table 11.}
 Correspondence of typical floors and structural details  (for structural section \(g_{1}\)) \\
\begin{tabular}{|c|c | c  c c c c c c c c | }
\hline
 No.&Typical floor&Structural details:&\(d_{1}\)&\(d_{2}\)&\(d_{3}\)&\(d_{4}\)
       &\(d_{5}\)&\(d_{6}\)&\(d_{7}\)&\(d_{8}\) \\


\hline

 1.&\(r_{1}\)&
    &\(0.0\)&\(0.0\)&\(0.0\)&\(0.0\)
   &\(0.0\)&\(0.0\)&\(0.0\)&\(0.0\)\\

 2.&\(r_{2}\)&
    &\(19.0\)&\(28.0\)&\(21.0\)&\(0.0\) &\(0.0\)
   &\(2.0\)&\(2.0\)&\(1.0\)\\

 3.&\(r_{3}\)&
    &\(16.0\)&\(0.0\)&\(27.0\)&\(19.0\)&\(0.0\)
   &\(2.0\)&\(5.0\)&\(1.0\)\\

 4.&\(r_{4}\)&
    &\(16.0\)&\(0.0\)&\(27.0\)&\(19.0\)&\(0.0\)
   &\(2.0\)&\(6.0\)&\(1.0\)\\

 5.&\(r_{5}\)&
    &\(16.0\)&\(0.0\)&\(27.0\)&\(19.0\)&\(0.0\)
   &\(2.0\)&\(6.0\)&\(1.0\)\\

 6.&\(r_{6}\)&
    &\(16.0\)&\(0.0\)&\(27.0\)&\(19.0\)&\(0.0\)
   &\(2.0\)&\(6.0\)&\(1.0\)\\

 7.&\(r_{7}\)&
    &\(20.0\)&\(0.0\)&\(33.0\)&\(6.0\)&\(22.0\)
   &\(1.0\)&\(5.0\)&\(0.0\)\\

 8.&\(r_{8}\)&
    &\(10.0\)&\(0.0\)&\(3.0\)&\(0.0\)&\(6.0\)
   &\(0.0\)&\(1.0\)&\(0.0\)\\

\hline
\end{tabular}
\end{small}
\end{center}

\begin{center}
\begin{small}
 {\bf Table 12.}
 Correspondence of typical floors and structural details  (for structural section \(g_{2}\)) \\
\begin{tabular}{|c|c | c  c c c c  c c c c| }
\hline
 No.&Typical flor&Structural details:&\(d_{1}\)&\(d_{2}\)&\(d_{3}\)&\(d_{4}\)
       &\(d_{5}\)&\(d_{6}\)&\(d_{7}\)&\(d_{8}\)\\


\hline

 1.&\(r_{1}\)&
    &\(0.0\)&\(0.0\)&\(0.0\)&\(0.0\)
   &\(0.0\)&\(0.0\)&\(0.0\)&\(0.0\)\\

 2.&\(r_{2}\)&
    &\(19.0\)&\(28.0\)&\(21.0\)&\(0.0\) &\(0.0\)
   &\(2.0\)&\(2.0\)&\(1.0\)\\

 3.&\(r_{3}\)&
    &\(16.0\)&\(0.0\)&\(27.0\)&\(19.0\)&\(0.0\)
   &\(2.0\)&\(5.0\)&\(1.0\)\\

 4.&\(r_{4}\)&
    &\(16.0\)&\(0.0\)&\(27.0\)&\(19.0\)&\(0.0\)
   &\(2.0\)&\(6.0\)&\(1.0\)\\

 5.&\(r_{5}\)&
    &\(16.0\)&\(0.0\)&\(27.0\)&\(19.0\)&\(0.0\)
   &\(2.0\)&\(6.0\)&\(1.0\)\\

 6.&\(r_{6}\)&
    &\(16.0\)&\(0.0\)&\(27.0\)&\(19.0\)&\(0.0\)
   &\(2.0\)&\(6.0\)&\(1.0\)\\

 7.&\(r_{7}\)&
    &\(20.0\)&\(0.0\)&\(33.0\)&\(6.0\)&\(22.0\)
   &\(1.0\)&\(5.0\)&\(0.0\)\\

 8.&\(r_{8}\)&
    &\(10.0\)&\(0.0\)&\(3.0\)&\(0.0\)&\(6.0\)
   &\(0.0\)&\(1.0\)&\(0.0\)\\

\hline
\end{tabular}
\end{small}
\end{center}

\begin{center}
\begin{small}
 {\bf Table 13.}
 Correspondence of typical floors and structural details  (for structural section \(g_{5}\)) \\
\begin{tabular}{|c|c | c  c c c c  c c c c| }
\hline
 No.&Typical flor&Structural details:&\(d_{1}\)&\(d_{2}\)&\(d_{3}\)&\(d_{4}\)
       &\(d_{5}\)&\(d_{6}\)&\(d_{7}\)&\(d_{8}\)\\


\hline

 1.&\(r_{1}\)&
    &\(0.0\)&\(0.0\)&\(0.0\)&\(0.0\)
   &\(0.0\)&\(0.0\)&\(0.0\)&\(0.0\)\\

 2.&\(r_{2}\)&
    &\(21.0\)&\(30.0\)&\(0.0\)&\(24.0\) &\(0.0\)
   &\(2.0\)&\(2.0\)&\(1.0\)\\

 3.&\(r_{3}\)&
    &\(18.0\)&\(0.0\)&\(30.0\)&\(22.0\)&\(0.0\)
   &\(2.0\)&\(5.0\)&\(1.0\)\\

 4.&\(r_{4}\)&
    &\(19.0\)&\(0.0\)&\(29.0\)&\(22.0\)&\(0.0\)
   &\(2.0\)&\(6.0\)&\(1.0\)\\

 5.&\(r_{5}\)&
    &\(18.0\)&\(0.0\)&\(30.0\)&\(22.0\)&\(0.0\)
   &\(2.0\)&\(6.0\)&\(1.0\)\\

 6.&\(r_{6}\)&
    &\(18.0\)&\(0.0\)&\(30.0\)&\(22.0\)&\(0.0\)
   &\(2.0\)&\(6.0\)&\(1.0\)\\

 7.&\(r_{7}\)&
    &\(23.0\)&\(0.0\)&\(38.0\)&\(6.0\)&\(26.0\)
   &\(1.0\)&\(5.0\)&\(0.0\)\\

 8.&\(r_{8}\)&
    &\(10.0\)&\(0.0\)&\(3.0\)&\(0.0\)&\(6.0\)
   &\(0.0\)&\(1.0\)&\(0.0\)\\

\hline
\end{tabular}
\end{small}
\end{center}

\begin{center}
\begin{small}
 {\bf Table 14.}
 Correspondence of typical floors and structural details  (for structural section \(g_{9}\)) \\
\begin{tabular}{|c|c | c  c c c c   c c c c| }
\hline
 No.&Typical floor&Structural details:&\(d_{1}\)&\(d_{2}\)&\(d_{3}\)&\(d_{4}\)
       &\(d_{5}\)&\(d_{6}\)&\(d_{7}\)&\(d_{8}\)\\


\hline

 1.&\(r_{1}\)&
    &\(0.0\)&\(0.0\)&\(0.0\)&\(0.0\)
   &\(0.0\)&\(0.0\)&\(0.0\)&\(0.0\)\\

 2.&\(r_{2}\)&
    &\(22.0\)&\(28.0\)&\(0.0\)&\(24.0\) &\(0.0\)
   &\(2.0\)&\(2.0\)&\(1.0\)\\

 3.&\(r_{3}\)&
    &\(19.0\)&\(0.0\)&\(27.0\)&\(22.0\)&\(0.0\)
   &\(2.0\)&\(5.0\)&\(1.0\)\\

 4.&\(r_{4}\)&
    &\(19.0\)&\(0.0\)&\(27.0\)&\(22.0\)&\(0.0\)
   &\(2.0\)&\(6.0\)&\(1.0\)\\

 5.&\(r_{5}\)&
    &\(19.0\)&\(0.0\)&\(27.0\)&\(22.0\)&\(0.0\)
   &\(2.0\)&\(6.0\)&\(1.0\)\\

 6.&\(r_{6}\)&
    &\(19.0\)&\(0.0\)&\(27.0\)&\(22.0\)&\(0.0\)
   &\(2.0\)&\(6.0\)&\(1.0\)\\

 7.&\(r_{7}\)&
    &\(25.0\)&\(0.0\)&\(45.0\)&\(6.0\)&\(29.0\)
   &\(1.0\)&\(5.0\)&\(0.0\)\\

 8.&\(r_{8}\)&
    &\(10.0\)&\(0.0\)&\(3.0\)&\(0.0\)&\(6.0\)
   &\(0.0\)&\(1.0\)&\(0.0\)\\

\hline
\end{tabular}
\end{small}
\end{center}

\begin{center}
\begin{small}
 {\bf Table 15.}
 Correspondence of typical floors and structural details  (for structural section \(w_{1}\)) \\
\begin{tabular}{|c|c | c  c c c   c c c c c| }
\hline
 No.&Typical floor&Structural details:&\(d_{1}\)&\(d_{2}\)&\(d_{3}\)&\(d_{4}\)
       &\(d_{5}\)&\(d_{6}\)&\(d_{7}\)&\(d_{8}\)\\


\hline

 1.&\(r_{1}\)&
    &\(2.0\)&\(2.0\)&\(0.0\)&\(0.0\)
   &\(0.0\)&\(0.0\)&\(0.0\)&\(0.0\)\\

 2.&\(r_{2}\)&
    &\(1.0\)&\(2.0\)&\(0.0\)&\(0.0\) &\(0.0\)
   &\(0.0\)&\(0.0\)&\(2.0\)\\

 3.&\(r_{3}\)&
    &\(1.0\)&\(0.0\)&\(2.0\)&\(0.0\)&\(0.0\)
   &\(0.0\)&\(0.0\)&\(2.0\)\\

 4.&\(r_{4}\)&
    &\(1.0\)&\(0.0\)&\(2.0\)&\(0.0\)&\(0.0\)
   &\(0.0\)&\(0.0\)&\(2.0\)\\

 5.&\(r_{5}\)&
    &\(1.0\)&\(0.0\)&\(2.0\)&\(0.0\)&\(0.0\)
   &\(0.0\)&\(0.0\)&\(2.0\)\\

 6.&\(r_{6}\)&
    &\(1.0\)&\(0.0\)&\(2.0\)&\(0.0\)&\(0.0\)
   &\(0.0\)&\(0.0\)&\(1.0\)\\

 7.&\(r_{7}\)&
    &\(0.0\)&\(0.0\)&\(2.0\)&\(0.0\)&\(0.0\)
   &\(0.0\)&\(0.0\)&\(2.0\)\\

 8.&\(r_{8}\)&
    &\(0.0\)&\(0.0\)&\(0.0\)&\(0.0\)&\(0.0\)
   &\(0.0\)&\(0.0\)&\(0.0\)\\

\hline
\end{tabular}
\end{small}
\end{center}

\newpage
\begin{center}
\begin{small}
 {\bf Table 16.}
 Correspondence of typical floors and structural details  (for structural section \(w_{2}\)) \\
\begin{tabular}{|c|c | c  c c c c  c c c c | }
\hline
 No.&Typical floor&Structural details:&\(d_{1}\)&\(d_{2}\)&\(d_{3}\)&\(d_{4}\)
       &\(d_{5}\)&\(d_{6}\)&\(d_{7}\)&\(d_{8}\)\\


\hline

 1.&\(r_{1}\)&
    &\(0.0\)&\(1.0\)&\(0.0\)&\(0.0\)
   &\(0.0\)&\(0.0\)&\(0.0\)&\(0.0\)\\

 2.&\(r_{2}\)&
    &\(0.0\)&\(1.0\)&\(0.0\)&\(0.0\) &\(0.0\)
   &\(0.0\)&\(0.0\)&\(0.0\)\\

 3.&\(r_{3}\)&
    &\(0.0\)&\(0.0\)&\(1.0\)&\(0.0\)&\(0.0\)
   &\(0.0\)&\(0.0\)&\(0.0\)\\

 4.&\(r_{4}\)&
    &\(0.0\)&\(0.0\)&\(1.0\)&\(0.0\)&\(0.0\)
   &\(0.0\)&\(0.0\)&\(0.0\)\\

 5.&\(r_{5}\)&
    &\(0.0\)&\(0.0\)&\(1.0\)&\(0.0\)&\(0.0\)
   &\(0.0\)&\(0.0\)&\(0.0\)\\

 6.&\(r_{6}\)&
    &\(0.0\)&\(0.0\)&\(1.0\)&\(0.0\)&\(0.0\)
   &\(0.0\)&\(0.0\)&\(0.0\)\\

 7.&\(r_{7}\)&
    &\(0.0\)&\(0.0\)&\(0.0\)&\(0.0\)&\(0.0\)
   &\(0.0\)&\(0.0\)&\(0.0\)\\

 8.&\(r_{8}\)&
    &\(0.0\)&\(0.0\)&\(0.0\)&\(0.0\)&\(0.0\)
   &\(0.0\)&\(0.0\)&\(0.0\)\\

\hline
\end{tabular}
\end{small}
\end{center}

\begin{center}
\begin{small}
 {\bf Table 17.}
 Correspondence of typical floors and structural details  (for structural section \(w_{3}\)) \\
\begin{tabular}{|c|c | c  c c c c  c c c c | }
\hline
 No.&Typical floor&Structural details:&\(d_{1}\)&\(d_{2}\)&\(d_{3}\)&\(d_{4}\)
       &\(d_{5}\)&\(d_{6}\)&\(d_{7}\)&\(d_{8}\)\\


\hline

 1.&\(r_{1}\)&
    &\(1.0\)&\(1.0\)&\(0.0\)&\(0.0\)
   &\(0.0\)&\(0.0\)&\(0.0\)&\(0.0\)\\

 2.&\(r_{2}\)&
    &\(1.0\)&\(1.0\)&\(0.0\)&\(0.0\) &\(0.0\)
   &\(0.0\)&\(0.0\)&\(0.0\)\\

 3.&\(r_{3}\)&
    &\(1.0\)&\(0.0\)&\(1.0\)&\(0.0\)&\(0.0\)
   &\(0.0\)&\(0.0\)&\(0.0\)\\

 4.&\(r_{4}\)&
    &\(1.0\)&\(0.0\)&\(1.0\)&\(0.0\)&\(0.0\)
   &\(0.0\)&\(0.0\)&\(0.0\)\\

 5.&\(r_{5}\)&
   &\(1.0\)&\(0.0\)&\(1.0\)&\(0.0\)&\(0.0\)
   &\(0.0\)&\(0.0\)&\(0.0\)\\

 6.&\(r_{6}\)&
   &\(1.0\)&\(0.0\)&\(1.0\)&\(0.0\)&\(0.0\)
   &\(0.0\)&\(0.0\)&\(0.0\)\\

 7.&\(r_{7}\)&
   &\(1.0\)&\(0.0\)&\(1.0\)&\(0.0\)&\(0.0\)
   &\(0.0\)&\(0.0\)&\(0.0\)\\

 8.&\(r_{8}\)&
   &\(0.0\)&\(0.0\)&\(0.0\)&\(0.0\)&\(0.0\)
   &\(0.0\)&\(0.0\)&\(0.0\)\\

\hline
\end{tabular}
\end{small}
\end{center}

\begin{center}
\begin{small}
 {\bf Table 18.}
 Correspondence of typical floors and structural details  (for structural section \(w_{6}\)) \\
\begin{tabular}{|c|c | c  c c c c  c c c c | }
\hline
 No.&Typical floor&Structural details:&\(d_{1}\)&\(d_{2}\)&\(d_{3}\)&\(d_{4}\)
       &\(d_{5}\)&\(d_{6}\)&\(d_{7}\)&\(d_{8}\)\\


\hline

 1.&\(r_{1}\)&
   &\(3.0\)&\(2.0\)&\(0.0\)&\(0.0\)
   &\(0.0\)&\(0.0\)&\(0.0\)&\(0.0\)\\

 2.&\(r_{2}\)&
   &\(3.0\)&\(2.0\)&\(0.0\)&\(0.0\)
   &\(0.0\)&\(0.0\)&\(0.0\)&\(2.0\)\\

 3.&\(r_{3}\)&
   &\(3.0\)&\(0.0\)&\(2.0\)&\(0.0\)
   &\(0.0\)&\(0.0\)&\(0.0\)&\(2.0\)\\

 4.&\(r_{4}\)&
   &\(3.0\)&\(0.0\)&\(2.0\)&\(0.0\)
   &\(0.0\)&\(0.0\)&\(0.0\)&\(2.0\)\\

 5.&\(r_{5}\)&
   &\(3.0\)&\(0.0\)&\(2.0\)&\(0.0\)
   &\(0.0\)&\(0.0\)&\(0.0\)&\(2.0\)\\

 6.&\(r_{6}\)&
   &\(3.0\)&\(0.0\)&\(2.0\)&\(0.0\)
   &\(0.0\)&\(0.0\)&\(0.0\)&\(1.0\)\\

 7.&\(r_{7}\)&
   &\(2.0\)&\(0.0\)&\(3.0\)&\(0.0\)
   &\(0.0\)&\(0.0\)&\(0.0\)&\(1.0\)\\

 8.&\(r_{8}\)&
   &\(0.0\)&\(0.0\)&\(0.0\)&\(0.0\)
   &\(0.0\)&\(0.0\)&\(0.0\)&\(0.0\)\\

\hline
\end{tabular}
\end{small}
\end{center}

\begin{center}
\begin{small}
 {\bf Table 19.}
 Correspondence of typical floors and structural details  (for structural section \(w_{7}\)) \\
\begin{tabular}{|c|c | c  c c c c  c c c c | }
\hline
 No.&Typical floor&Structural details:&\(d_{1}\)&\(d_{2}\)&\(d_{3}\)&\(d_{4}\)
       &\(d_{5}\)&\(d_{6}\)&\(d_{7}\)&\(d_{8}\)\\


\hline

 1.&\(r_{1}\)&
   &\(3.0\)&\(2.0\)&\(0.0\)&\(0.0\)
   &\(0.0\)&\(0.0\)&\(0.0\)&\(0.0\)\\

 2.&\(r_{2}\)&
   &\(3.0\)&\(2.0\)&\(0.0\)&\(0.0\)
   &\(0.0\)&\(0.0\)&\(0.0\)&\(2.0\)\\

 3.&\(r_{3}\)&
   &\(3.0\)&\(2.0\)&\(0.0\)&\(0.0\)
   &\(0.0\)&\(0.0\)&\(0.0\)&\(2.0\)\\

 4.&\(r_{4}\)&
   &\(3.0\)&\(2.0\)&\(0.0\)&\(0.0\)
   &\(0.0\)&\(0.0\)&\(0.0\)&\(2.0\)\\

 5.&\(r_{5}\)&
   &\(3.0\)&\(2.0\)&\(0.0\)&\(0.0\)
   &\(0.0\)&\(0.0\)&\(0.0\)&\(2.0\)\\

 6.&\(r_{6}\)&
   &\(3.0\)&\(2.0\)&\(0.0\)&\(0.0\)
   &\(0.0\)&\(0.0\)&\(0.0\)&\(1.0\)\\

 7.&\(r_{7}\)&
   &\(3.0\)&\(2.0\)&\(0.0\)&\(0.0\)
   &\(0.0\)&\(0.0\)&\(0.0\)&\(1.0\)\\

 8.&\(r_{8}\)&
   &\(0.0\)&\(0.0\)&\(0.0\)&\(0.0\)
   &\(0.0\)&\(0.0\)&\(0.0\)&\(0.0\)\\

\hline
\end{tabular}
\end{small}
\end{center}


\begin{center}
\begin{small}
 {\bf Table 20.} Buildings under planning \\
\begin{tabular}{|c|l|l|c| c|c|c|}
\hline
  Nom.&Notation of &Moscow address&Number of&General&Time of  &Start\\
      &building        &  &floors   &  square &assembly &time \\

\hline

 1.&\(a_{1}\) (V1A)&Vorontzovo, kor. 1A
     &\(18\)&\(17.5\)&\(9.0\)&\(0.5\)\\

 2.&\(a_{2}\) (V2)&Vorontzovo, kor. 2
     &\(22\)&\(16.4\)&\(6.2\)&\(8.0\)\\

 3.&\(a_{3}\) (V5A)&Vorontzovo, kor. 5A
     &\(18\)&\(13.3\)&\(4.5\)&\(6.5\)\\

 4.&\(a_{4}\) (V6A)&Vorontzovo, kor. 6A
     &\(18\)&\(17.7\)&\(4.8\)&\(7.0\)\\

 5.&\(a_{5}\) (B6B)&Vorontzovo, kor. 6B
     &\(22\)&\(24.0\)&\(6.4\)&\(8.8\)\\

 6.&\(a_{6}\) (V2A)&Vorontzovo, kor. 2A
     &\(22\)&\(11.3\)&\(3.0\)&\(9.5\)\\

 7.&\(a_{7}\) (B2A)&Bibirevo, kor. 2A
     &\(22\)&\(16.3\)&\(4.3\)&\(11.8\)\\

 8.&\(a_{8}\) (B2B)&Bibirevo, kor. 2B
     &\(22\)&\(22.7\)&\(6.1\)&\(9.7\)\\

 9.&\(a_{9}\) (B4)&Bibirevo, kor. 4
     &\(22\)&\(29.0\)&\(7.8\)&\(11.0\)\\

\hline
\end{tabular}
\end{small}
\end{center}

 Configurations of the buildings by the typical sections are depicted
 in Table 21.

 The considered initial solution (i.e., assembly plan) for eight flows
 (assembly teams) is depicted in Fig. 12
 (i.e., \(8\) processor schedule \(S^{0}\)).
 The planning time period includes 19 months
 (since January 1982 to July 1983).

\begin{center}
\begin{small}
 {\bf Table 21.} Configuration of buildings by typical sections \\
%
\end{center}

 Table 22 contains the required structural sections
 for each month of the above-mentioned schedule.

\begin{center}
\begin{small}
 {\bf Table 22.} Requirements of structural sections in schedule (for each month) \\
\begin{tabular}{|c|c|c|c|c|c |c|c|c|c|c |c| }
\hline
 No.&  Month
     &\(g_{1}\)&\(g_{2}\)&\(g_{5}\)&\(g_{9}\)
     &\(w_{1}\)&\(w_{2}\)&\(w_{3}\)&\(w_{6}\)&\(w_{7}\)\\

\hline

 1.& 1/1982
   &\(0.091\)&\(0.045\)&\(0.045\)&\(0.0\)
   &\(0.136\) &\(0.0\)&\(0.182\)&\(0.0\)&\(0.0\)\\

 2.& 2/1982
   &\(0.182\)&\(0.091\)&\(0.091\)&\(0.0\)
   &\(0.273\) &\(0.0\)&\(0.364\)&\(0.0\)&\(0.0\)\\

 3.& 3/1982
   &\(0.182\)&\(0.091\)&\(0.091\)&\(0.0\)
   &\(0.273\) &\(0.0\)&\(0.364\)&\(0.0\)&\(0.0\)\\

 4.& 4/1982
   &\(0.182\)&\(0.091\)&\(0.091\)&\(0.0\)
   &\(0.273\) &\(0.0\)&\(0.364\)&\(0.0\)&\(0.0\)\\

 5.& 5/1982
   &\(0.182\)&\(0.091\)&\(0.091\)&\(0.0\)
   &\(0.273\) &\(0.0\)&\(0.364\)&\(0.0\)&\(0.0\)\\

 6.& 6/1982
   &\(0.182\)&\(0.091\)&\(0.091\)&\(0.0\)
   &\(0.273\) &\(0.0\)&\(0.364\)&\(0.0\)&\(0.0\)\\

 7.& 7/1982
   &\(0.274\)&\(0.183\)&\(0.183\)&\(0.0\)
   &\(0.456\) &\(0.0\)&\(0.731\)&\(0.0\)&\(0.0\)\\
\hline
\end{tabular}
\end{small}
\end{center}

\newpage

 Further, requirements of typical floors
 for each plan month are presented:

 (1) requirements of typical floors via \(g_{1}\) (Table 23),

 (2) requirements of typical floors via \(g_{2}\) (Table 24),

 (3) requirements of typical floors via \(g_{5}\) (Table 25),

 (4) requirements of typical floors via \(g_{9}\) (Table 26),




 (5) requirements of typical floors via \(w_{1}\) (Table 27),

 (6) requirements of typical floors via \(w_{2}\) (Table 28),

 (7) requirements of typical floors via \(w_{3}\) (Table 29),

 (8) requirements of typical floors via \(w_{6}\) (Table 30),
 and

 (9) requirements of typical floors via \(w_{7}\) (Table 31).

\begin{center}
\begin{small}
 {\bf Table 23.} Requirements of typical floors via \(g_{1}\) \\
\begin{tabular}{|c|c | c  c c c c  c c c c| }
\hline
 No.&  Month&Typical floor: &\(r_{1}\)&\(r_{2}\)&\(r_{3}\)&\(r_{4}\)
       &\(r_{5}\)&\(r_{6}\)&\(r_{7}\)&\(r_{8}\)\\

\hline

 1.& 1/1982&
   &\(0.0\)&\(2.0\)&\(0.0\)&\(0.1\)
   &\(0.0\)&\(0.0\)&\(0.0\)&\(0.0\)\\

 2.& 2/1982&
   &\(0.0\)&\(0.0\)&\(0.0\)&\(4.2\)
   &\(0.0\)&\(0.0\)&\(0.0\)&\(0.0\)\\

 3.& 3/1982&
   &\(0.0\)&\(0.0\)&\(0.0\)&\(4.2\)
   &\(0.0\)&\(0.0\)&\(0.0\)&\(0.0\)\\

 4.& 4/1982&
   &\(0.0\)&\(0.0\)&\(0.0\)&\(4.2\)
   &\(0.0\)&\(0.0\)&\(0.0\)&\(0.0\)\\

 5.& 5/1982&
   &\(0.0\)&\(0.0\)&\(0.0\)&\(4.2\)
   &\(0.0\)&\(0.0\)&\(0.0\)&\(0.0\)\\

 6.& 6/1982&
   &\(0.0\)&\(0.0\)&\(0.0\)&\(4.2\)
   &\(0.0\)&\(0.0\)&\(0.0\)&\(0.0\)\\

 7.& 7/1982&
   &\(0.0\)&\(1.0\)&\(0.0\)&\(1.9\)
   &\(3.4\)&\(0.0\)&\(0.0\)&\(0.0\)\\
\hline
\end{tabular}
\end{small}
\end{center}

\begin{center}
\begin{small}
 {\bf Table 24.} Requirements of typical floors via \(g_{2}\) \\
\begin{tabular}{|c|c|c c c c c  c c c c| }
\hline
 No.&  Month& Typical floor: & \(r_{1}\)&\(r_{2}\)&\(r_{3}\)&\(r_{4}\)
   &\(r_{5}\)&\(r_{6}\)&\(r_{7}\)&\(r_{8}\)\\

\hline

 1.& 1/1982&
   &\(0.0\)&\(1.0\)&\(0.0\)&\(0.1\)
   &\(0.0\)&\(0.0\)&\(0.0\)&\(0.0\)\\

 2.& 2/1982&
   &\(0.0\)&\(0.0\)&\(0.0\)&\(2.1\)
   &\(0.0\)&\(0.0\)&\(0.0\)&\(0.0\)\\

 3.& 3/1982&
   &\(0.0\)&\(0.0\)&\(0.0\)&\(2.1\)
   &\(0.0\)&\(0.0\)&\(0.0\)&\(0.0\)\\

 4.& 4/1982&
   &\(0.0\)&\(0.0\)&\(0.0\)&\(2.1\)
   &\(0.0\)&\(0.0\)&\(0.0\)&\(0.0\)\\

 5.& 5/1982&
   &\(0.0\)&\(0.0\)&\(0.0\)&\(2.1\)
   &\(0.0\)&\(0.0\)&\(0.0\)&\(0.0\)\\

 6.& 6/1982&
   &\(0.0\)&\(0.0\)&\(0.0\)&\(2.1\)
   &\(0.0\)&\(0.0\)&\(0.0\)&\(0.0\)\\

 7.& 7/1982&
   &\(0.0\)&\(1.0\)&\(0.0\)&\(1.5\)
   &\(1.7\)&\(0.0\)&\(0.0\)&\(0.0\)\\
\hline
\end{tabular}
\end{small}
\end{center}

\newpage
\begin{center}
\begin{small}
 {\bf Table 25.} Requirements of typical floors via \(g_{5}\) \\
\begin{tabular}{|c|c|c c c c c  c c c c| }
\hline
 No.&  Month&Typical floor:&\(r_{1}\)&\(r_{2}\)&\(r_{3}\)&\(r_{4}\)
   &\(r_{5}\)&\(r_{6}\)&\(r_{7}\)&\(r_{8}\)\\

\hline

 1.& 1/1982&
   &\(0.0\)&\(1.0\)&\(0.0\)&\(0.1\)
   &\(0.0\)&\(0.0\)&\(0.0\)&\(0.0\)\\

 2.& 2/1982&
   &\(0.0\)&\(0.0\)&\(0.0\)&\(2.1\)
   &\(0.0\)&\(0.0\)&\(0.0\)&\(0.0\)\\

 3.& 3/1982&
   &\(0.0\)&\(0.0\)&\(0.0\)&\(2.1\)
   &\(0.0\)&\(0.0\)&\(0.0\)&\(0.0\)\\

 4.& 4/1982&
   &\(0.0\)&\(0.0\)&\(0.0\)&\(2.1\)
   &\(0.0\)&\(0.0\)&\(0.0\)&\(0.0\)\\

 5.& 5/1982&
   &\(0.0\)&\(0.0\)&\(0.0\)&\(2.1\)
   &\(0.0\)&\(0.0\)&\(0.0\)&\(0.0\)\\

 6.& 6/1982&
   &\(0.0\)&\(0.0\)&\(0.0\)&\(2.1\)
   &\(0.0\)&\(0.0\)&\(0.0\)&\(0.0\)\\

 7.& 7/1982&
   &\(0.0\)&\(1.0\)&\(0.0\)&\(1.5\)
   &\(1.7\)&\(0.0\)&\(0.0\)&\(0.0\)\\
\hline
\end{tabular}
\end{small}
\end{center}


\begin{center}
\begin{small}
 {\bf Table 26.} Requirements of typical floors via \(g_{9}\) \\
\begin{tabular}{|c|c|c c c c c  c c c c| }
\hline
 No.&  Month&Typical floor:   &\(r_{1}\)&\(r_{2}\)&\(r_{3}\)&\(r_{4}\)
   &\(r_{5}\)&\(r_{6}\)&\(r_{7}\)&\(r_{8}\)\\

\hline

 1.& 1/1982&
   &\(0.0\)&\(0.0\)&\(0.0\)&\(0.0\)
   &\(0.0\)&\(0.0\)&\(0.0\)&\(0.0\)\\

 2.& 2/1982&
   &\(0.0\)&\(0.0\)&\(0.0\)&\(0.0\)
   &\(0.0\)&\(0.0\)&\(0.0\)&\(0.0\)\\

 3.& 3/1982&
   &\(0.0\)&\(0.0\)&\(0.0\)&\(0.0\)
   &\(0.0\)&\(0.0\)&\(0.0\)&\(0.0\)\\

 4.& 4/1982&
   &\(0.0\)&\(0.0\)&\(0.0\)&\(0.0\)
   &\(0.0\)&\(0.0\)&\(0.0\)&\(0.0\)\\

 5.& 5/1982&
   &\(0.0\)&\(0.0\)&\(0.0\)&\(0.0\)
   &\(0.0\)&\(0.0\)&\(0.0\)&\(0.0\)\\

 6.& 6/1982&
   &\(0.0\)&\(0.0\)&\(0.0\)&\(0.0\)
   &\(0.0\)&\(0.0\)&\(0.0\)&\(0.0\)\\

 7.& 7/1982&
   &\(0.0\)&\(0.0\)&\(0.0\)&\(0.0\)
   &\(0.0\)&\(0.0\)&\(0.0\)&\(0.0\)\\
\hline
\end{tabular}
\end{small}
\end{center}

\newpage
\begin{center}
\begin{small}
 {\bf Table 27.} Requirements of typical floors via \(w_{1}\) \\
\begin{tabular}{|c|c|c cccc cccc| }
\hline
 No.&  Month&Typical floor:  &\(r_{1}\)&\(r_{2}\)&\(r_{3}\)&\(r_{4}\)
   &\(r_{5}\)&\(r_{6}\)&\(r_{7}\)&\(r_{8}\)\\

\hline

 1.& 1/1982&
   &\(0.0\)&\(3.0\)&\(0.0\)&\(0.2\)
   &\(0.0\)&\(0.0\)&\(0.0\)&\(0.0\)\\

 2.& 2/1982&
   &\(0.0\)&\(0.0\)&\(0.0\)&\(6.3\)
   &\(0.0\)&\(0.0\)&\(0.0\)&\(0.0\)\\

 3.& 3/1982&
   &\(0.0\)&\(0.0\)&\(0.0\)&\(6.3\)
   &\(0.0\)&\(0.0\)&\(0.0\)&\(0.0\)\\

 4.& 4/1982&
   &\(0.0\)&\(0.0\)&\(0.0\)&\(6.3\)
   &\(0.0\)&\(0.0\)&\(0.0\)&\(0.0\)\\

 5.& 5/1982&
   &\(0.0\)&\(0.0\)&\(0.0\)&\(6.3\)
   &\(0.0\)&\(0.0\)&\(0.0\)&\(0.0\)\\

 6.& 6/1982&
   &\(0.0\)&\(0.0\)&\(0.0\)&\(6.3\)
   &\(0.0\)&\(0.0\)&\(0.0\)&\(0.0\)\\

 7.& 7/1982&
   &\(0.0\)&\(2.0\)&\(0.0\)&\(3.4\)
   &\(5.2\)&\(0.0\)&\(0.0\)&\(0.0\)\\
\hline
\end{tabular}
\end{small}
\end{center}

\begin{center}
\begin{small}
 {\bf Table 28.} Requirements of typical floors via \(w_{2}\) \\
\begin{tabular}{|c|c|c cccc cccc| }
\hline
 No.&Month&Typical floor:&\(r_{1}\)&\(r_{2}\)&\(r_{3}\)&\(r_{4}\)
   &\(r_{5}\)&\(r_{6}\)&\(r_{7}\)&\(r_{8}\)\\

\hline

 1.& 1/1982&
   &\(0.0\)&\(0.0\)&\(0.0\)&\(0.0\)
   &\(0.0\)&\(0.0\)&\(0.0\)&\(0.0\)\\

 2.& 2/1982&
   &\(0.0\)&\(0.0\)&\(0.0\)&\(0.0\)
   &\(0.0\)&\(0.0\)&\(0.0\)&\(0.0\)\\

 3.& 3/1982&
   &\(0.0\)&\(0.0\)&\(0.0\)&\(0.0\)
   &\(0.0\)&\(0.0\)&\(0.0\)&\(0.0\)\\

 4.& 4/1982&
   &\(0.0\)&\(0.0\)&\(0.0\)&\(0.0\)
   &\(0.0\)&\(0.0\)&\(0.0\)&\(0.0\)\\

 5.& 5/1982&
   &\(0.0\)&\(0.0\)&\(0.0\)&\(0.0\)
   &\(0.0\)&\(0.0\)&\(0.0\)&\(0.0\)\\

 6.& 6/1982&
   &\(0.0\)&\(0.0\)&\(0.0\)&\(0.0\)
   &\(0.0\)&\(0.0\)&\(0.0\)&\(0.0\)\\

 7.& 7/1982&
   &\(0.0\)&\(0.0\)&\(0.0\)&\(0.0\)
   &\(0.0\)&\(0.0\)&\(0.0\)&\(0.0\)\\
\hline
\end{tabular}
\end{small}
\end{center}

\newpage
\begin{center}
\begin{small}
 {\bf Table 29.} Requirements of typical floors via \(w_{3}\) \\
\begin{tabular}{|c|c|c cccc cccc| }
\hline
 No.&  Month&Typical floor:   &\(r_{1}\)&\(r_{2}\)&\(r_{3}\)&\(r_{4}\)
   &\(r_{5}\)&\(r_{6}\)&\(r_{7}\)&\(r_{8}\)\\

\hline

 1.& 1/1982&
   &\(0.0\)&\(4.0\)&\(0.0\)&\(0.2\)
   &\(0.0\)&\(0.0\)&\(0.0\)&\(0.0\)\\

 2.& 2/1982&
   &\(0.0\)&\(0.0\)&\(0.0\)&\(8.4\)
   &\(0.0\)&\(0.0\)&\(0.0\)&\(0.0\)\\

 3.& 3/1982&
   &\(0.0\)&\(0.0\)&\(0.0\)&\(8.4\)
   &\(0.0\)&\(0.0\)&\(0.0\)&\(0.0\)\\

 4.& 4/1982&
   &\(0.0\)&\(0.0\)&\(0.0\)&\(8.4\)
   &\(0.0\)&\(0.0\)&\(0.0\)&\(0.0\)\\

 5.& 5/1982&
   &\(0.0\)&\(0.0\)&\(0.0\)&\(8.4\)
   &\(0.0\)&\(0.0\)&\(0.0\)&\(0.0\)\\

 6.& 6/1982&
   &\(0.0\)&\(0.0\)&\(0.0\)&\(8.4\)
   &\(0.0\)&\(0.0\)&\(0.0\)&\(0.0\)\\

 7.& 7/1982&
   &\(0.0\)&\(4.0\)&\(0.0\)&\(6.1\)
   &\(6.9\)&\(0.0\)&\(0.0\)&\(0.0\)\\
\hline
\end{tabular}
\end{small}
\end{center}

\begin{center}
\begin{small}
 {\bf Table 30.} Requirements of typical floors via \(w_{6}\) \\
\begin{tabular}{|c|c|c cccc cccc| }
\hline
 No.&Month&Typical floor:&\(r_{1}\)&\(r_{2}\)&\(r_{3}\)&\(r_{4}\)
   &\(r_{5}\)&\(r_{6}\)&\(r_{7}\)&\(r_{8}\)\\

\hline

 1.& 1/1982&
   &\(0.0\)&\(0.0\)&\(0.0\)&\(0.0\)
   &\(0.0\)&\(0.0\)&\(0.0\)&\(0.0\)\\

 2.& 2/1982&
   &\(0.0\)&\(0.0\)&\(0.0\)&\(0.0\)
   &\(0.0\)&\(0.0\)&\(0.0\)&\(0.0\)\\

 3.& 3/1982&
   &\(0.0\)&\(0.0\)&\(0.0\)&\(0.0\)
   &\(0.0\)&\(0.0\)&\(0.0\)&\(0.0\)\\

 4.& 4/1982&
   &\(0.0\)&\(0.0\)&\(0.0\)&\(0.0\)
   &\(0.0\)&\(0.0\)&\(0.0\)&\(0.0\)\\

 5.& 5/1982&
   &\(0.0\)&\(0.0\)&\(0.0\)&\(0.0\)
   &\(0.0\)&\(0.0\)&\(0.0\)&\(0.0\)\\

 6.& 6/1982&
   &\(0.0\)&\(0.0\)&\(0.0\)&\(0.0\)
   &\(0.0\)&\(0.0\)&\(0.0\)&\(0.0\)\\

 7.& 7/1982&
   &\(0.0\)&\(0.0\)&\(0.0\)&\(0.0\)
   &\(0.0\)&\(0.0\)&\(0.0\)&\(0.0\)\\
\hline
\end{tabular}
\end{small}
\end{center}

\newpage
\begin{center}
\begin{small}
 {\bf Table 31.} Requirements of typical floors via \(w_{7}\) \\
\begin{tabular}{|c|c|c cccc cccc| }
\hline
 No.&  Month&Typical floor:  &\(r_{1}\)&\(r_{2}\)&\(r_{3}\)&\(r_{4}\)
   &\(r_{5}\)&\(r_{6}\)&\(r_{7}\)&\(r_{8}\)\\

\hline

 1.& 1/1982&
   &\(0.0\)&\(0.0\)&\(0.0\)&\(0.0\)
   &\(0.0\)&\(0.0\)&\(0.0\)&\(0.0\)\\

 2.& 2/1982&
   &\(0.0\)&\(0.0\)&\(0.0\)&\(0.0\)
   &\(0.0\)&\(0.0\)&\(0.0\)&\(0.0\)\\

 3.& 3/1982&
   &\(0.0\)&\(0.0\)&\(0.0\)&\(0.0\)
   &\(0.0\)&\(0.0\)&\(0.0\)&\(0.0\)\\

 4.& 4/1982&
   &\(0.0\)&\(0.0\)&\(0.0\)&\(0.0\)
   &\(0.0\)&\(0.0\)&\(0.0\)&\(0.0\)\\

 5.& 5/1982&
   &\(0.0\)&\(0.0\)&\(0.0\)&\(0.0\)
   &\(0.0\)&\(0.0\)&\(0.0\)&\(0.0\)\\

 6.& 6/1982&
   &\(0.0\)&\(0.0\)&\(0.0\)&\(0.0\)
   &\(0.0\)&\(0.0\)&\(0.0\)&\(0.0\)\\

 7.& 7/1982&
   &\(0.0\)&\(0.0\)&\(0.0\)&\(0.0\)
   &\(0.0\)&\(0.0\)&\(0.0\)&\(0.0\)\\
\hline
\end{tabular}
\end{small}
\end{center}

 Further, requirements of the examined assembly schedule
 by typical floors
 ( \(\{r_{1},...,r_{8}  \} \) )
 for each month are presented:
 (1) for January 1982 (Table 32),
 (2) for February 1982 (Table 33),
 (3) for March 1982 (Table 34),
 (4) for April 1982 (Table 35),
 (5) for May 1982 (Table 36),
 (6) for June 1982 (Table 37),
 (7) for July 1982 (Table 38),
 (8) for August 1982 (Table 39),
 (9) for September 1982 (Table 40),
 (10) for October 1982 (Table 41),
 (11) for November 1982 (Table 42),
 (12) for December 1982 (Table 43),
 (13) for January 1983 (Table 44),
 (14) for February 1983 (Table 45),
 (15) for March 1983 (Table 46),
 (16) for April 1983 (Table 47),
 (17) for May 1983 (Table 48),
 (18) for June 1983 (Table 49),
 (19) for July 1983 (Table 50).

\begin{center}
\begin{small}
 {\bf Table 32.} Requirements of typical structural sections/floors in 1/1982 \\
\begin{tabular}{|c|c|c cccc cccc| }
\hline
 No.&  Typical section&Typical floor:&\(r_{1}\)&\(r_{2}\)&\(r_{3}\)&\(r_{4}\)
   &\(r_{5}\)&\(r_{6}\)&\(r_{7}\)&\(r_{8}\)\\


\hline

 1.& \(g_{1}\)&
   &\(0.0\)&\(2.0\)&\(0.0\)&\(0.11\)
   &\(0.0\)&\(0.0\)&\(0.0\)&\(0.0\)\\

 2.&\(g_{2}\)&
   &\(0.0\)&\(1.0\)&\(0.0\)&\(0.06\)
   &\(0.0\)&\(0.0\)&\(0.0\)&\(0.0\)\\

 3.& \(g_{5}\)&
   &\(0.0\)&\(1.0\)&\(0.0\)&\(0.06\)
   &\(0.0\)&\(0.0\)&\(0.0\)&\(0.0\)\\

 4.& \(g_{9}\)&
   &\(0.0\)&\(0.0\)&\(0.0\)&\(0.0\)
   &\(0.0\)&\(0.0\)&\(0.0\)&\(0.0\)\\




 8.& \(w_{1}\)&
   &\(0.0\)&\(3.0\)&\(0.0\)&\(0.17\)
   &\(0.0\)&\(0.0\)&\(0.0\)&\(0.0\)\\

 9.& \(w_{2}\)&
   &\(0.0\)&\(0.0\)&\(0.0\)&\(0.0\)
   &\(0.0\)&\(0.0\)&\(0.0\)&\(0.0\)\\

 10.& \(w_{3}\)&
   &\(0.0\)&\(4.0\)&\(0.0\)&\(0.22\)
   &\(0.0\)&\(0.0\)&\(0.0\)&\(0.0\)\\

 11.& \(w_{6}\)&
   &\(0.0\)&\(0.0\)&\(0.0\)&\(0.0\)
   &\(0.0\)&\(0.0\)&\(0.0\)&\(0.0\)\\

 12.& \(w_{7}\)&
   &\(0.0\)&\(0.0\)&\(0.0\)&\(0.0\)
   &\(0.0\)&\(0.0\)&\(0.0\)&\(0.0\)\\

\hline
\end{tabular}
\end{small}
\end{center}

\begin{center}
\begin{small}
 {\bf Table 33.} Requirements of typical structural sections/floors in 2/1982 \\
\begin{tabular}{|c|c|c cccc cccc| }
\hline
 No.&  Typical section&Typical floor:&\(r_{1}\)&\(r_{2}\)&\(r_{3}\)&\(r_{4}\)
   &\(r_{5}\)&\(r_{6}\)&\(r_{7}\)&\(r_{8}\)\\


\hline

 1.& \(g_{1}\)&
   &\(0.0\)&\(0.0\)&\(0.0\)&\(4.22\)
   &\(0.0\)&\(0.0\)&\(0.0\)&\(0.0\)\\

 2.&\(g_{2}\)&
   &\(0.0\)&\(0.0\)&\(0.0\)&\(2.11\)
   &\(0.0\)&\(0.0\)&\(0.0\)&\(0.0\)\\

 3.& \(g_{5}\)&
   &\(0.0\)&\(0.0\)&\(0.0\)&\(2.11\)
   &\(0.0\)&\(0.0\)&\(0.0\)&\(0.0\)\\

 4.& \(g_{9}\)&
   &\(0.0\)&\(0.0\)&\(0.0\)&\(0.0\)
   &\(0.0\)&\(0.0\)&\(0.0\)&\(0.0\)\\




 5.& \(w_{1}\)&
   &\(0.0\)&\(0.0\)&\(0.0\)&\(6.33\)
   &\(0.0\)&\(0.0\)&\(0.0\)&\(0.0\)\\

 6.& \(w_{2}\)&
   &\(0.0\)&\(0.0\)&\(0.0\)&\(0.0\)
   &\(0.0\)&\(0.0\)&\(0.0\)&\(0.0\)\\

 7.& \(w_{3}\)&
   &\(0.0\)&\(0.0\)&\(0.0\)&\(8.44\)
   &\(0.0\)&\(0.0\)&\(0.0\)&\(0.0\)\\

 8.& \(w_{6}\)&
   &\(0.0\)&\(0.0\)&\(0.0\)&\(0.0\)
   &\(0.0\)&\(0.0\)&\(0.0\)&\(0.0\)\\

 9.& \(w_{7}\)&
   &\(0.0\)&\(0.0\)&\(0.0\)&\(0.0\)
   &\(0.0\)&\(0.0\)&\(0.0\)&\(0.0\)\\

\hline
\end{tabular}
\end{small}
\end{center}

\newpage
\begin{center}
\begin{small}
 {\bf Table 34.} Requirements of typical structural sections/floors in 3/1982 \\
\begin{tabular}{|c|c|c cccc cccc| }
\hline
 No.&Typical section&Typical floor:&\(r_{1}\)&\(r_{2}\)&\(r_{3}\)&\(r_{4}\)
   &\(r_{5}\)&\(r_{6}\)&\(r_{7}\)&\(r_{8}\)\\


\hline

 1.& \(g_{1}\)&
   &\(0.0\)&\(0.0\)&\(0.0\)&\(4.22\)
   &\(0.0\)&\(0.0\)&\(0.0\)&\(0.0\)\\

 2.&\(g_{2}\)&
   &\(0.0\)&\(0.0\)&\(0.0\)&\(2.11\)
   &\(0.0\)&\(0.0\)&\(0.0\)&\(0.0\)\\

 3.& \(g_{5}\)&
   &\(0.0\)&\(0.0\)&\(0.0\)&\(2.11\)
   &\(0.0\)&\(0.0\)&\(0.0\)&\(0.0\)\\

 4.& \(g_{9}\)&
   &\(0.0\)&\(0.0\)&\(0.0\)&\(0.0\)
   &\(0.0\)&\(0.0\)&\(0.0\)&\(0.0\)\\




 5.& \(w_{1}\)&
   &\(0.0\)&\(0.0\)&\(0.0\)&\(6.33\)
   &\(0.0\)&\(0.0\)&\(0.0\)&\(0.0\)\\

 6.& \(w_{2}\)&
   &\(0.0\)&\(0.0\)&\(0.0\)&\(0.0\)
   &\(0.0\)&\(0.0\)&\(0.0\)&\(0.0\)\\

 7.& \(w_{3}\)&
   &\(0.0\)&\(0.0\)&\(0.0\)&\(8.44\)
   &\(0.0\)&\(0.0\)&\(0.0\)&\(0.0\)\\

 8.& \(w_{6}\)&
   &\(0.0\)&\(0.0\)&\(0.0\)&\(0.0\)
   &\(0.0\)&\(0.0\)&\(0.0\)&\(0.0\)\\

 9.& \(w_{7}\)&
   &\(0.0\)&\(0.0\)&\(0.0\)&\(0.0\)
   &\(0.0\)&\(0.0\)&\(0.0\)&\(0.0\)\\

\hline
\end{tabular}
\end{small}
\end{center}

\begin{center}
\begin{small}
 {\bf Table 35.} Requirements of typical structural sections/floors in 4/1982 \\
\begin{tabular}{|c|c|c cccc cccc| }
\hline
 No.&Typical section&Typical floor:&\(r_{1}\)&\(r_{2}\)&\(r_{3}\)&\(r_{4}\)
   &\(r_{5}\)&\(r_{6}\)&\(r_{7}\)&\(r_{8}\)\\


\hline

 1.& \(g_{1}\)&
   &\(0.0\)&\(0.0\)&\(0.0\)&\(4.22\)
   &\(0.0\)&\(0.0\)&\(0.0\)&\(0.0\)\\

 2.&\(g_{2}\)&
   &\(0.0\)&\(0.0\)&\(0.0\)&\(2.11\)
   &\(0.0\)&\(0.0\)&\(0.0\)&\(0.0\)\\

 3.& \(g_{5}\)&
   &\(0.0\)&\(0.0\)&\(0.0\)&\(2.11\)
   &\(0.0\)&\(0.0\)&\(0.0\)&\(0.0\)\\

 4.& \(g_{9}\)&
   &\(0.0\)&\(0.0\)&\(0.0\)&\(0.0\)
   &\(0.0\)&\(0.0\)&\(0.0\)&\(0.0\)\\




 5.& \(w_{1}\)&
   &\(0.0\)&\(0.0\)&\(0.0\)&\(6.33\)
   &\(0.0\)&\(0.0\)&\(0.0\)&\(0.0\)\\

 6.& \(w_{2}\)&
   &\(0.0\)&\(0.0\)&\(0.0\)&\(0.0\)
   &\(0.0\)&\(0.0\)&\(0.0\)&\(0.0\)\\

 7.& \(w_{3}\)&
   &\(0.0\)&\(0.0\)&\(0.0\)&\(8.44\)
   &\(0.0\)&\(0.0\)&\(0.0\)&\(0.0\)\\

 8.& \(w_{6}\)&
   &\(0.0\)&\(0.0\)&\(0.0\)&\(0.0\)
   &\(0.0\)&\(0.0\)&\(0.0\)&\(0.0\)\\

 9.& \(w_{7}\)&
   &\(0.0\)&\(0.0\)&\(0.0\)&\(0.0\)
   &\(0.0\)&\(0.0\)&\(0.0\)&\(0.0\)\\

\hline
\end{tabular}
\end{small}
\end{center}


\begin{center}
\begin{small}
 {\bf Table 36.} Requirements of typical structural sections/floors in 5/1982 \\
\begin{tabular}{|c|c|c cccc cccc| }
\hline
 No.&Typical section&Typical floor:&\(r_{1}\)&\(r_{2}\)&\(r_{3}\)&\(r_{4}\)
   &\(r_{5}\)&\(r_{6}\)&\(r_{7}\)&\(r_{8}\)\\


\hline

 1.& \(g_{1}\)&
   &\(0.0\)&\(0.0\)&\(0.0\)&\(4.22\)
   &\(0.0\)&\(0.0\)&\(0.0\)&\(0.0\)\\

 2.&\(g_{2}\)&
   &\(0.0\)&\(0.0\)&\(0.0\)&\(2.11\)
   &\(0.0\)&\(0.0\)&\(0.0\)&\(0.0\)\\

 3.& \(g_{5}\)&
   &\(0.0\)&\(0.0\)&\(0.0\)&\(2.11\)
   &\(0.0\)&\(0.0\)&\(0.0\)&\(0.0\)\\

 4.& \(g_{9}\)&
   &\(0.0\)&\(0.0\)&\(0.0\)&\(0.0\)
   &\(0.0\)&\(0.0\)&\(0.0\)&\(0.0\)\\




 5.& \(w_{1}\)&
   &\(0.0\)&\(0.0\)&\(0.0\)&\(6.33\)
   &\(0.0\)&\(0.0\)&\(0.0\)&\(0.0\)\\

 6.& \(w_{2}\)&
   &\(0.0\)&\(0.0\)&\(0.0\)&\(0.0\)
   &\(0.0\)&\(0.0\)&\(0.0\)&\(0.0\)\\

 7.& \(w_{3}\)&
   &\(0.0\)&\(0.0\)&\(0.0\)&\(8.44\)
   &\(0.0\)&\(0.0\)&\(0.0\)&\(0.0\)\\

 8.& \(w_{6}\)&
   &\(0.0\)&\(0.0\)&\(0.0\)&\(0.0\)
   &\(0.0\)&\(0.0\)&\(0.0\)&\(0.0\)\\

 9.& \(w_{7}\)&
   &\(0.0\)&\(0.0\)&\(0.0\)&\(0.0\)
   &\(0.0\)&\(0.0\)&\(0.0\)&\(0.0\)\\

\hline
\end{tabular}
\end{small}
\end{center}

\begin{center}
\begin{small}
 {\bf Table 37.} Requirements of typical structural sections/floors in 6/1982 \\
\begin{tabular}{|c|c|c cccc cccc| }
\hline
 No.&Typical section&Typical floor:&\(r_{1}\)&\(r_{2}\)&\(r_{3}\)&\(r_{4}\)
   &\(r_{5}\)&\(r_{6}\)&\(r_{7}\)&\(r_{8}\)\\


\hline

 1.& \(g_{1}\)&
   &\(0.0\)&\(0.0\)&\(0.0\)&\(4.22\)
   &\(0.0\)&\(0.0\)&\(0.0\)&\(0.0\)\\

 2.&\(g_{2}\)&
   &\(0.0\)&\(0.0\)&\(0.0\)&\(2.11\)
   &\(0.0\)&\(0.0\)&\(0.0\)&\(0.0\)\\

 3.& \(g_{5}\)&
   &\(0.0\)&\(0.0\)&\(0.0\)&\(2.11\)
   &\(0.0\)&\(0.0\)&\(0.0\)&\(0.0\)\\

 4.& \(g_{9}\)&
   &\(0.0\)&\(0.0\)&\(0.0\)&\(0.0\)
   &\(0.0\)&\(0.0\)&\(0.0\)&\(0.0\)\\




 5.& \(w_{1}\)&
   &\(0.0\)&\(0.0\)&\(0.0\)&\(6.33\)
   &\(0.0\)&\(0.0\)&\(0.0\)&\(0.0\)\\

 6.& \(w_{2}\)&
   &\(0.0\)&\(0.0\)&\(0.0\)&\(0.0\)
   &\(0.0\)&\(0.0\)&\(0.0\)&\(0.0\)\\

 7.& \(w_{3}\)&
   &\(0.0\)&\(0.0\)&\(0.0\)&\(8.44\)
   &\(0.0\)&\(0.0\)&\(0.0\)&\(0.0\)\\

 8.& \(w_{6}\)&
   &\(0.0\)&\(0.0\)&\(0.0\)&\(0.0\)
   &\(0.0\)&\(0.0\)&\(0.0\)&\(0.0\)\\

 9.& \(w_{7}\)&
   &\(0.0\)&\(0.0\)&\(0.0\)&\(0.0\)
   &\(0.0\)&\(0.0\)&\(0.0\)&\(0.0\)\\

\hline
\end{tabular}
\end{small}
\end{center}

\begin{center}
\begin{small}
 {\bf Table 38.} Requirements of typical structural sections/floors in 7/1982 \\
\begin{tabular}{|c|c|c cccc cccc| }
\hline
 No.&Typical section&Typical floor:&\(r_{1}\)&\(r_{2}\)&\(r_{3}\)&\(r_{4}\)
   &\(r_{5}\)&\(r_{6}\)&\(r_{7}\)&\(r_{8}\)\\


\hline

 1.& \(g_{1}\)&
   &\(0.0\)&\(1.0\)&\(0.0\)&\(1.91\)
   &\(3.44\)&\(0.0\)&\(0.0\)&\(0.0\)\\

 2.&\(g_{2}\)&
   &\(0.0\)&\(1.0\)&\(0.0\)&\(1.52\)
   &\(1.72\)&\(0.0\)&\(0.0\)&\(0.0\)\\

 3.& \(g_{5}\)&
   &\(0.0\)&\(1.0\)&\(0.0\)&\(1.52\)
   &\(1.72\)&\(0.0\)&\(0.0\)&\(0.0\)\\

 4.& \(g_{9}\)&
   &\(0.0\)&\(0.0\)&\(0.0\)&\(0.0\)
   &\(0.0\)&\(0.0\)&\(0.0\)&\(0.0\)\\




 5.& \(w_{1}\)&
   &\(0.0\)&\(2.0\)&\(0.0\)&\(3.43\)
   &\(5.17\)&\(0.0\)&\(0.0\)&\(0.0\)\\

 6.& \(w_{2}\)&
   &\(0.0\)&\(0.0\)&\(0.0\)&\(0.0\)
   &\(0.0\)&\(0.0\)&\(0.0\)&\(0.0\)\\

 7.& \(w_{3}\)&
   &\(0.0\)&\(4.0\)&\(0.0\)&\(6.08\)
   &\(6.08\)&\(6.89\)&\(0.0\)&\(0.0\)\\

 8.& \(w_{6}\)&
   &\(0.0\)&\(0.0\)&\(0.0\)&\(0.0\)
   &\(0.0\)&\(0.0\)&\(0.0\)&\(0.0\)\\

 9.& \(w_{7}\)&
   &\(0.0\)&\(0.0\)&\(0.0\)&\(0.0\)
   &\(0.0\)&\(0.0\)&\(0.0\)&\(0.0\)\\

\hline
\end{tabular}
\end{small}
\end{center}

\begin{center}
\begin{small}
 {\bf Table 39.} Requirements of typical structural sections/floors in 8/1982 \\
\begin{tabular}{|c|c|c cccc cccc| }
\hline
 No.&Typical section&Typical floor:&\(r_{1}\)&\(r_{2}\)&\(r_{3}\)&\(r_{4}\)
   &\(r_{5}\)&\(r_{6}\)&\(r_{7}\)&\(r_{8}\)\\


\hline

 1.& \(g_{1}\)&
   &\(0.0\)&\(0.0\)&\(0.0\)&\(4.26\)
   &\(4.22\)&\(0.0\)&\(0.0\)&\(0.0\)\\

 2.&\(g_{2}\)&
   &\(0.0\)&\(1.0\)&\(0.0\)&\(7.26\)
   &\(2.11\)&\(0.0\)&\(0.0\)&\(0.0\)\\

 3.& \(g_{5}\)&
   &\(0.0\)&\(1.0\)&\(0.0\)&\(7.26\)
   &\(2.11\)&\(0.0\)&\(0.0\)&\(0.0\)\\

 4.& \(g_{9}\)&
   &\(0.0\)&\(1.0\)&\(0.0\)&\(3.0\)
   &\(0.0\)&\(0.0\)&\(0.0\)&\(0.0\)\\




 5.& \(w_{1}\)&
   &\(0.0\)&\(1.0\)&\(0.0\)&\(11.52\)
   &\(6.33\)&\(0.0\)&\(0.0\)&\(0.0\)\\

 6.& \(w_{2}\)&
   &\(0.0\)&\(1.0\)&\(0.0\)&\(3.0\)
   &\(0.0\)&\(0.0\)&\(0.0\)&\(0.0\)\\

 7.& \(w_{3}\)&
   &\(0.0\)&\(6.0\)&\(0.0\)&\(35.04\)
   &\(8.44\)&\(0.0\)&\(0.0\)&\(0.0\)\\

 8.& \(w_{6}\)&
   &\(0.0\)&\(0.0\)&\(0.0\)&\(0.0\)
   &\(0.0\)&\(0.0\)&\(0.0\)&\(0.0\)\\

 9.& \(w_{7}\)&
   &\(0.0\)&\(0.0\)&\(0.0\)&\(0.0\)
   &\(0.0\)&\(0.0\)&\(0.0\)&\(0.0\)\\

\hline
\end{tabular}
\end{small}
\end{center}

\begin{center}
\begin{small}
 {\bf Table 40.} Requirements of typical structural sections/floors in 9/1982 \\
\begin{tabular}{|c|c|c cccc cccc| }
\hline
 No.&Typical section&Typical floor:&\(r_{1}\)&\(r_{2}\)&\(r_{3}\)&\(r_{4}\)
   &\(r_{5}\)&\(r_{6}\)&\(r_{7}\)&\(r_{8}\)\\


\hline

 1.& \(g_{1}\)&
   &\(0.0\)&\(0.0\)&\(0.0\)&\(4.26\)
   &\(2.33\)&\(1.89\)&\(0.0\)&\(0.0\)\\

 2.&\(g_{2}\)&
   &\(0.0\)&\(2.71\)&\(5.42\)&\(8.26\)
   &\(1.17\)&\(0.94\)&\(0.0\)&\(0.0\)\\

 3.& \(g_{5}\)&
   &\(0.0\)&\(0.71\)&\(0.0\)&\(8.26\)
   &\(1.17\)&\(0.94\)&\(0.0\)&\(0.0\)\\

 4.& \(g_{9}\)&
   &\(0.0\)&\(2.43\)&\(2.71\)&\(4.0\)
   &\(0.0\)&\(0.0\)&\(0.0\)&\(0.0\)\\




 5.& \(w_{1}\)&
   &\(0.0\)&\(1.71\)&\(2.71\)&\(12.52\)
   &\(3.50\)&\(2.83\)&\(0.0\)&\(0.0\)\\

 6.& \(w_{2}\)&
   &\(0.0\)&\(2.43\)&\(2.71\)&\(4.0\)
   &\(0.0\)&\(0.0\)&\(0.0\)&\(0.0\)\\

 7.& \(w_{3}\)&
   &\(0.0\)&\(9.71\)&\(10.84\)&\(41.04\)
   &\(4.67\)&\(3.78\)&\(0.0\)&\(0.0\)\\

 8.& \(w_{6}\)&
   &\(0.0\)&\(0.0\)&\(0.0\)&\(0.0\)
   &\(0.0\)&\(0.0\)&\(0.0\)&\(0.0\)\\

 9.& \(w_{7}\)&
   &\(0.0\)&\(1.0\)&\(2.71\)&\(0.0\)
   &\(0.0\)&\(0.0\)&\(0.0\)&\(0.0\)\\

\hline
\end{tabular}
\end{small}
\end{center}

\begin{center}
\begin{small}
 {\bf Table 41.} Requirements of typical structural sections/floors in 10/1982 \\
\begin{tabular}{|c|c|c cccc cccc| }
\hline
 No.&Typical section&Typical floor:&\(r_{1}\)&\(r_{2}\)&\(r_{3}\)&\(r_{4}\)
   &\(r_{5}\)&\(r_{6}\)&\(r_{7}\)&\(r_{8}\)\\


\hline

 1.& \(g_{1}\)&
   &\(0.0\)&\(1.0\)&\(0.13\)&\(1.35\)
   &\(2.91\)&\(0.11\)&\(2.0\)&\(2.0\)\\

 2.&\(g_{2}\)&
   &\(0.0\)&\(2.29\)&\(8.82\)&\(10.19\)
   &\(2.91\)&\(0.06\)&\(1.0\)&\(1.0\)\\

 3.& \(g_{5}\)&
   &\(0.0\)&\(2.29\)&\(6.24\)&\(5.35\)
   &\(2.91\)&\(0.06\)&\(1.0\)&\(1.0\)\\

 4.& \(g_{9}\)&
   &\(0.0\)&\(1.57\)&\(7.98\)&\(6.42\)
   &\(0.0\)&\(0.0\)&\(0.0\)&\(0.0\)\\




 5.& \(w_{1}\)&
   &\(0.0\)&\(3.29\)&\(7.67\)&\(9.12\)
   &\(5.82\)&\(0.17\)&\(3.0\)&\(0.0\)\\

 6.& \(w_{2}\)&
   &\(0.0\)&\(1.57\)&\(7.98\)&\(6.42\)
   &\(0.0\)&\(0.0\)&\(0.0\)&\(0.0\)\\

 7.& \(w_{3}\)&
   &\(0.0\)&\(12.29\)&\(43.51\)&\(39.08\)
   &\(11.64\)&\(0.22\)&\(4.0\)&\(0.0\)\\

 8.& \(w_{6}\)&
   &\(0.0\)&\(0.0\)&\(0.0\)&\(0.0\)
   &\(0.0\)&\(0.0\)&\(0.0\)&\(0.0\)\\

 9.& \(w_{7}\)&
   &\(0.0\)&\(0.0\)&\(1.29\)&\(2.42\)
   &\(0.0\)&\(0.0\)&\(0.0\)&\(0.0\)\\

\hline
\end{tabular}
\end{small}
\end{center}


\begin{center}
\begin{small}
 {\bf Table 42.} Requirements of typical structural sections/floors in 11/1982 \\
\begin{tabular}{|c|c|c cccc cccc| }
\hline
 No.&Typical section&Typical floor:&\(r_{1}\)&\(r_{2}\)&\(r_{3}\)&\(r_{4}\)
   &\(r_{5}\)&\(r_{6}\)&\(r_{7}\)&\(r_{8}\)\\


\hline

 1.& \(g_{1}\)&
   &\(0.0\)&\(0.12\)&\(3.77\)&\(0.0\)
   &\(2.09\)&\(1.0\)&\(1.0\)&\(1.0\)\\

 2.&\(g_{2}\)&
   &\(0.0\)&\(0.12\)&\(5.66\)&\(16.76\)
   &\(6.09\)&\(1.0\)&\(1.0\)&\(1.0\)\\

 3.& \(g_{5}\)&
   &\(0.0\)&\(0.12\)&\(5.66\)&\(9.34\)
   &\(6.09\)&\(1.0\)&\(1.0\)&\(1.0\)\\

 4.& \(g_{9}\)&
   &\(0.0\)&\(0.24\)&\(5.21\)&\(9.40\)
   &\(4.0\)&\(0.0\)&\(0.0\)&\(0.0\)\\




 5.& \(w_{1}\)&
   &\(0.0\)&\(0.24\)&\(9.43\)&\(13.05\)
   &\(8.18\)&\(2.0\)&\(2.0\)&\(0.0\)\\

 6.& \(w_{2}\)&
   &\(0.0\)&\(0.24\)&\(5.21\)&\(9.40\)
   &\(4.0\)&\(0.0\)&\(0.0\)&\(0.0\)\\

 7.& \(w_{3}\)&
   &\(0.0\)&\(0.82\)&\(33.06\)&\(63.60\)
   &\(32.36\)&\(4.0\)&\(4.0\)&\(0.0\)\\

 8.& \(w_{6}\)&
   &\(0.0\)&\(0.0\)&\(0.0\)&\(0.0\)
   &\(0.0\)&\(0.0\)&\(0.0\)&\(0.0\)\\

 9.& \(w_{7}\)&
   &\(0.0\)&\(0.0\)&\(0.0\)&\(3.71\)
   &\(0.0\)&\(0.0\)&\(0.0\)&\(0.0\)\\

\hline
\end{tabular}
\end{small}
\end{center}

\begin{center}
\begin{small}
 {\bf Table 43.} Requirements of typical structural sections/floors in 12/1982 \\
\begin{tabular}{|c|c|c cccc cccc| }
\hline
 No.&Typical section&Typical floor:&\(r_{1}\)&\(r_{2}\)&\(r_{3}\)&\(r_{4}\)
   &\(r_{5}\)&\(r_{6}\)&\(r_{7}\)&\(r_{8}\)\\


\hline

 1.& \(g_{1}\)&
   &\(0.0\)&\(1.88\)&\(2.49\)&\(3.67\)
   &\(0.0\)&\(0.0\)&\(0.0\)&\(0.0\)\\

 2.&\(g_{2}\)&
   &\(0.0\)&\(1.88\)&\(2.49\)&\(19.16\)
   &\(4.17\)&\(1.0\)&\(1.0\)&\(1.0\)\\

 3.& \(g_{5}\)&
   &\(0.0\)&\(0.88\)&\(2.16\)&\(11.74\)
   &\(4.17\)&\(1.0\)&\(1.0\)&\(1.0\)\\

 4.& \(g_{9}\)&
   &\(0.0\)&\(2.76\)&\(4.55\)&\(14.51\)
   &\(1.0\)&\(1.0\)&\(1.0\)&\(1.0\)\\




 5.& \(w_{1}\)&
   &\(0.0\)&\(2.76\)&\(4.65\)&\(19.12\)
   &\(4.17\)&\(1.0\)&\(1.0\)&\(0.0\)\\

 6.& \(w_{2}\)&
   &\(0.0\)&\(2.76\)&\(4.55\)&\(14.51\)
   &\(1.0\)&\(1.0\)&\(1.0\)&\(0.0\)\\

 7.& \(w_{3}\)&
   &\(0.0\)&\(11.18\)&\(16.69\)&\(83.40\)
   &\(18.67\)&\(6.0\)&\(6.0\)&\(0.0\)\\

 8.& \(w_{6}\)&
   &\(0.0\)&\(0.0\)&\(0.0\)&\(0.0\)
   &\(0.0\)&\(0.0\)&\(0.0\)&\(0.0\)\\

 9.& \(w_{7}\)&
   &\(0.0\)&\(1.0\)&\(0.34\)&\(3.71\)
   &\(0.0\)&\(0.0\)&\(0.0\)&\(0.0\)\\

\hline
\end{tabular}
\end{small}
\end{center}

\begin{center}
\begin{small}
 {\bf Table 44.} Requirements of typical structural sections/floors in 1/1983 \\
\begin{tabular}{|c|c|c cccc cccc| }
\hline
 No.&Typical section&Typical floor:&\(r_{1}\)&\(r_{2}\)&\(r_{3}\)&\(r_{4}\)
   &\(r_{5}\)&\(r_{6}\)&\(r_{7}\)&\(r_{8}\)\\


\hline

 1.& \(g_{1}\)&
   &\(0.0\)&\(0.0\)&\(5.60\)&\(6.46\)
   &\(0.0\)&\(0.0\)&\(0.0\)&\(0.0\)\\

 2.&\(g_{2}\)&
   &\(0.0\)&\(0.0\)&\(5.60\)&\(12.35\)
   &\(6.93\)&\(1.0\)&\(1.0\)&\(1.0\)\\

 3.& \(g_{5}\)&
   &\(0.0\)&\(0.0\)&\(1.94\)&\(8.34\)
   &\(1.83\)&\(1.0\)&\(1.0\)&\(1.0\)\\

 4.& \(g_{9}\)&
   &\(0.0\)&\(0.0\)&\(7.55\)&\(15.75\)
   &\(2.55\)&\(0.0\)&\(0.0\)&\(0.0\)\\




 5.& \(w_{1}\)&
   &\(0.0\)&\(0.0\)&\(7.55\)&\(15.95\)
   &\(4.38\)&\(1.0\)&\(1.0\)&\(0.0\)\\

 6.& \(w_{2}\)&
   &\(0.0\)&\(0.0\)&\(7.55\)&\(15.75\)
   &\(2.55\)&\(0.0\)&\(0.0\)&\(0.0\)\\

 7.& \(w_{3}\)&
   &\(0.0\)&\(0.0\)&\(31.90\)&\(71.23\)
   &\(17.53\)&\(4.0\)&\(4.0\)&\(0.0\)\\

 8.& \(w_{6}\)&
   &\(0.0\)&\(0.0\)&\(0.0\)&\(0.0\)
   &\(0.0\)&\(0.0\)&\(0.0\)&\(0.0\)\\

 9.& \(w_{7}\)&
   &\(0.0\)&\(0.0\)&\(3.66\)&\(2.85\)
   &\(2.55\)&\(0.0\)&\(0.0\)&\(0.0\)\\

\hline
\end{tabular}
\end{small}
\end{center}

\begin{center}
\begin{small}
 {\bf Table 45.} Requirements of typical structural sections/floors in 2/1983 \\
\begin{tabular}{|c|c|c cccc cccc| }
\hline
 No.&Typical section&Typical floor:&\(r_{1}\)&\(r_{2}\)&\(r_{3}\)&\(r_{4}\)
   &\(r_{5}\)&\(r_{6}\)&\(r_{7}\)&\(r_{8}\)\\


\hline

 1.& \(g_{1}\)&
   &\(0.0\)&\(0.0\)&\(0.0\)&\(11.85\)
   &\(0.21\)&\(0.0\)&\(0.0\)&\(0.0\)\\

 2.&\(g_{2}\)&
   &\(0.0\)&\(0.0\)&\(0.0\)&\(12.87\)
   &\(7.66\)&\(2.0\)&\(0.74\)&\(0.0\)\\

 3.& \(g_{5}\)&
   &\(0.0\)&\(0.0\)&\(0.0\)&\(7.52\)
   &\(2.76\)&\(0.0\)&\(0.0\)&\(0.0\)\\

 4.& \(g_{9}\)&
   &\(0.0\)&\(0.0\)&\(0.0\)&\(16.84\)
   &\(7.75\)&\(1.0\)&\(0.37\)&\(0.0\)\\




 5.& \(w_{1}\)&
   &\(0.0\)&\(0.0\)&\(0.0\)&\(19.37\)
   &\(5.42\)&\(1.0\)&\(0.37\)&\(0.0\)\\

 6.& \(w_{2}\)&
   &\(0.0\)&\(0.0\)&\(0.0\)&\(16.84\)
   &\(7.75\)&\(1.0\)&\(0.37\)&\(0.0\)\\

 7.& \(w_{3}\)&
   &\(0.0\)&\(0.0\)&\(0.0\)&\(76.86\)
   &\(31.43\)&\(4.0\)&\(1.47\)&\(0.0\)\\

 8.& \(w_{6}\)&
   &\(0.0\)&\(0.0\)&\(0.0\)&\(0.0\)
   &\(0.0\)&\(0.0\)&\(0.0\)&\(0.0\)\\

 9.& \(w_{7}\)&
   &\(0.0\)&\(0.0\)&\(0.0\)&\(5.35\)
   &\(2.45\)&\(1.0\)&\(0.37\)&\(0.0\)\\

\hline
\end{tabular}%
\end{small}
\end{center}


 Finally,
 the scheduling problem can be considered the following.
 There are the following problem components:

 (1) set of processors (teams) \(P= \{ P_{1},...,P_{8}\}\);

 (2) set of buildings under planning: \(A=\{a_{1},...,a_{9}\}\);

 (3)
 19 time intervals (months) of the scheduling process
  (\(\tau_{1}\),...,\(\tau_{\epsilon}\),...,\(\tau_{19}\));

 (4) schedule \(S\) (e.g., Fig. 12);

 (5) clustering solution for schedule \(S\):
 \(\widetilde{X}(S) =  \{ X_{\tau_{1}},..., X_{\tau_{\epsilon}} ,...,X_{\tau_{19}} \}\),
  \( X_{\tau_{\epsilon}} \)
 corresponds to the set of the required  details
 for time interval \(\tau_{\epsilon}\)  (\(\epsilon=\overline{1,19} \));

 (6) estimates as the required typical details (panels)
 for each time interval \(\tau_{\epsilon }\) (\(\epsilon =\overline{1,19}\) ):
    \( \gamma_{d_{1}} ( X_{\tau_{\epsilon }} )\) as the number of required details \(d_{1}\)
  at during time interval \(\tau_{\epsilon }\),
     and so on (Table 51).
  In Table 51,
  \(\mu(\gamma_{d_{\kappa}}) =
 \frac{  \gamma_{d_{\kappa}} } { \sum_{\kappa=\overline{1,8}} \gamma_{d_{\kappa}}  }
 \),
%
  ~\( \kappa = \overline{1,8} \) );

 (7) productivity (by details \(d_{1},...,d_{8}\)) of the production system is
 \(\widehat{\gamma}_{0}=\{ \gamma^{0}_{d_{1}},...,\gamma^{0}_{d_{\kappa}},...,\gamma^{0}_{d_{8}}\}
 \)


\begin{center}
\begin{small}
 {\bf Table 51.} Requirements of initial schedule \(S^{0}\)
  by typical structural details (for each month/interval \(\tau_{\epsilon}\)) \\
\begin{tabular}{|c| cccc cccc| }
\hline
   \(\tau_{\epsilon}\)
  &\(\gamma_{d_{1}}\);\(\mu(\gamma_{d_{1}})\)
  &\(\gamma_{d_{2}}\);\(\mu(\gamma_{d_{2}})\)
  &\(\gamma_{d_{3}}\);\(\mu(\gamma_{d_{3}})\)
  &\(\gamma_{d_{4}}\);\(\mu(\gamma_{d_{4}})\)
  &\(\gamma_{d_{5}}\);\(\mu(\gamma_{d_{5}})\)
  &\(\gamma_{d_{6}}\);\(\mu(\gamma_{d_{6}})\)
  &\(\gamma_{d_{7}}\);\(\mu(\gamma_{d_{7}})\)
  &\(\gamma_{d_{8}}\);\(\mu(\gamma_{d_{8}})\)
 \\

\hline

 1  &\(79;24.7\)&\(122;37.9\)&\(27;8.6\)&\(70;21.9\)&\(0;0.0\)
   &\(8;2.6\)&\(9;2.9\)&\(4;1.3\)\\

 2  &\(137;21.7\)&\(0;0.0\)&\(253;40.0\)&\(166;26.3\)&\(0;0.0\)
   &\(16;2.7\)&\(50;8.0\)&\(8;1.3\)\\

  3 &\(137;21.7\)&\(0;0.0\)&\(253;40.0\)&\(166;26.3\)&\(0;0.0\)
   &\(16;2.7\)&\(50;8.0\)&\(8;1.3\)\\

  4 &\(137;21.7\)&\(0;0.0\)&\(253;40.0\)&\(166;26.3\)&\(0;0.0\)
   &\(16;2.7\)&\(50;8.0\)&\(8;1.3\)\\

  5  &\(137;21.7\)&\(0;0.0\)&\(253;40.0\)&\(166;26.3\)&\(0;0.0\)
   &\(16;2.7\)&\(50;8.0\)&\(8;1.3\)\\

  6 &\(137;21.7\)&\(0;0.0\)&\(253;40.0\)&\(166;26.3\)&\(0;0.0\)
   &\(16;2.7\)&\(50;8.0\)&\(8;1.3\)\\

  7  &\(250;22.3\)&\(92;8.2\)&\(377;33.7\)&\(279;24.9\)&\(0;0.0\)
   &\(29;2.6\)&\(77;6.9\)&\(14;1.3\)\\

 8  &\(576;22.5\)&\(93;3.7\)&\(932;36.5\)&\(659;25.8\)&\(0;0.0\)
   &\(66;2.6\)&\(187;7.3\)&\(41;1.6\)\\

  9  &\(842;23.0\)&\(181;5.0\)&\(1300;35.6\)&\(912;24.9\)&\(0;0.0\)
   &\(94;2.6\)&\(251;6.9\)&\(72;2.0\)\\

 10  &\(1250;23.0\)&\(222;4.1\)&\(1866;34.3\)&\(1277;23.5\)&\(109;2.0\)
   &\(129;2.4\)&\(347;6.4\)&\(101;1.9\)\\

  11  &\(1468;22.6\)&\(18;0.3\)&\(2448;37.7\)&\(1615;24.9\)&\(84;1.3\)
   &\(158;2.4\)&\(461;7.1\)&\(122;1.9\)\\

 12 &\(1562;22.9\)&\(231;3.4\)&\(2385;35.3\)&\(1654;24.2\)&\(94;1.4\)
   &\(164;2.4\)&\(459;6.7\)&\(139;2.0\)\\

 13  &\(1446;22.8\)&\(0;0.0\)&\(2418;38.1\)&\(1589;25.8\)&\(59;0.9\)
   &\(155;2.5\)&\(452;7.1\)&\(146;2.3\)\\

 14  &\(1296;22.7\)&\(0;0.0\)&\(2187;38.3\)&\(1452;25.5\)&\(26;0.5\)
   &\(142;2.5\)&\(428;7.5\)&\(138;2.4\)\\

 15 &\(1156;22.5\)&\(0;0.0\)&\(1938;37.7\)&\(1244;24.2\)&\(80;1.6\)
   &\(121;2.4\)&\(373;7.3\)&\(114;2.2\)\\

  16  &\(965;21.5\)&\(0;0.0\)&\(1554;34.6\)&\(810;18.1\)&\(305;6.8\)
   &\(83;1.9\)&\(279;6.2\)&\(76;1.7\)\\

  17  &\(289;23.3\)&\(0;0.0\)&\(453;36.6\)&\(305;24.6\)&\(16;1.3\)
   &\(29;2.4\)&\(90;7.3\)&\(26;2.1\)\\

  18  &\(261;22.6\)&\(0;0.0\)&\(447;38.6\)&\(305;26.4\)&\(0;0.0\)
   &\(29;2.5\)&\(88;7.6\)&\(26;2.3\)\\

 19  &\(276;21.0\)&\(0;0.0\)&\(409;31.2\)&\(164;12.5\)&\(154;11.8\)
   &\(17;1.4\)&\(68;5.2\)&\(13;1.0\)\\

\hline
\end{tabular}
\end{small}
\end{center}

 The planning problem (as coordination/balancing)
  can be considered as the following:

 ~~

 Find the schedule \(S\)
 and corresponding
 \( \widetilde{X} = \{X_{\tau_{1}},...,X_{\tau_{\epsilon}},...,X_{\tau_{19}}   \} \)

 such  that
   \(
  (\gamma_{d_{1}} ( X_{\tau_{\epsilon }} ),.., \gamma_{d_{8}} ( X_{\tau_{\epsilon }} ))
      \preceq  \widehat{\gamma}_{0}
   ~~ \forall \tau_{\epsilon} = \overline{1,19} \).


~~

 In addition,
 various objective functions can be taking into account.

 The solving scheme may be based of correction (improvement)
 of the initial schedule (here it is \(S^{0}\)).
 The following types of the correction operations
 of the initial schedule
 can be considered:
 (i) right shift of  building \(a_{\zeta}\) in the schedule,
 (ii) left shift of  building \(a_{\zeta}\) in the schedule,
 (iii) exchange of two buildings \(a_{\zeta_{1}}\) and \(a_{\zeta_{2}}\)
 in the schedule:
  (a) case 1: \(a_{\zeta_{1}}\) and \(a_{\zeta_{2}}\)
  belong to the same processor schedule,
  (b) case 2: \(a_{\zeta_{1}}\) and \(a_{\zeta_{2}}\)
  belong to the different processor schedules.

   Note,  the more long building contains less part of
 literal walls (panels) (type \(d_{1}\)).

 In the considered example, the possible correction operations
 of  \(S^{0}\) are pointed out in Table 56.
 Here,
 estimates of profit (i.e., usefulness for the balancing)
  and ``cost'' have only illustrative character.

\begin{center}
 {\bf Table 56.} Correction operations for \(S^{0}\) \( \Longrightarrow \) \(S'\)\\
\begin{tabular}{|c|l|l|c|c |c| }
\hline
 No.  &Building&Change/improvement&Binary&Profit& ``Cost''\\
 \(\iota\)&   &item/operation&variable    &\(c_{\iota,j}\)&\(b_{\iota,j}\)\\
\hline

 1.& \(a_{6}\)
  &\(V^{1}_{1}:\) none &\(y_{1,1}\)&\(0.0\)&\(0.0\)\\

  &&\(V^{1}_{2}\): right shift, 3 days&\(y_{1,2}\)&\(0.5\)&\(1.0\)\\

  &&\(V^{1}_{3}\): right shift, 7 days&\(y_{1,3}\)&\(1.5\)&\(2.0\)\\

  &&\(V^{1}_{4}\): right shift, 14 days&\(y_{1,4}\)&\(2.5\)&\(3.0\)\\

  &&\(V^{1}_{5}\): right shift, 21 days&\(y_{1,5}\)&\(3.5\)&\(4.0\)\\

\hline
 2.& \(a_{7}\)
  &\(V^{2}_{1}:\) none &\(y_{2,1}\)&\(0.0\)&\(0.0\)\\

  &&\(V^{2}_{2}\): right shift, 3 days&\(y_{2,2}\)&\(0.3\)&\(0.5\)\\

  &&\(V^{2}_{3}\): right shift, 7 days&\(y_{2,3}\)&\(1.0\)&\(0.8\)\\

  &&\(V^{2}_{4}\): right shift, 14 days&\(y_{2,4}\)&\(1.5\)&\(1.0\)\\

\hline
 3.& \(a_{8}\)
  &\(V^{3}_{1}:\) none &\(y_{3,1}\)&\(0.0\)&\(0.0\)\\

  &&\(V^{3}_{2}\): right shift, 7 days&\(y_{3,2}\)&\(1.5\)&\(1.0\)\\

  &&\(V^{3}_{3}\): right shift, 14 days&\(y_{3,3}\)&\(2.5\)&\(1.5\)\\

  &&\(V^{3}_{4}\): right shift, 21 days&\(y_{3,4}\)&\(3.5\)&\(2.0\)\\

\hline
 4.& \(a_{3}\),\(a_{6}\)
  &\(V^{4}_{1}:\) none &\(y_{4,1}\)&\(0.0\)&\(0.0\)\\

  &&\(V^{4}_{2}\): position exchange&\(y_{4,2}\)&\(1.5\)&\(2.0\)\\

\hline
\end{tabular}
\end{center}


%
\begin{center}
\begin{picture}(115,87)
\put(07,00){\makebox(0,0)[bl]{Fig. 13. New  eight-processor
 schedule for assembly of buildings}}

\put(10,17){\vector(1,0){100}}

\put(111,15.5){\makebox(0,0)[bl]{\(t\)}}

\put(10,15){\line(0,1){04}} \put(10.1,15){\line(0,1){04}}

\put(15,15.5){\line(0,1){03}} \put(20,15.5){\line(0,1){03}}
\put(25,15.5){\line(0,1){03}} \put(30,15.5){\line(0,1){03}}
\put(35,15.5){\line(0,1){03}} \put(40,15.5){\line(0,1){03}}
\put(45,15.5){\line(0,1){03}} \put(50,15.5){\line(0,1){03}}
\put(55,15.5){\line(0,1){03}} \put(60,15.5){\line(0,1){03}}
\put(65,15.5){\line(0,1){03}}

\put(69.9,15){\line(0,1){04}} \put(70.1,15){\line(0,1){04}}
\put(70,15){\line(0,1){04}}

\put(75,15.5){\line(0,1){03}} \put(80,15.5){\line(0,1){03}}
\put(85,15.5){\line(0,1){03}} \put(90,15.5){\line(0,1){03}}
\put(95,15.5){\line(0,1){03}} \put(100,15.5){\line(0,1){03}}
\put(105,15.5){\line(0,1){03}}

\put(9,11){\makebox(0,0)[bl]{\(0\)}}
\put(14,11){\makebox(0,0)[bl]{\(1\)}}
\put(19,11){\makebox(0,0)[bl]{\(2\)}}
\put(24,11){\makebox(0,0)[bl]{\(3\)}}
\put(29,11){\makebox(0,0)[bl]{\(4\)}}
\put(34,11){\makebox(0,0)[bl]{\(5\)}}
\put(39,11){\makebox(0,0)[bl]{\(6\)}}
\put(44,11){\makebox(0,0)[bl]{\(7\)}}
\put(49,11){\makebox(0,0)[bl]{\(8\)}}
\put(54,11){\makebox(0,0)[bl]{\(9\)}}
\put(58,11){\makebox(0,0)[bl]{\(10\)}}
\put(63,11){\makebox(0,0)[bl]{\(11\)}}
\put(68,11){\makebox(0,0)[bl]{\(12\)}}

\put(74,11){\makebox(0,0)[bl]{\(1\)}}
\put(79,11){\makebox(0,0)[bl]{\(2\)}}
\put(84,11){\makebox(0,0)[bl]{\(3\)}}
\put(89,11){\makebox(0,0)[bl]{\(4\)}}
\put(94,11){\makebox(0,0)[bl]{\(5\)}}
\put(99,11){\makebox(0,0)[bl]{\(6\)}}
\put(104,11){\makebox(0,0)[bl]{\(7\)}}

\put(26,06){\makebox(0,0)[bl]{\(1~~~9~~~8~~~2\) ~~year}}
\put(75,06){\makebox(0,0)[bl]{\(1~~~9~~~8~~~3\) ~~year}}

\put(00,82.5){\makebox(0,0)[bl]{Team}}
\put(02.5,79){\makebox(0,0)[bl]{\(P_{1}\)}}

\put(10,79){\line(0,1){06}} \put(105,79){\line(0,1){06}}
\put(10,80){\line(1,0){95}} \put(10,84){\line(1,0){95}}

\put(00,74.5){\makebox(0,0)[bl]{Team}}
\put(02.5,71){\makebox(0,0)[bl]{\(P_{2}\)}}

\put(10,71){\line(0,1){06}} \put(105,71){\line(0,1){06}}
\put(10,72){\line(1,0){95}} \put(10,76){\line(1,0){95}}

\put(50,72){\makebox(0,0)[bl]{\(a_{4}\)(V5B)}}
\put(45,72){\line(0,1){04}} \put(68,72){\line(0,1){04}}


\put(78.5,72){\makebox(0,0)[bl]{\(a_{7}\)(B2A)}}
\put(78,72){\line(0,1){04}} \put(92,72){\line(0,1){04}}

\put(00,66.5){\makebox(0,0)[bl]{Team}}
\put(02.5,63){\makebox(0,0)[bl]{\(P_{3}\)}}

\put(10,63){\line(0,1){06}} \put(105,63){\line(0,1){06}}
\put(10,64){\line(1,0){95}} \put(10,68){\line(1,0){95}}

\put(23,64){\makebox(0,0)[bl]{\(a_{1}\)(V1A)}}
\put(12,64){\line(0,1){04}} \put(48,64){\line(0,1){04}}

\put(53,64){\makebox(0,0)[bl]{\(a_{6}\)(V6B)}}
\put(48,64){\line(0,1){04}} \put(72.5,64){\line(0,1){04}}

\put(00,58.5){\makebox(0,0)[bl]{Team}}
\put(02.5,55){\makebox(0,0)[bl]{\(P_{4}\)}}

\put(10,55){\line(0,1){06}} \put(105,55){\line(0,1){06}}
\put(10,56){\line(1,0){95}} \put(10,60){\line(1,0){95}}

\put(47,56){\makebox(0,0)[bl]{\(a_{3}\)(V5A)}}
\put(42,56){\line(0,1){04}} \put(65,56){\line(0,1){04}}

\put(78,56){\makebox(0,0)[bl]{\(a_{9}\)(B4)}}
\put(65,56){\line(0,1){04}} \put(102,56){\line(0,1){04}}

\put(00,50.5){\makebox(0,0)[bl]{Team}}
\put(02.5,47.5){\makebox(0,0)[bl]{\(P_{5}\)}}

\put(10,47){\line(0,1){06}} \put(105,47){\line(0,1){06}}
\put(10,48){\line(1,0){95}} \put(10,52){\line(1,0){95}}

\put(64,48){\makebox(0,0)[bl]{\(a_{5}\)(B6A)}}
\put(54,48){\line(0,1){04}} \put(87,48){\line(0,1){04}}

\put(00,42.5){\makebox(0,0)[bl]{Team}}
\put(02.5,39){\makebox(0,0)[bl]{\(P_{6}\)}}

\put(10,39){\line(0,1){06}} \put(105,39){\line(0,1){06}}
\put(10,40){\line(1,0){95}} \put(10,44){\line(1,0){95}}

\put(60,40){\makebox(0,0)[bl]{\(a_{2}\)(V2)}}
\put(50,40){\line(0,1){04}} \put(83,40){\line(0,1){04}}

\put(00,34.5){\makebox(0,0)[bl]{Team}}
\put(02.5,31){\makebox(0,0)[bl]{\(P_{7}\)}}

\put(10,31){\line(0,1){06}} \put(105,31){\line(0,1){06}}
\put(10,32){\line(1,0){95}} \put(10,36){\line(1,0){95}}


\put(00,26.5){\makebox(0,0)[bl]{Team}}
\put(02.5,23){\makebox(0,0)[bl]{\(P_{8}\)}}

\put(10,23){\line(0,1){06}} \put(105,23){\line(0,1){06}}
\put(10,24){\line(1,0){95}} \put(10,28){\line(1,0){95}}


\put(83,24){\makebox(0,0)[bl]{\(a_{8}\)(B2B)}}
\put(74.5,24){\line(0,1){04}} \put(104.6,24){\line(0,1){04}}

\end{picture}
\end{center}

 First, it is assumed the correction operations are independent
 (in other case, it will be necessary to use morphological clique problem
 \cite{lev98,lev12a,lev15}).
%
%
 Thus, multiple choice problem can be used
%
 (\(q_{1}=4\), \(q_{2}=4\), \(q_{3}=4\) , \(q_{4}=2\)):
 \[
 \max \sum_{\iota=1}^{4} \sum_{j=1}^{q_{\iota}} y_{\iota,j} c_{\iota,j} ~~~~
 s.t. ~~~\sum_{\iota=1}^{4}  \sum_{j=1}^{q_{\iota}}  y_{\iota,j} b_{\iota,j} < b_{2}^{constr},
 ~~ \sum_{j=1}^{q_{\iota}} y_{\iota,j} \leq 1 ~~
 \forall \iota=\overline{1,4}, ~ \forall j=\overline{1,q_{\iota}}
 ~~~~ \forall y_{\iota,j} \in \{0,1\}.\]
%
%
 Here, a simplified greedy heuristic can be used
 (i.e., series packing of items via value
 \(  \frac{ c_{\iota,j} } { b_{\iota,j}} \)).
 The illustration of the resultant improvement solution (as improvement of \(S^{0}\))):~
 \( I(S^{0})=
 V^{1}_{1} \star V^{2}_{4} \star V^{3}_{4} \star V^{4}_{1} \).
%
The binary solution is
  \(y_{1,1} = 0\),
 \(y_{2,4} = 1\),
 \(y_{3,4} = 1\),
 \(y_{4,1} = 0\).
 The corresponding schedule  \(S'\) is depicted in Fig. 13.
 Fig. 14 depicts an illustration for the analysis of
 balance by structural details production and requirements
 (by  literal walls (panels)
  for initial schedule \(S^{0}\)
  and
  for corrected (improved) schedule \(S'\).

%

%
\begin{center}
\begin{picture}(115,178)
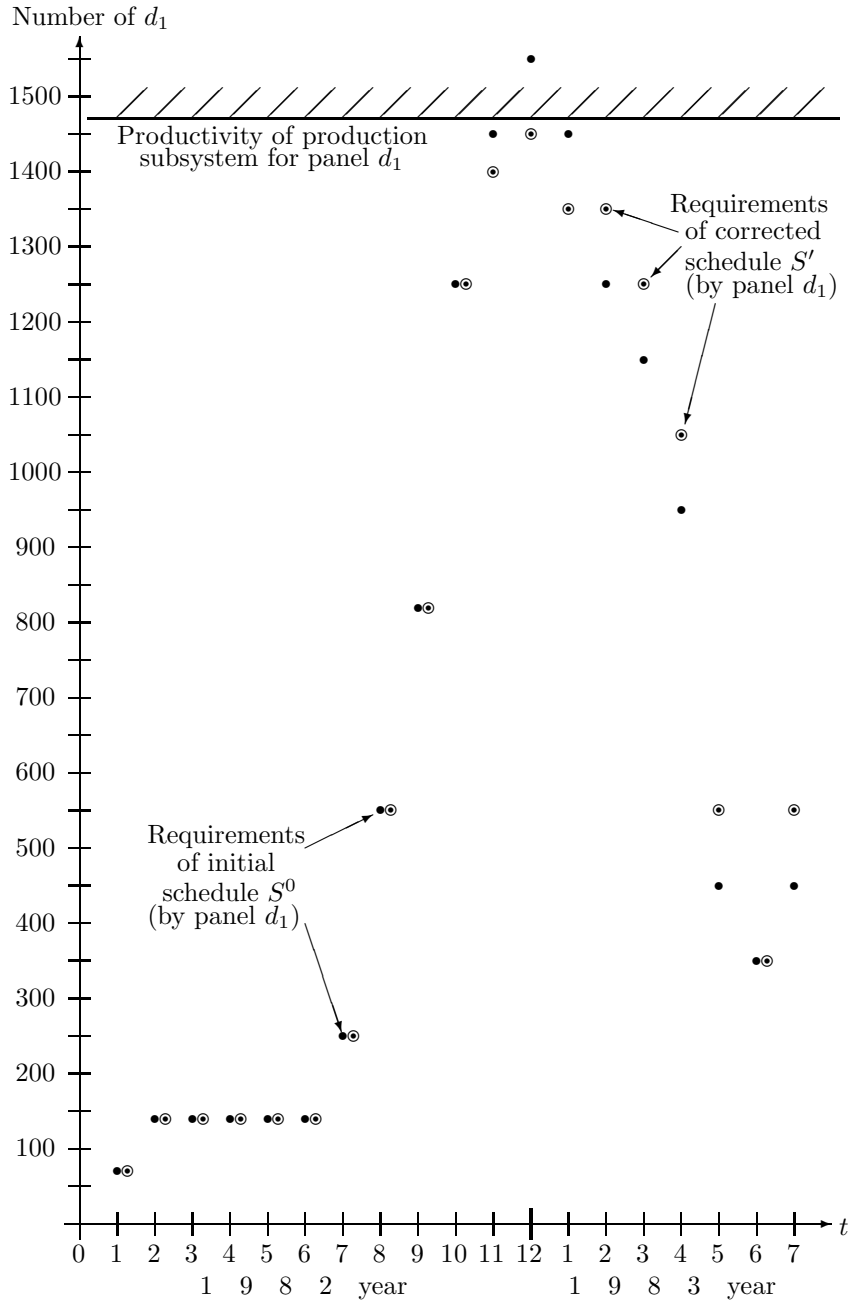
\put(01,00){\makebox(0,0)[bl]{Fig. 14. Analysis of
 balance by structural details
 (by number of detail \(d_{1}\))}}

\put(11,162.1){\line(1,0){100}} \put(11,162){\line(1,0){100}}

\put(15,158){\makebox(0,0)[bl]{Productivity of production}}
\put(18,155){\makebox(0,0)[bl]{subsystem for panel \(d_{1}\)}}

\put(15,162.2){\line(1,1){04}} \put(20,162.2){\line(1,1){04}}
\put(25,162.2){\line(1,1){04}} \put(30,162.2){\line(1,1){04}}
\put(35,162.2){\line(1,1){04}} \put(40,162.2){\line(1,1){04}}
\put(45,162.2){\line(1,1){04}} \put(50,162.2){\line(1,1){04}}

\put(55,162.2){\line(1,1){04}} \put(60,162.2){\line(1,1){04}}
\put(65,162.2){\line(1,1){04}} \put(70,162.2){\line(1,1){04}}
\put(75,162.2){\line(1,1){04}} \put(80,162.2){\line(1,1){04}}
\put(85,162.2){\line(1,1){04}} \put(90,162.2){\line(1,1){04}}

\put(95,162.2){\line(1,1){04}} \put(100,162.2){\line(1,1){04}}
\put(105,162.2){\line(1,1){04}}

\put(15,22){\circle*{0.9}} 

\put(20,29){\circle*{0.9}} 

\put(25,29){\circle*{0.9}} 

\put(30,29){\circle*{0.9}} 

\put(35,29){\circle*{0.9}} 

\put(40,29){\circle*{0.9}} 

\put(45,40){\circle*{0.9}} 

\put(50,70){\circle*{0.9}} 

\put(55,97){\circle*{0.9}} 

\put(60,140){\circle*{0.9}} 

\put(65,160){\circle*{0.9}} 

\put(70,170){\circle*{0.9}} 

\put(19,65){\makebox(0,0)[bl]{Requirements}}
\put(23,62){\makebox(0,0)[bl]{of initial}}
\put(21,58){\makebox(0,0)[bl]{schedule \(S^{0}\)}}
\put(19,54){\makebox(0,0)[bl]{(by panel \(d_{1}\)) }}

\put(40,65){\vector(2,1){09}}

\put(40,55){\vector(1,-3){04.8}}

\put(75,160){\circle*{0.9}} 

\put(80,140){\circle*{0.9}} 

\put(85,130){\circle*{0.9}} 

\put(90,110){\circle*{0.9}} 

\put(95,60){\circle*{0.9}} 

\put(100,50){\circle*{0.9}} 

\put(105,60){\circle*{0.9}} 

\put(16.4,22){\circle*{0.6}} \put(16.4,22){\circle{1.4}}

\put(21.4,29){\circle*{0.6}} \put(21.4,29){\circle{1.4}}

\put(26.4,29){\circle*{0.6}} \put(26.4,29){\circle{1.4}}

\put(31.4,29){\circle*{0.6}} \put(31.4,29){\circle{1.4}}

\put(36.4,29){\circle*{0.6}} \put(36.4,29){\circle{1.4}}

\put(41.4,29){\circle*{0.6}} \put(41.4,29){\circle{1.4}}

\put(46.4,40){\circle*{0.6}} \put(46.4,40){\circle{1.4}}

\put(51.4,70){\circle*{0.6}} \put(51.4,70){\circle{1.4}}

\put(56.4,97){\circle*{0.6}} \put(56.4,97){\circle{1.4}}

\put(61.4,140){\circle*{0.6}} \put(61.4,140){\circle{1.4}}

\put(65,155){\circle*{0.6}} \put(65,155){\circle{1.4}}

\put(70,160){\circle*{0.6}} \put(70,160){\circle{1.4}}

\put(88.5,149){\makebox(0,0)[bl]{Requirements}}
\put(90.5,146){\makebox(0,0)[bl]{of corrected}}
\put(90.5,142){\makebox(0,0)[bl]{schedule \(S'\)}}
\put(90.5,138){\makebox(0,0)[bl]{(by panel \(d_{1}\)) }}

\put(89.6,147){\vector(-3,1){08.6}}
\put(90,145){\vector(-1,-1){04}}
\put(94.5,137.5){\vector(-1,-4){4}}

\put(75,150){\circle*{0.6}} \put(75,150){\circle{1.4}}

\put(80,150){\circle*{0.6}} \put(80,150){\circle{1.4}}

\put(85,140){\circle*{0.6}} \put(85,140){\circle{1.4}}

\put(90,120){\circle*{0.6}} \put(90,120){\circle{1.4}}

\put(95,70){\circle*{0.6}} \put(95,70){\circle{1.4}}

\put(101.4,50){\circle*{0.6}} \put(101.4,50){\circle{1.4}}

\put(105,70){\circle*{0.6}} \put(105,70){\circle{1.4}}

\put(10,15){\vector(0,1){158}}
\put(01,174){\makebox(0,0)[bl]{Number of \(d_{1}\)}}

\put(8.5,20){\line(1,0){03}} \put(8.5,25){\line(1,0){03}}
\put(8.5,30){\line(1,0){03}} \put(8.5,35){\line(1,0){03}}
\put(8.5,40){\line(1,0){03}} \put(8.5,45){\line(1,0){03}}
\put(8.5,50){\line(1,0){03}} \put(8.5,55){\line(1,0){03}}
\put(8.5,60){\line(1,0){03}} \put(8.5,65){\line(1,0){03}}
\put(8.5,70){\line(1,0){03}} \put(8.5,75){\line(1,0){03}}
\put(8.5,80){\line(1,0){03}} \put(8.5,85){\line(1,0){03}}
\put(8.5,90){\line(1,0){03}} \put(8.5,95){\line(1,0){03}}
\put(8.5,100){\line(1,0){03}} \put(8.5,105){\line(1,0){03}}
\put(8.5,110){\line(1,0){03}} \put(8.5,115){\line(1,0){03}}
\put(8.5,120){\line(1,0){03}} \put(8.5,125){\line(1,0){03}}
\put(8.5,130){\line(1,0){03}} \put(8.5,135){\line(1,0){03}}
\put(8.5,140){\line(1,0){03}} \put(8.5,145){\line(1,0){03}}
\put(8.5,150){\line(1,0){03}} \put(8.5,155){\line(1,0){03}}
\put(8.5,160){\line(1,0){03}} \put(8.5,165){\line(1,0){03}}
\put(8.5,170){\line(1,0){03}}


\put(00.5,164){\makebox(0,0)[bl]{\(1500\)}}
\put(00.5,154){\makebox(0,0)[bl]{\(1400\)}}
\put(00.5,144){\makebox(0,0)[bl]{\(1300\)}}
\put(00.5,134){\makebox(0,0)[bl]{\(1200\)}}
\put(00.5,124){\makebox(0,0)[bl]{\(1100\)}}

\put(00.5,114){\makebox(0,0)[bl]{\(1000\)}}
\put(01.5,104){\makebox(0,0)[bl]{\(900\)}}
\put(01.5,094){\makebox(0,0)[bl]{\(800\)}}
\put(01.5,084){\makebox(0,0)[bl]{\(700\)}}
\put(01.5,074){\makebox(0,0)[bl]{\(600\)}}

\put(01.5,064){\makebox(0,0)[bl]{\(500\)}}
\put(01.5,054){\makebox(0,0)[bl]{\(400\)}}
\put(01.5,044){\makebox(0,0)[bl]{\(300\)}}
\put(01.5,034){\makebox(0,0)[bl]{\(200\)}}
\put(01.5,024){\makebox(0,0)[bl]{\(100\)}}

\put(10,15){\vector(1,0){100}}

\put(111,13.5){\makebox(0,0)[bl]{\(t\)}}

\put(10,13){\line(0,1){04}} \put(08,15){\line(1,0){04}}

\put(15,13.5){\line(0,1){03}} \put(20,13.5){\line(0,1){03}}
\put(25,13.5){\line(0,1){03}} \put(30,13.5){\line(0,1){03}}
\put(35,13.5){\line(0,1){03}} \put(40,13.5){\line(0,1){03}}
\put(45,13.5){\line(0,1){03}} \put(50,13.5){\line(0,1){03}}
\put(55,13.5){\line(0,1){03}} \put(60,13.5){\line(0,1){03}}
\put(65,13.5){\line(0,1){03}}

\put(69.9,13){\line(0,1){04}} \put(70.1,13){\line(0,1){04}}
\put(70,13){\line(0,1){04}}

\put(75,13.5){\line(0,1){03}} \put(80,13.5){\line(0,1){03}}
\put(85,13.5){\line(0,1){03}} \put(90,13.5){\line(0,1){03}}
\put(95,13.5){\line(0,1){03}} \put(100,13.5){\line(0,1){03}}
\put(105,13.5){\line(0,1){03}}

\put(09,10){\makebox(0,0)[bl]{\(0\)}}
\put(14,10){\makebox(0,0)[bl]{\(1\)}}
\put(19,10){\makebox(0,0)[bl]{\(2\)}}
\put(24,10){\makebox(0,0)[bl]{\(3\)}}
\put(29,10){\makebox(0,0)[bl]{\(4\)}}
\put(34,10){\makebox(0,0)[bl]{\(5\)}}
\put(39,10){\makebox(0,0)[bl]{\(6\)}}
\put(44,10){\makebox(0,0)[bl]{\(7\)}}
\put(49,10){\makebox(0,0)[bl]{\(8\)}}
\put(54,10){\makebox(0,0)[bl]{\(9\)}}
\put(58,10){\makebox(0,0)[bl]{\(10\)}}
\put(63,10){\makebox(0,0)[bl]{\(11\)}}
\put(68,10){\makebox(0,0)[bl]{\(12\)}}

\put(74,10){\makebox(0,0)[bl]{\(1\)}}
\put(79,10){\makebox(0,0)[bl]{\(2\)}}
\put(84,10){\makebox(0,0)[bl]{\(3\)}}
\put(89,10){\makebox(0,0)[bl]{\(4\)}}
\put(94,10){\makebox(0,0)[bl]{\(5\)}}
\put(99,10){\makebox(0,0)[bl]{\(6\)}}
\put(104,10){\makebox(0,0)[bl]{\(7\)}}

\put(26,05){\makebox(0,0)[bl]{\(1~~~9~~~8~~~2\) ~~year}}
\put(75,05){\makebox(0,0)[bl]{\(1~~~9~~~8~~~3\) ~~year}}

\end{picture}
\end{center}

\newpage
\section{Brief description of two other applications}

\subsection{General framework for supply chain management }

 A special generalized scheme for
  modular manufacturing and assembly
  supply chain is shown in Fig. 15.
 This scheme corresponds to multi-processor scheduling.
 The considered goal consists in coordination (i.e., balancing)
 between stage of modules manufacturing and transportation
 and stage of assemblies.

 Thus, this approach may be considered as a version of
 coordinated supply chain management
 \cite{thom96}.
%
%

\begin{center}
\begin{picture}(114,72)
\put(07,00){\makebox(0,0)[bl]{Fig. 15.
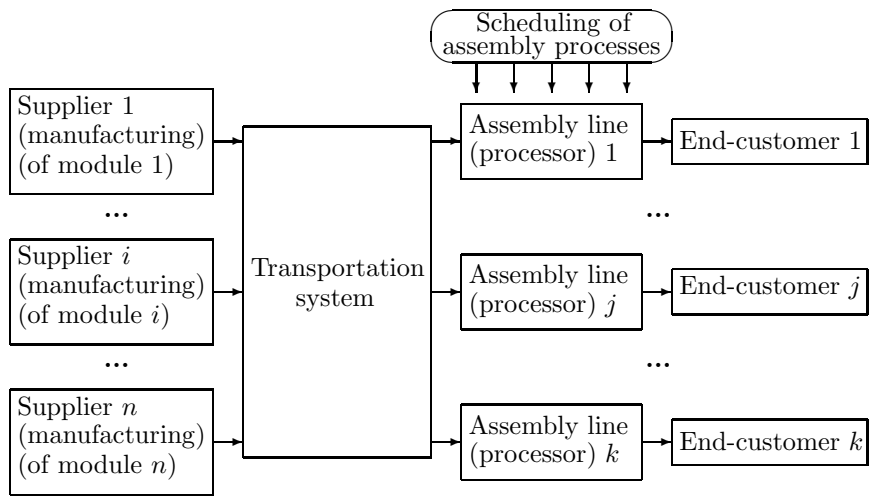
 Modular manufacturing and assembly (as supply chains) }}
\put(00,46){\line(1,0){27}} \put(00,60){\line(1,0){27}}
\put(00,46){\line(0,1){14}} \put(27,46){\line(0,1){14}}

\put(01,56){\makebox(0,0)[bl]{Supplier \(1\)}}
\put(01,52){\makebox(0,0)[bl]{(manufacturing)}}
\put(01,48){\makebox(0,0)[bl]{(of module 1)}}

\put(27,53){\vector(1,0){4}}

\put(12.5,43){\makebox(0,0)[bl]{{\bf ...}}}

\put(00,26){\line(1,0){27}} \put(00,40){\line(1,0){27}}
\put(00,26){\line(0,1){14}} \put(27,26){\line(0,1){14}}

\put(01,36){\makebox(0,0)[bl]{Supplier \(i\)}}
\put(01,32){\makebox(0,0)[bl]{(manufacturing)}}
\put(01,28){\makebox(0,0)[bl]{(of module \(i\))}}

\put(27,33){\vector(1,0){4}}

\put(12.5,23){\makebox(0,0)[bl]{{\bf ...}}}

\put(00,06){\line(1,0){27}} \put(00,20){\line(1,0){27}}
\put(00,06){\line(0,1){14}} \put(27,06){\line(0,1){14}}

\put(01,16){\makebox(0,0)[bl]{Supplier \(n\)}}
\put(01,12){\makebox(0,0)[bl]{(manufacturing)}}
\put(01,08){\makebox(0,0)[bl]{(of module \(n\))}}

\put(27,13){\vector(1,0){4}}

\put(31,11){\line(1,0){25}} \put(31,55){\line(1,0){25}}
\put(31,11){\line(0,1){44}} \put(56,11){\line(0,1){44}}

\put(32,34.5){\makebox(0,0)[bl]{Transportation}}
\put(38,30.5){\makebox(0,0)[bl]{system}}

\put(72,67){\oval(32,07)}

\put(62,67){\makebox(0,0)[bl]{Scheduling of}}
\put(57.5,64){\makebox(0,0)[bl]{assembly processes}}

\put(62,63.5){\vector(0,-1){4}} \put(67,63.5){\vector(0,-1){4}}
\put(72,63.5){\vector(0,-1){4}} \put(77,63.5){\vector(0,-1){4}}
\put(82,63.5){\vector(0,-1){4}}

\put(60,48){\line(1,0){24}} \put(60,58){\line(1,0){24}}
\put(60,48){\line(0,1){10}} \put(84,48){\line(0,1){10}}

\put(61,53.5){\makebox(0,0)[bl]{Assembly line }}
\put(61,49.5){\makebox(0,0)[bl]{(processor) \(1\)}}

\put(56,53){\vector(1,0){4}} \put(84,53){\vector(1,0){4}}

\put(88,50){\line(1,0){26}} \put(88,56){\line(1,0){26}}
\put(88,50){\line(0,1){06}} \put(114,50){\line(0,1){06}}

\put(89,52){\makebox(0,0)[bl]{End-customer \(1\)}}

\put(84.5,43){\makebox(0,0)[bl]{{\bf ...}}}

\put(60,28){\line(1,0){24}} \put(60,38){\line(1,0){24}}
\put(60,28){\line(0,1){10}} \put(84,28){\line(0,1){10}}

\put(61,33.5){\makebox(0,0)[bl]{Assembly line }}
\put(61,29.5){\makebox(0,0)[bl]{(processor) \(j\)}}

\put(56,33){\vector(1,0){4}} \put(84,33){\vector(1,0){4}}

\put(88,30){\line(1,0){26}} \put(88,36){\line(1,0){26}}
\put(88,30){\line(0,1){06}} \put(114,30){\line(0,1){06}}

\put(89,32){\makebox(0,0)[bl]{End-customer \(j\)}}

\put(84.5,23){\makebox(0,0)[bl]{{\bf ...}}}

\put(60,08){\line(1,0){24}} \put(60,18){\line(1,0){24}}
\put(60,08){\line(0,1){10}} \put(84,08){\line(0,1){10}}

\put(61,13.5){\makebox(0,0)[bl]{Assembly line }}
\put(61,9.5){\makebox(0,0)[bl]{(processor) \(k\)}}

\put(56,13){\vector(1,0){4}} \put(84,13){\vector(1,0){4}}

\put(88,10){\line(1,0){26}} \put(88,16){\line(1,0){26}}
\put(88,10){\line(0,1){06}} \put(114,10){\line(0,1){06}}

\put(89,12){\makebox(0,0)[bl]{End-customer \(k\)}}

\end{picture}
\end{center}

\subsection{Planning of multichannel flow in data transmission}

 There are \(n\) initial object-based message sequences
  (they correspond to data sources):

 \(A_{1} = <a_{11},a_{12},a_{13},... >\), ... ,
 \(A_{i} = <a_{i1},a_{i2},a_{i3},... >\), ... ,
 \(A_{n} = <a_{n1},a_{n2},a_{n3},... >\).

 Each message sequence \(A_{i}\) contains a set of standard objects/component
 (e.g., \(b_{i1}\),\(b_{i2}\),\(b_{i3}\),\(b_{i4}\),... )
 with precedence constraint (as a chain):
 ~\(b_{i1} \leftarrow  b_{i2} \leftarrow b_{i3} \leftarrow b_{i4} \leftarrow ...  \).
%

%
 The planning problem consists in scheduling of the messages
 on the basis of \(k\) sub-channels while taking into account
  special requirements as time-interval balancing
 by message objects
 (i.e., for time interval \(\tau_{1}\),\(\tau_{2}\),\(\tau_{3}\), ... ).
 It is assumed
 each message can be sent via each subchannel and with a time
 shift (a delay).

 A scheme of the considered
 time-balanced object-based message scheduling process is shown in
 Fig. 16.

~~

\begin{center}
\begin{picture}(144,60)
\put(17,00){\makebox(0,0)[bl]{Fig. 16.
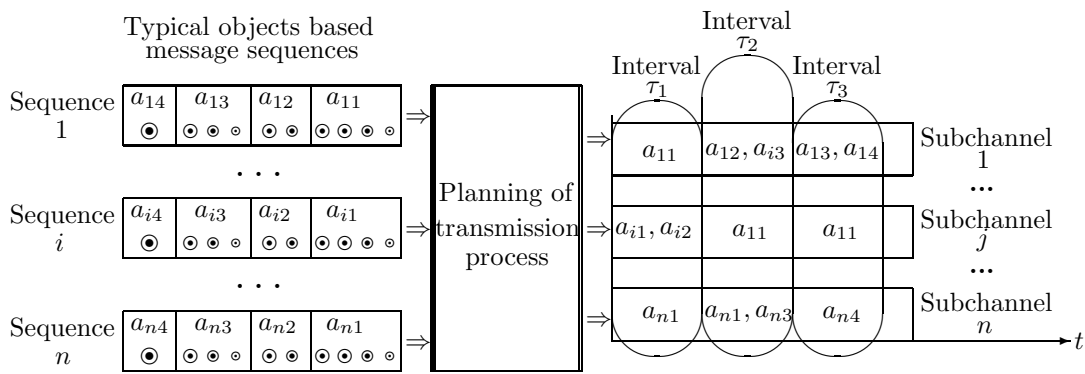
 Scheme of time-balanced object-based message scheduling}}

\put(15,50){\makebox(0,0)[bl]{Typical objects based}}
\put(18,47){\makebox(0,0)[bl]{message sequences}}

\put(00,40){\makebox(0,0)[bl]{Sequence}}
\put(06,37){\makebox(0,0)[bl]{\(1\)}}

\put(15,36){\line(1,0){37}} \put(15,44){\line(1,0){37}}
\put(15,36){\line(0,1){08}} \put(52,36){\line(0,1){08}}

\put(22,36){\line(0,1){08}} \put(32,36){\line(0,1){08}}
\put(40,36){\line(0,1){08}}

\put(52.5,39){\makebox(0,0)[bl]{\(\Rightarrow \)}}

\put(16,41){\makebox(0,0)[bl]{\(a_{14}\)}}
\put(24.5,41){\makebox(0,0)[bl]{\(a_{13}\)}}
\put(33,41){\makebox(0,0)[bl]{\(a_{12}\)}}
\put(42,41){\makebox(0,0)[bl]{\(a_{11}\)}}

\put(18.5,38){\circle*{1.2}} \put(18.5,38){\circle{2.0}}

\put(24,38){\circle*{0.8}} \put(24,38){\circle{1.8}}
\put(27,38){\circle*{0.7}} \put(27,38){\circle{1.4}}
\put(30,38){\circle*{0.5}} \put(30,38){\circle{1.2}}

\put(34.5,38){\circle*{0.8}} \put(34.5,38){\circle{1.8}}
\put(37.5,38){\circle*{0.7}} \put(37.5,38){\circle{1.4}}

\put(41.5,38){\circle*{0.8}} \put(41.5,38){\circle{1.8}}
\put(44.5,38){\circle*{0.8}} \put(44.5,38){\circle{1.8}}
\put(47.5,38){\circle*{0.7}} \put(47.5,38){\circle{1.4}}
\put(50.5,38){\circle*{0.5}} \put(50.5,38){\circle{1.2}}

\put(30,32){\makebox(0,0)[bl]{{\bf .~.~.}}}

\put(00,25){\makebox(0,0)[bl]{Sequence}}
\put(06,22){\makebox(0,0)[bl]{\(i\)}}

\put(15,21){\line(1,0){37}} \put(15,29){\line(1,0){37}}
\put(15,21){\line(0,1){08}} \put(52,21){\line(0,1){08}}

\put(52.5,24){\makebox(0,0)[bl]{\(\Rightarrow \)}}

\put(22,21){\line(0,1){08}} \put(32,21){\line(0,1){08}}
\put(40,21){\line(0,1){08}}

\put(16,26){\makebox(0,0)[bl]{\(a_{i4}\)}}
\put(24.5,26){\makebox(0,0)[bl]{\(a_{i3}\)}}
\put(33,26){\makebox(0,0)[bl]{\(a_{i2}\)}}
\put(42,26){\makebox(0,0)[bl]{\(a_{i1}\)}}

\put(18.5,23){\circle*{1.2}} \put(18.5,23){\circle{2.0}}

\put(24,23){\circle*{0.8}} \put(24,23){\circle{1.8}}
\put(27,23){\circle*{0.7}} \put(27,23){\circle{1.4}}
\put(30,23){\circle*{0.5}} \put(30,23){\circle{1.2}}

\put(34.5,23){\circle*{0.8}} \put(34.5,23){\circle{1.8}}
\put(37.5,23){\circle*{0.7}} \put(37.5,23){\circle{1.4}}

\put(41.5,23){\circle*{0.8}} \put(41.5,23){\circle{1.8}}
\put(44.5,23){\circle*{0.8}} \put(44.5,23){\circle{1.8}}
\put(47.5,23){\circle*{0.7}} \put(47.5,23){\circle{1.4}}
\put(50.5,23){\circle*{0.5}} \put(50.5,23){\circle{1.2}}

\put(30,17){\makebox(0,0)[bl]{{\bf .~.~.}}}

\put(00,10){\makebox(0,0)[bl]{Sequence}}
\put(06,07){\makebox(0,0)[bl]{\(n\)}}

\put(15,06){\line(1,0){37}} \put(15,14){\line(1,0){37}}
\put(15,06){\line(0,1){08}} \put(52,06){\line(0,1){08}}

\put(52.5,09){\makebox(0,0)[bl]{\(\Rightarrow \)}}

\put(22,06){\line(0,1){08}} \put(32,06){\line(0,1){08}}
\put(40,06){\line(0,1){08}}

\put(16,11){\makebox(0,0)[bl]{\(a_{n4}\)}}
\put(24.5,11){\makebox(0,0)[bl]{\(a_{n3}\)}}
\put(33,11){\makebox(0,0)[bl]{\(a_{n2}\)}}
\put(42,11){\makebox(0,0)[bl]{\(a_{n1}\)}}

\put(18.5,08){\circle*{1.2}} \put(18.5,08){\circle{2.0}}

\put(24,08){\circle*{0.8}} \put(24,08){\circle{1.8}}
\put(27,08){\circle*{0.7}} \put(27,08){\circle{1.4}}
\put(30,08){\circle*{0.5}} \put(30,08){\circle{1.2}}

\put(34.5,08){\circle*{0.8}} \put(34.5,08){\circle{1.8}}
\put(37.5,08){\circle*{0.7}} \put(37.5,08){\circle{1.4}}

\put(41.5,08){\circle*{0.8}} \put(41.5,08){\circle{1.8}}
\put(44.5,08){\circle*{0.8}} \put(44.5,08){\circle{1.8}}
\put(47.5,08){\circle*{0.7}} \put(47.5,08){\circle{1.4}}
\put(50.5,08){\circle*{0.5}} \put(50.5,08){\circle{1.2}}


\put(56,06){\line(1,0){20}} \put(56,44){\line(1,0){20}}
\put(56,06){\line(0,1){38}} \put(76,06){\line(0,1){38}}
\put(56.4,06){\line(0,1){38}} \put(75.6,06){\line(0,1){38}}

\put(57,28){\makebox(0,0)[bl]{Planning of }}
\put(56.5,24){\makebox(0,0)[bl]{transmission}}
\put(60.5,19.5){\makebox(0,0)[bl]{process}}

\put(76.5,36){\makebox(0,0)[bl]{\(\Rightarrow \)}}
\put(76.5,24){\makebox(0,0)[bl]{\(\Rightarrow \)}}
\put(76.5,12){\makebox(0,0)[bl]{\(\Rightarrow \)}}

\put(80,10){\line(0,1){30}}

\put(80,10){\vector(1,0){61}}
\put(141.5,09){\makebox(0,0)[bl]{\(t\)}}

\put(80,39){\line(1,0){40}} \put(80,32){\line(1,0){40}}

\put(120,32){\line(0,1){07}}

\put(121,36){\makebox(0,0)[bl]{Subchannel}}
\put(128.5,33){\makebox(0,0)[bl]{\(1\)}}

\put(127.5,30){\makebox(0,0)[bl]{{\bf ...}}}

\put(80,28){\line(1,0){40}} \put(80,21){\line(1,0){40}}

\put(120,21){\line(0,1){07}}

\put(121,25){\makebox(0,0)[bl]{Subchannel}}
\put(128.5,22){\makebox(0,0)[bl]{\(j\)}}

\put(127.5,19){\makebox(0,0)[bl]{{\bf ...}}}

\put(80,17){\line(1,0){40}} \put(120,10){\line(0,1){07}}

\put(121,14){\makebox(0,0)[bl]{Subchannel}}
\put(128.5,11){\makebox(0,0)[bl]{\(n\)}}

\put(86,25){\oval(12,34)}

\put(80,45.5){\makebox(0,0)[bl]{Interval}}
\put(84.5,42.5){\makebox(0,0)[bl]{\(\tau_{1}\)}}

\put(84,34){\makebox(0,0)[bl]{\(a_{11}\)}}
\put(80.4,23.5){\makebox(0,0)[bl]{\(a_{i1},a_{i2}\)}}
\put(84,12.5){\makebox(0,0)[bl]{\(a_{n1}\)}}

\put(98,28){\oval(12,40)}

\put(92,51.5){\makebox(0,0)[bl]{Interval}}
\put(96.5,48.5){\makebox(0,0)[bl]{\(\tau_{2}\)}}

\put(92.4,34){\makebox(0,0)[bl]{\(a_{12},a_{i3}\)}}
\put(96,23.5){\makebox(0,0)[bl]{\(a_{11}\)}}
\put(92.3,12.5){\makebox(0,0)[bl]{\(a_{n1},a_{n3}\)}}

\put(110,25){\oval(12,34)}

\put(104,45.5){\makebox(0,0)[bl]{Interval}}
\put(108.5,42.5){\makebox(0,0)[bl]{\(\tau_{3}\)}}

\put(104.4,34){\makebox(0,0)[bl]{\(a_{13},a_{14}\)}}
\put(108,23.5){\makebox(0,0)[bl]{\(a_{11}\)}}
\put(108,12.5){\makebox(0,0)[bl]{\(a_{n4}\)}}

\end{picture}
\end{center}


\newpage
\section{Conclusion}

 In the paper, a new problem of
 time-interval balancing in multi-processor scheduling of composite modular
 is proposed and described.
 A big illustrative real world example
 for scheduling in homebuilding is presented
 and two prospective applied domains are pointed out.

 The following future research directions can be  pointed out:
%
%
 (1) consideration of the suggested planning problem
 (and close just-in-time scheduling problems)
 from the viewpoint of restructuring of solutions
 \cite{lev15restr};
 (2) analysis and implementation of various solving frameworks
  for the considered planning problem;
%
 (3) taking into account uncertainty in the balancing models;
%
%
 (4) study of other possible application domains for the problem
 (e.g.,
 in communication networking).
%

\section{Acknowledgments}


 The research was
  partially supported by the Russian Science Foundation
 (grant of
 Institute for Information Transmission Problems of Russian
 Academy of Sciences
 14-50-00150, ``Digital technologies and their applications'').





%

\end{document}